# Unboxing Engagement in YouTube Influencer Videos:
# An Attention-Based Approach

Prashant Rajaram*
Puneet Manchanda

This Version: April 30, 2025

* Rajaram (*prajaram@ivey.ca*) is Assistant Professor of Marketing at the Ivey Business School, Western University, and Manchanda (*pmanchan@umich.edu*) is Isadore and Leon Winkelman Professor and Professor of Marketing, at the Stephen M. Ross School of Business, University of Michigan. The authors would like to thank David Jurgens, Mengxia Zhang, Eric Schwartz, Zhenling Jiang, Hortense Fong, Jun Li, Yiqi Li, Yu Song, the Marketing faculty and doctoral students at the Ross School of Business, seminar participants at Ivey Business School, University of Wisconsin-Madison, Singapore Management University, Bocconi University, National University of Singapore, University of Manitoba, Bank of Canada, Bass FORMS Conference 2021, AIM Conference 2021, ISMS Marketing Science Conference 2021, Joint Statistical Meeting 2021, KDD 2021, ISMS Doctoral Consortium 2022, MSI Webinar 2022, Workshop by Global Institute for AI and Business Analytics 2023, JSM 2023 and Symposium on AI in Marketing 2024 for their valuable comments and feedback.

## Abstract

Influencer marketing has become a widely used strategy for reaching customers. Despite growing interest among influencers and brand partners in predicting engagement with influencer videos, there has been little research on the relative importance of different video data modalities in predicting engagement. We analyze unstructured data from long-form YouTube influencer videos—spanning text, audio, and video images—using an interpretable deep learning framework that leverages model attention to video elements. This framework enables strong out-of-sample prediction, followed by ex-post interpretation using a novel approach that prunes spurious associations. Our prediction-based results reveal that "what is said" through words (text) is more important than "how it is said" through imagery (video images) or acoustics (audio) in predicting video engagement. Interpretation-based findings show that during the critical onset period of a video (first 30 seconds), auditory stimuli (e.g., brand mentions and music) are associated with sentiment expressed in verbal engagement (comments), while visual stimuli (e.g., video images of humans and packaged goods) are linked with sentiment expressed through non-verbal engagement (the thumbs-up/down ratio). We validate our approach through multiple methods, connect our findings to relevant theory, and discuss implications for influencers, brands and agencies.

**Keywords**: *Influencer Marketing, Social Media Engagement, Interpretable Deep Learning, Video Analysis, Model Attention*

## 1. Introduction

Influencers have the capacity to shape the opinions of others within their social network. The vast majority of influencers today are individuals who have cultivated a loyal following over time by producing professional-quality videos that signal authority and credibility (Digital Marketing Institute, 2024). Their growing popularity has contributed to the rapid expansion of the influencer marketing industry which is expected to reach a global valuation of $22.2B at the end of 2025 from $9.7B in 2020 (Statista, 2023). In this ecosystem, influencer marketing agencies play a central role, often charging brands monthly retainer fees up to $18,000 for a range of services, including influencer sourcing, strategy development, content creation, and performance tracking (Fields, 2025). For both influencers and agencies, driving engagement with both organic and sponsored content is essential, as higher engagement levels enhance visibility and increase opportunities for brand endorsements (Alain, 2023). Furthermore, for brands, engagement with sponsored content is a critical performance metric used to assess campaign success (Influencer Marketing Hub, 2022). Consequently, the ability to predict video engagement before launch is of strategic importance to agencies, enabling them to serve as reliable brokers between influencers and sponsoring brands (Quuu Blog, 2024).

Despite the importance of predicting engagement in influencer videos, prior research reveals three key gaps. First, past literature has not examined the relative importance of different video modalities—namely text, audio, and video images—in predicting engagement with social media videos (details in Section 2.1). Second, while several studies have explored the relationship between specific features within each modality and online engagement, the evidence remains mixed across platforms and contexts (details in Section 6.1). This inconsistency hinders the generalizability of prior findings and underscores the need to re-evaluate these relationships within the platform and context of interest. Third, most studies treat engagement measures — such as likes, comments and shares —as interchangeable. As a result, differences between these engagement types have received limited attention, likely due to their high correlations (Hartmann et al., 2021; Hughes et al., 2019; Lee et al., 2018). As a result, we lack a nuanced understanding of how video features may be differentially associated with distinct engagement outcomes.

In this paper, we address these gaps by developing an "interpretable deep leaning framework" and applying it to publicly available influencer videos on YouTube, one of the most widely used platforms for long-form influencer content (Shaikh, 2024). We examine multiple engagement outcomes, distinguishing between the level of engagement and the sentiment of engagement. Each of these is further categorized into non-verbal and verbal measures of engagement (details in Section 4.2). These measures are conditional on video views and, as such, are not highly correlated with one another, reflecting distinct underlying constructs.

There are several challenges in implementing our framework. First, past approaches in the marketing literature using deep learning methods have documented a tradeoff between predictive ability and interpretability (Dzyabura et al., 2023; Liu et al., 2020; Liu et al., 2019). Specifically, deep learning models that rely on ex-ante *handcrafted features* as input tend to offer high interpretability, as the relationships between input features and outcomes are transparent and easier to explain. However, this approach can suffer from limited predictive ability because handcrafted features fail to fully capture the rich and complex latent constructs embedded within each modality. In contrast, deep learning models that ingest *raw unstructured data*—such as text, audio, or video images—are better at uncovering these latent constructs, leading to stronger out-of-sample predictive performance. Yet, this predictive strength comes at the cost of interpretability, as the internal representations and decision logic of these models are often opaque, making it difficult to understand how specific inputs influence outcomes.

Second, the analysis of unstructured data is computationally very demanding (in terms of both resources and time) to get good prediction performance, requiring large datasets to prevent overfitting. Hence, any proposed framework needs to carry out analyses using feasible resources on an adequately sized dataset in a reasonable amount of time. Third, an interpretation approach to understand the role of video features must overcome the costly challenge of experimentally examining all possible combinations among the triad of video features, their location in the video, and marketing outcomes of interest. Fourth, while the typical approach in the machine learning literature for validating model reasoning involves human judges (e.g., validating sentiment analysis results based on human interpretation of word meanings), this approach is not feasible when dealing with marketing outcomes such as engagement, views, or sales, where no ground truth links these outcomes to specific video features.

We address the aforementioned challenges as follows. Our framework uses *raw unstructured data* across different modalities as input to individual deep learning models designed for each modality. Predictions from each model are combined with structured features using machine learning methods to capture the overall effect, thus achieving strong predictive performance. We contrast this with a multimodal approach that takes embeddings from unstructured data as input and captures complex interactions between modalities during the training process. We examine and validate the relative importance of modalities using both approaches. This is followed by a novel ex-post interpretation approach that helps explain the relationships captured by the individual models between video features and engagement measures.

To tackle the computational demands, we employ transfer learning on each modality of data (text, audio and video images) (Liu et al., 2020; Yang et al., 2025) that benefits from using models pretrained on millions of observations. Hence, when fine-tuning these models on our data, we are able to work with a moderate sized sample of 1620 videos, using reasonable computational resources (48GB GPU and 128GB RAM) employed for a feasible amount of time (less than 700 hours) (details in Section 7.1). Transfer learning also prevents overfitting and enhances interpretation by leveraging insights from the millions of observations in the pre-trained data.

We address the challenges posed by the burden of examining numerous potential relationships and the impracticality of using human judges to validate model reasoning, through two steps of our novel approach to interpret the prediction results of our deep learning models. Our interpretation approach adapts the theoretical framework of (visual) attention capture and transfer from the print advertising literature (Pieters & Wedel, 2004) to our context where we capture *model* attention (not *visual* attention) in each modality. In the first step, we find significant correlations between features of interest (to researchers and influencers) and the attention measures attributed by the models to these features. In the second step, we find significant correlations between the features and engagement measures predicted by our model. Relationships that are only correlated in only one of the steps but not both steps are spurious (Draelos & Carin, 2020; Vashishth et al., 2019). We examine 108 possible hypotheses and find 53 spurious associations (correlated in only one of the steps), 34 null effects, and shortlist 21 robust relationships (correlated in both steps) for future causal testing. Using simulated data, we validate the efficacy of our interpretation approach and demonstrate its superiority over commonly used benchmark feature selection methods.

Our prediction results demonstrate that unstructured data in text (captions/transcript) capture more variation in all the engagement measures than unstructured data in video images or audio. This shows that "what is said" in words is more influential than "how it is said" in imagery or acoustics for predicting video engagement. This is consistent with findings in the advertising literature that emphasize the greater importance of message content over stylistic elements in enhancing ad effectiveness (MacInnis & Jaworski, 1989; Tellis, 2003). We also find that video data in the beginning explains more variation in engagement than data in the middle or end, consistent with the fact that many viewers do not finish YouTube videos (Bump, 2021).

Our interpretation approach identifies more robust associations for features in the beginning of the video compared to the middle or end, likely due to the higher salience of stimuli at the video's onset, consistent with our prediction results. Some of our key findings for the beginning 30 seconds are as follows. First, we find that mentioning brand names, in captions/transcript is associated with a *decrease* in the sentiment of verbal engagement but not sentiment of non-verbal engagement. Second, we find that an increase in duration of music is associated with an *increase* in the sentiment of verbal engagement but not sentiment of non-verbal engagement. Third, we find that an increase in size of human images (packaged goods) displayed in video images is associated with an *increase* (*decrease*) in the sentiment of non-verbal engagement but not sentiment of verbal engagement.

During the critical initial period, the text and audio features we study are associated with sentiment of verbal engagement, whereas the video image features we study are associated with sentiment of non-verbal engagement. This suggests that viewers tend to express their sentiment toward *auditory* stimuli (e.g., brand mentions and music) through verbal engagement in comments. In contrast, sentiment toward *visual* stimuli (e.g., images of humans and packaged goods) is more often expressed through non-verbal engagement by clicking thumbs-up or thumbs-down. This aligns with prior research in sensory marketing, which shows that different sensory modalities elicit distinct emotional and cognitive responses (Krishna, 2012), potentially leading to different forms of engagement behavior. Our findings offer novel hypotheses on the role of auditory and visual stimuli in videos that future research can formally test through causal experimentation.

In summary, this paper makes three main contributions to the marketing literature. First, to the best of our knowledge, it is the first paper that rigorously documents the relative importance of video modalities (text, audio and video images) in influencer videos in predicting engagement on social media. Second, our ex-post interpretation approach prunes spurious associations and uncovers robust novel associations between video features and engagement measures for long-form videos on YouTube, an important yet understudied platform. Third, our paper draws a distinction between features associated with verbal and non-verbal social media engagement measures, a distinction not pinpointed in prior research.

Our findings and methodological approach are relevant to multiple audiences. For academics, we provide a theory-based understanding of the association between influencer video modalities/features and distinct forms of engagement on YouTube. For YouTube influencers and brand agencies, our results offer a foundation for identifying video modalities and their features to manipulate in A/B testing to assess changes in engagement outcomes. For researchers and marketing practitioners across domains such as advertising, education and politics, we offer a methodological framework that can be used to analyze any *local* video feature of interest, while pruning spurious associations to identify promising relationships for formal causal testing.

## 2. Related Literature

In this section, we review the literature on influencer marketing and unstructured data analysis (using deep learning) and describe how our work builds on it.

### 2.1 Influencer Marketing

The growing body of literature on influencer marketing has examined its impact across text, audio and video data. Research on textual data reveals that high influencer expertise increases engagement when the advertising intent is to raise awareness (Hughes et al., 2019). Likewise, Weibo posts with more brand mentions are associated with an increase in informativeness, leading to more reposts (Leung et al., 2022).

Recent studies have focused on audio and video data in influencer marketing. Findings suggest softer voice tones in sponsored Instagram videos increase positive sentiment (Hwang et al., 2022), while sponsored YouTube videos can lead to a loss in subscribers (Cheng & Zhang, 2024). Similarly early brand disclosure in videos on Bilibili can reduce engagement (Chen et al., 2022). On TikTok, products advertised in engaging video segments lead to higher sales, and an increase in follower counts increases impressions (Tian et al., 2023; Yang et al., 2025). However, simultaneously employing large and small influencers has been shown to decrease sales (Gu et al., 2024). Similarly, sponsored streams on Twitch have been found unprofitable for most game publishers (Huang & Morozov, 2025). On audio platforms, unknown music creators can increase their follower base by seeding creators with less followers than established influencers (Lanz et al., 2019), and a novel framework has been developed to increase profit potential by buying future endorsements from prospective influencers (Lanz et al., 2024).

We contribute to this stream of literature by examining the relative importance of video modalities—text, audio and video images—in predicting social media engagement. While most of the past literature considers engagement as a linear combination of shares, comments and likes, our work extends this by developing distinct measures of the level and sentiment of engagement, which are further sub-divided into verbal and non-verbal engagement. This creates four unique engagement constructs whose differential association with video features has not been studied in prior work.

### 2.2 Unstructured Data Analysis in Marketing via Deep Learning

Deep learning's application in marketing literature has risen due to its capacity to capture complex, non-linear relationships which help improve predictions. Studies on textual data have employed Convolutional Neural Nets (CNNs), Long Short-Term Memory Cells (LSTMs) and Transformer-based architectures to study various outcomes including sales conversion (Liu et al., 2019), sentiment in reviews (Chakraborty et al., 2022), company survival (Zhang & Luo, 2023) and consumer perception of service (Puranam et al., 2021).

Image data have been analyzed using CNN-based architectures to classify and label images (Hartmann et al., 2021; He et al., 2023; Overgoor et al., 2022; Zhang et al., 2021) and to predict brand personality and product return rates (Dzyabura et al., 2023; Liu et al., 2020). Audio data have been analyzed use theory-based CNN architectures to predict emotional response to music (Fong et al., 2025). Video data analysis has also utilized deep learning, extracting engineered features to study their relationship with various outcomes, such as product preference while shopping (Lu et al., 2016), project success on Kickstarter (Li et al., 2019), consumer sentiment on Instagram (Hwang et al., 2022), video completion of educational courseware (Zhou et al., 2021) and number of subscribers on YouTube (Cheng & Zhang, 2024). Transfer learning has also been applied to identify engaging video segments on TikTok (Yang et al., 2025). There is also research embedding multimodal data to predict business outcomes (Lee et al., 2024), to suggest logo features for a new brand (Dew et al., 2022), to suggest new designs (Burnap et al., 2023) and to use as a control variable (Tian et al., 2023).

We contribute to this stream of literature by developing an interpretable deep learning framework that not only predicts well but also interprets model reasoning through attention mechanisms. Our novel interpretation approach is grounded in model attention theory from deep learning literature unlike past studies that adapted causal inference methods from econometrics to analyze unstructured data.

## 3. Interpretation Approach: Frameworks and Theory

### 3.1 Interpretable Deep Learning Framework

Our interpretable deep learning framework, depicted in Figure 1, uses unstructured video data as input to individual deep learning models to predict engagement. To avoid the computational demands associated with training millions of videos, we use a transfer learning approach. Our base models for each data type (text, audio and images) have been pre-trained in prior research on millions of observations at a high computational cost. By virtue of being pre-trained, these models have already learned basic patterns in unstructured data, making it easier for the model to identify patterns in our data sample. We customize these models with additional architectures, including attention mechanisms, that enable interpretation of inner workings of the models. Finetuning these models on our sample allows us to capture the relationship between our unstructured video data and our engagement measures.

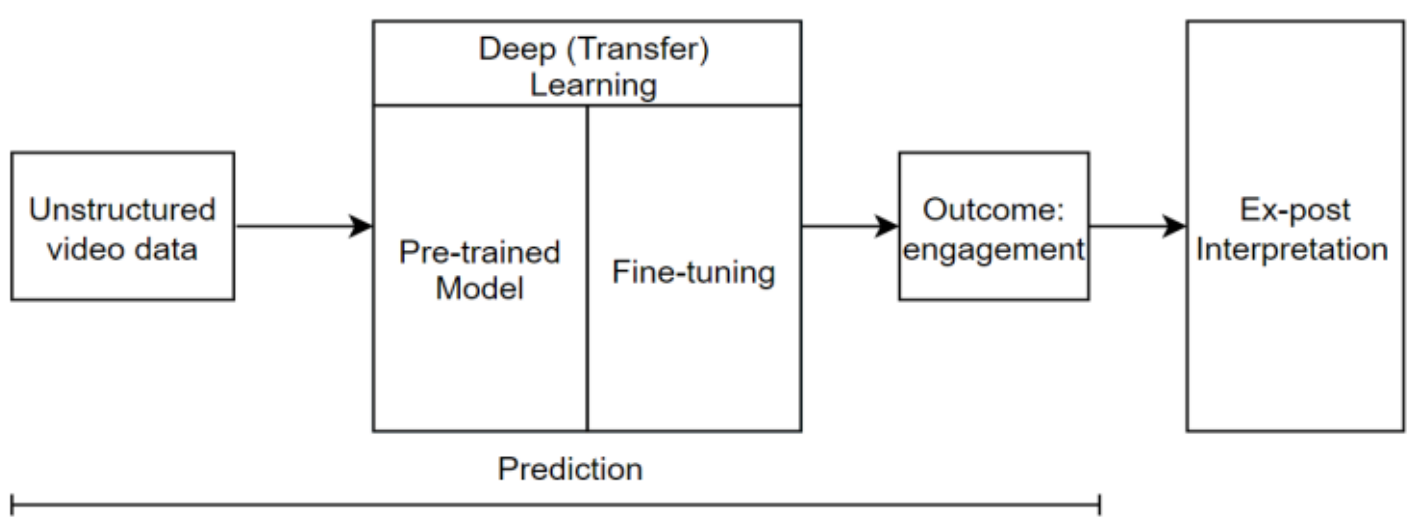

*Figure 1: Interpretable Deep Learning*

Using raw unstructured video data as input, rather than handcrafted features, enables our models to achieve strong predictive performance consistent with the approach in prior literature (Dzyabura et al., 2023; Liu et al., 2020; Liu et al., 2019). After predictions are made, we implement ex-post interpretation by analyzing the trained models to uncover the captured relationships. We accomplish this by engineering features ex-post from the unstructured data which were supplied as input to the deep learning models and then study the *attention* (importance) attributed by the model to these features (stimuli) while originally predicting engagement. We explain the theory behind interpretation in 3.2.

### 3.2 Ex-Post Interpretation Framework

We propose a framework adapted from the visual attention theory in print advertising by Pieters and Wedel (2004), tailored to the context of influencer videos. On entertainment platforms like YouTube and TikTok, bottom-up (saliency-based) attention is more common than top-down (volitional) attention (Yang et al., 2025). This type of attention where salient features in text, audio or video images capture viewer focus is particularly relevant for our setting.

The top of Figure 2 illustrates the relevant part of the framework of Pieters and Wedel (2004), while the bottom of the figure shows our adaptation. In print ads, a stimulus such as the size of a brand logo affects the reach of the ad through the mediating effect of visual attention to the stimulus. In our context, the stimuli comprise of features that can be *locally* identified within unstructured data (presence of text features, duration of audio features or size of image features) (Box I) so that we can measure attention attributed to those features. We capture *model attention (or importance)* to these features (Box II) and not viewer visual attention. These attention measures are more salient for those features that explain more variation in engagement (Box III)[1]

---

[1] The framework in Pieters and Wedel (2004) uses eye-tracking studies to measure visual attention. We complement this stream of work by analyzing secondary data and introduce the concept of *model attention* from the machine learning literature to the business literature. Our

(Selvaraju et al., 2017; Vashishth et al., 2019). The effect of a feature on engagement is thus mediated by model attention to the feature, making it a theoretically necessary condition for the feature to impact predicted engagement. This parallels the role of visual attention in print ads where attention to brand features is necessary for effective brand communication (Pieters & Wedel, 2004).

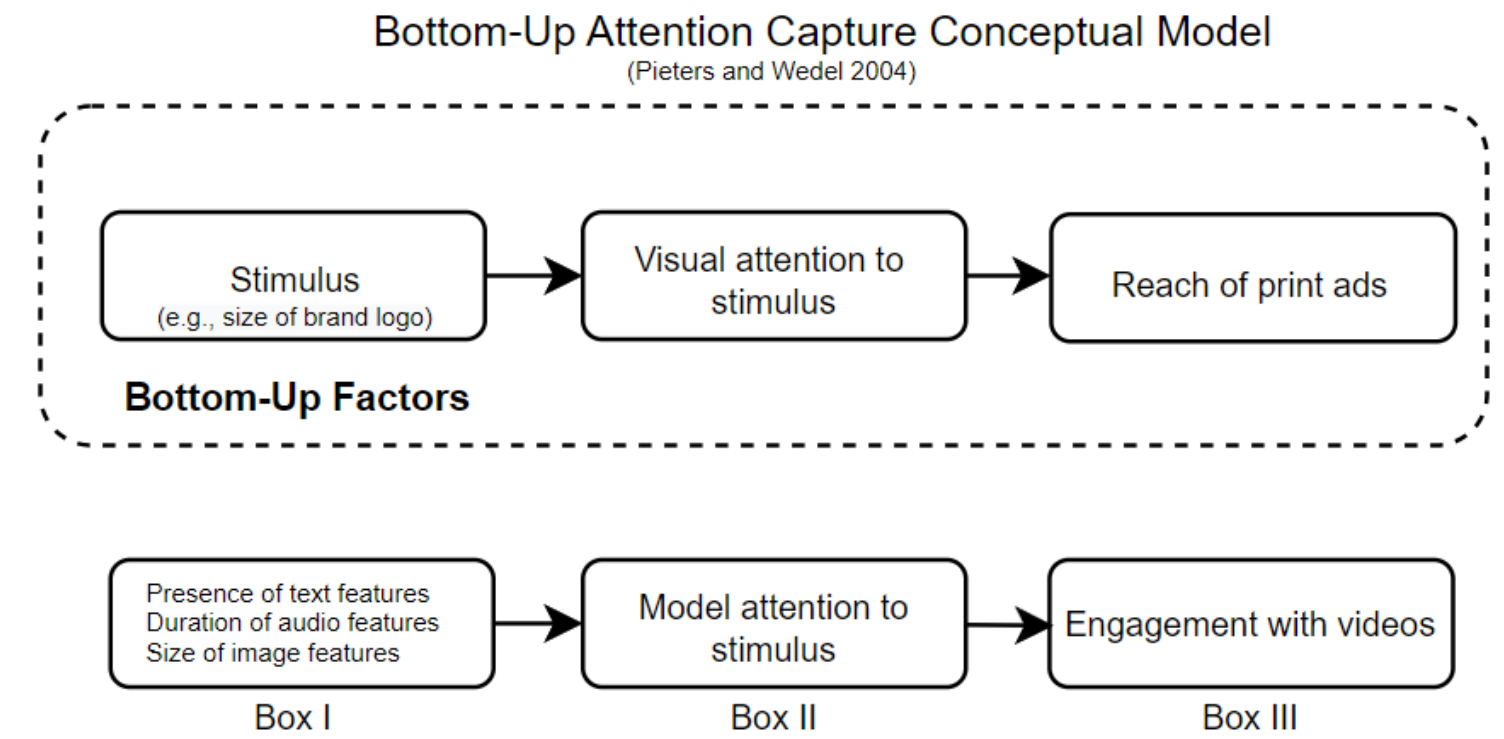

*Figure 2: Ex-Post Interpretation Framework*

However, an increase in model attention alone does not necessarily imply a significant effect on engagement. Past literature has demonstrated that deep learning models may assign attention to features spuriously (Jain & Wallace, 2019), and has identified conditions under which attention weights are interpretable (Wiegreffe & Pinter, 2019). Subsequent work has conducted systematic analyses showing that attention weights can provide reliable explanations when appropriately applied, but also emphasizes the need to exclude instances where these explanations are spurious (Vashishth et al., 2019). We detail the theory on types of spurious associations in Online Appendix A. While prior literature has often relied on human judgment to exclude spurious associations, this approach is not feasible in our context, as there is no ground truth linking video features to our engagement measures.

---

approach can be applied on public videos that potentially engage millions of viewers over many minutes of viewing, as opposed to the small number of viewers in conventional lab or field studies. It is also important to note that model attention is not independent of eye tracking attention, and extant research in machine learning has found that they have a statistically significant correlation (Selvaraju et al., 2017).

## 3.3 Two-Step Approach

To address spurious associations in deep learning models, we employ a two-step approach (see Figure 3).

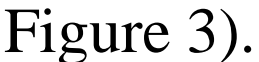

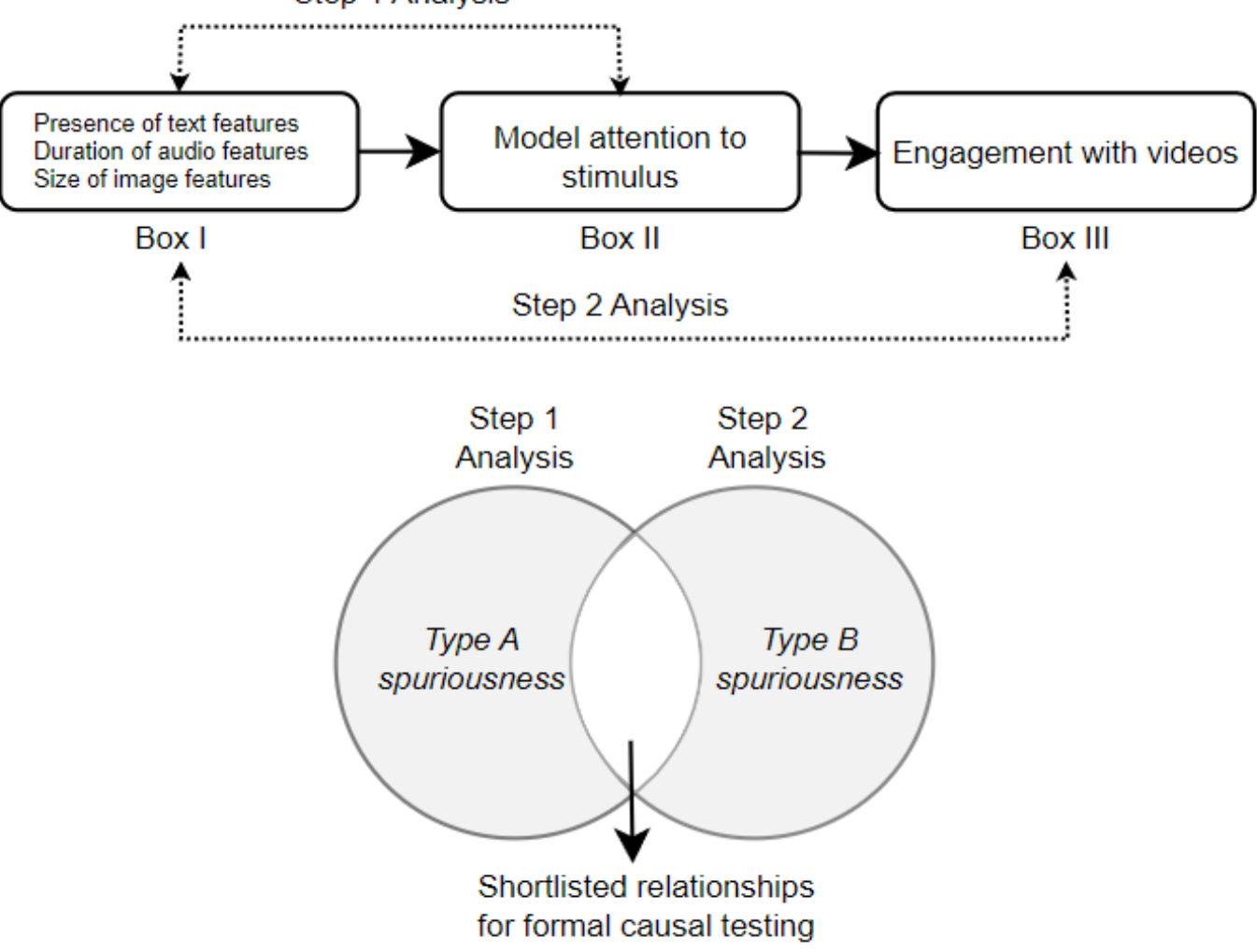

*Figure 3: Two-step approach*

In the first step, we assess whether attention is attributed to a feature by analyzing the correlation between the feature (Box I, Fig. 3) and the attention it receives (Box II, Fig. 3) during model training. This helps determine if features of interest are important in predicting engagement. However, spurious attention can occur if a feature is not genuinely correlated with predicted engagement, leading to Type A spuriousness. In order to identify and remove such spurious associations, we implement Step 2 of the analysis where we analyze the correlation between features (Box I, Fig. 3) and the *predicted* engagement measures returned by the models (Box III, Fig 3). We use predicted (estimated) engagement (and not observed engagement) as the dependent variable since it is theoretically influenced by the attention measures, making the Step 2 analysis directly comparable with the Step 1 analysis that uses (estimated) attention measures as the dependent variable. Now, some of the features can also be spuriously correlated with predicted engagement without the necessary mediating effect of an increase in attention, resulting in Type B spuriousness. Hence, we find relationships at the intersection of Step 1 & 2 which is visually illustrated using a Venn diagram at the bottom-half of Figure 3.

We exclude Type A spurious features from the left and Type B spurious features from the right of the Venn diagram, identifying features (Box I) that cause both a change in attention (Box II) and predicted engagement (Box III) (details in Online Appendix A). Analogously, our two-step approach can be understood through the lens of visual attention in print advertising where, at a minimum, the feature (stimulus) must be correlated with both the mediator (visual attention to the stimulus) and the outcome (reach of print ads), as a necessary condition for establishing causality.

To address the "brittleness" of deep learning models, which may converge to local optima, we implement a form of bootstrapping, by running each model multiple times and finding the average across iterations (details in Section 7.1). This ensures that we do not erroneously discard a causal association. While the two-step approach removes spurious associations, it is not a definitive method for identifying *only* causal features. Some relationships at the intersection may still be *coincidentally* correlated across both steps and remain confounded due to correlation with unobservables. Nonetheless, by pruning spurious associations, we reduce the effort required for future causal research. The number of spurious associations removed will increase as more local features are tested in Box I or more engagement measures are evaluated in Box III.

## 4. Data

We focus on influencer videos on YouTube, one of the most popular platforms for long-form content (Shaikh, 2024). To create our sample, we first shortlist 110 influencers identified by Forbes in February 2017 (O'Connor, 2017) across 11 product categories (Beauty, Entertainment, Fashion, Fitness, Food, Gaming, Home, Parenting, Pets, Tech & Business, Travel) with 10 influencers in each category. These influencers[2] are top performers, earning revenue from brand endorsements and primarily posting in English across Facebook, YouTube, Instagram, and Twitter. We refine our sample by excluding influencers not active on YouTube and those with fewer than 1,000 followers or more than 100 million followers. Additionally, we focus on influencers who post at least 50 videos to ensure sufficient variation in their content, narrowing our pool to 73 influencers.

We thus have a sample of YouTube influencers who primarily create content for adults, post only pre-recorded videos, and have between 1000 and 100M subscribers reflecting the typical range of YouTube subscribership. From this group, we randomly select 3 influencers per category, totaling 33 influencers who can be categorized by their subscriber count: 1000 to 10K (2), 10K to 100K (6), 100K to 1M (7), 1M to 10M (13), and 10M+ (5). While these influencers are on average top performers across social media platforms, many have only a small to mid-sized follower base on YouTube. Using the YouTube Data API v3, we scrape the titles and posting times of all videos posted by these 33 influencers until October 2019, resulting in a master list of 32,246 videos.[3] We then randomly select 50 public videos per influencer, creating a balanced sample of 1650 videos whose unstructured data (text, audio and images) is computationally feasible to analyze. After excluding videos with disabled likes, dislikes or comments, we are left with 1620 videos, from which we scrape all publicly available data in November 2019. Since we use transfer learning methods that benefit from using models pretrained on millions of observations (details in Section 5.1), we are able to work with a moderate sized sample of 1620 videos, thus allowing us to use feasible computational resources employed for a reasonable amount of time (details in Section 7.1).

[2] The criteria used by Forbes to identify these influencers include total reach, propensity for virality and level of engagement (across all social media channels) including endorsements and related offline business.

[3] Our usage of this data falls within the ambit of YouTube's fair use policy .

We assess whether the influencers in our sample comply with the U.S. Federal Trade Commission (FTC) guidelines, which mandate that sponsorships be disclosed early in a video using terms like 'ad' or 'sponsor' (FTC, 2020). By examining the captions/transcript in the beginning (and middle) of each video, we find that only 1% of videos include such disclosures. For instance, despite reports in the media indicating a partnership between a specific parenting influencer and a brand, related videos lack sponsorship disclosures. This overall low compliance suggests that our sample contains both paid and organic videos, often without indicators of whether a video is paid, consistent with findings in related research (Ershov et al., 2025).

### 4.1 Unstructured Data

We list the unstructured data which we supply as input to our individual deep learning models (details in Section 5.1) in Table 1. Text data comprise the title of the video, description below the video and captions/transcript of the video. For the video description, a maximum of 160 characters are visible in Google Search and even fewer characters are visible below the video before the 'Show More' link (Cournoyer, 2014). Hence, we truncate each description to the first 160 characters (160 c) as it is more likely to contribute to any potential association with our outcomes. Captions are present in 74% of the videos in our sample; for videos without captions, we use Google's Cloud Speech-to-Text API to transcribe the audio to English. We also utilize the raw audio data alongside captions/transcript to capture acoustic information (e.g., music). Image data comprise thumbnails and image frames captured at the standard sampling rate of one frame per second (fps) (Yang et al., 2025; Yue-Hei Ng et al., 2015). These frames have a resolution of 135x240 pixels, which is both visually clear and feasible for analysis.

For the unstructured data in captions/transcript, audio and video frames, we divide the data into segments of 30 seconds (e.g., for a 10 minute video, we have 20 segments) for a few reasons. First, the minimum duration of video content that needs to be viewed for an impression to be registered is 30 seconds (Parsons, 2017) which makes this an adequate threshold. Second, to manage computational costs, we limit data analysis to 30-second segments, which is feasible using a 48GB GPU with 128GB RAM. Third, we aim to study whether effects differ across various parts of the video, considering many viewers do not watch YouTube videos in their entirety (Bump, 2021).

| **Class** | **Features** |
|---|---|
| Text | Title |
| | Description (first 160 characters) |
| | Captions or Transcript (divided into 30 sec segments) |
| Audio | Audio file (divided into 30 sec segments) |
| Images | Thumbnail |
| | Video Frames (divided into 30 sec segments) at one fps |

*Table 1: Unstructured data*

## 4.2 Outcome Variables

In the industry, the most important criteria used by brands to evaluate influencers to partner with are engagement/clicks, followed by content type/category, impressions and sales (Influencer Marketing Hub, 2022). In making these decisions, brands generally consider overall engagement—rather than differentiating between paid and organic posts—since it better reflects an influencer's capacity to generate high-performing content (Alain, 2023). Consequently, influencers are motivated to build engagement across all their videos to increase their chances of securing brand sponsorships. Given that our sample includes both paid and organic videos (with incomplete indicators of paid content as discussed earlier), focusing on engagement across both types of videos is appropriate for our analysis.

We construct verbal and non-verbal engagement measures. Verbal engagement measures involve typing comments below a YouTube video. They get posted with your profile name below the video resulting in potential social consequences (Dwoskin, 2021). Non-verbal engagement can be completed with a simple click on a 'thumbs up' icon (like) or 'thumbs down' icon (dislike). They have no social consequences as they are anonymous on YouTube, and do not push the post onto the user's YouTube timeline (Dwoskin, 2021).

Traditional metrics often do not distinguish between verbal and non-verbal engagement, likely due to their high correlation (Hartmann et al., 2021; Hughes et al., 2019; Lee et al., 2018).[4] Additionally, YouTube's algorithm, which recommends videos based on expected watch time, can confound these engagement measures (Covington et al., 2016). To address these issues, we construct engagement measures that control for views, resulting in unique constructs (see Section 4.2.3) that are not highly correlated with each other or with views, minimizing the algorithm's influence.

#### 4.2.1 Engagement Level

We develop the following measure for non-verbal engagement level (NVL): (#likes + #dislikes) / (#views). It measures the total clicks on 'thumbs up' or 'thumbs down' conditional on video viewing. We measure verbal engagement level (VL) as follows: (#comments) / (#views), which measures the total comments conditional on video viewing. As both measures are exponentially distributed, they are transformed using their natural log, with 1 added to avoid log (0). log NVL = log ((#likes + #dislikes + 1) / #views) ranging from $-8.41$ to $-1.97$ and has a median of $-3.78$ (or approximately 228 likes and dislikes per 10K views).[5] log VL = log ((#comments + 1) /

[4] For example, in our sample of 1620 videos there is a high correlation between log views and log (comments+1) at 0.91, between log views and log (likes+1) at 0.95 and between log views and log (dislikes+1) at 0.92.

[5] Note that from Nov 2021 YouTube has made thumbs down (dislike) counts private (https://blog.youtube/news-and-events/update-to-youtube/).

views), ranging from −11.42 to −2.14 and has a median of – 6.21 (or approximately 20 comments per 10K views).

### 4.2.2 Sentiment of Engagement

Past research shows that the sentiment expressed in visual and verbal components of advertising can influence attitude towards the ad which in turn affects brand attitude including purchasing behavior (Mitchell, 1986). Moreover, the sentiment of user generated content on brand managed social media communities has been linked with sales (Goh et al., 2013). Therefore, understanding viewer sentiment towards videos can serve as a proxy for sales.

We develop the following measure for non-verbal sentiment (NVS): (#likes) / (#dislikes), which measures the likelihood of a viewer liking rather than disliking a video, with higher values indicating more positive sentiment. This measure is also transformed using log NVS = log ((#likes + 1) / (#dislikes + 1)), ranging from −0.92 to 6.83, and has a median of 3.99 (or 54 likes per dislike).

We capture verbal sentiment (VS) by analyzing the sentiment in comments using Google's Natural Language API (Li & Xie, 2020). Specifically, we capture the average sentiment of the top 25 comments below a video, with sentiment scores ranging from -1 (very negative) to +1 (very positive). [6] For comments made in a language not supported by the Google Natural Language API, we use the Google Translation API to translate them into English before analyzing sentiment. The distribution of average sentiment scores ranges from −0.90 to 0.90, and we use the median value (0.34) to categorize loveability into "positive" and "not positive" (neutral or negative). This binary approach helps create a more objective measure of sentiment since there can be minor variations in continuous measures based on the choice of the API (or algorithm/human coder). Using a categorical outcome for sentiment is consistent with the approach in previous research (Goh et al., 2013; Li & Xie, 2020). We assume that if viewers do not comment on a video, and comments are not disabled, the sentiment is neutral. As a robustness check, we scrape the top 50 or 100 comments from a random sample of 66 videos (2 videos per influencer) and explore use of decreasing weights. We find that sentiment from these measures is highly correlated with a simple average of the top 25 comments ($\rho \geq 0.88$).

[6] A tabulation shows that 99% of comments in our data are made by viewers and not the influencer and hence we do not separate the two.

#### 4.2.3 Summary of outcome variables

We have three continuous outcomes and one binary outcome, as summarized in Table 2a. The average magnitude of the Pearson correlation coefficient among the four outcomes is 0.28, indicating that these measures are not highly correlated with each other. A Principal Component Analysis (PCA) on these outcomes (including views) reveals that each measure loads heavily (≥0.83) on a distinct construct (see Table 2b), providing further evidence that each outcome variable captures unique information: {verbal or non-verbal} x {level or sentiment}.

| | Verbal Engagement (V) | Non-Verbal Engagement (NV) |
|---|---|---|
| Level of Engagement (L) | VL: $\frac{\#comments}{\#views}$ | NVL: $\frac{\#likes+\#dislikes}{\#views}$ |
| Sentiment of Engagement (S) | VS:<br>Positive or Not Positive | NVS: $\frac{\#likes}{\#dislikes}$ |

*Table 2a: Summary of outcome variables*

| | Factor 1 | Factor 2 | Factor 3 | Factor 4 | Factor 5 |
|---|---|---|---|---|---|
| Log NVL | 0.45 | 0.32 | 0.07 | 0.02 | 0.83 |
| Log VL | 0.96 | 0.04 | 0.00 | -0.09 | 0.28 |
| Log NVS | 0.05 | 0.94 | 0.24 | 0.10 | 0.23 |
| VS | -0.07 | 0.08 | -0.11 | 0.99 | 0.01 |
| Log views | 0.01 | 0.21 | 0.97 | -0.12 | 0.05 |

*Table 2b: Factor Loadings from a PCA of variables*

### 4.3 Structured Features

We also list the set of structured features that we supply as input to our Combined Model (Section 5.2) and Multimodal Model (Section 5.3) in Table 3.

| Fixed Effects | Influencer Fixed Effects (33) |
|---|---|
| Length | Video Length (min) |
| Tags | Number of video tags assigned by influencer (see Google (2020) for details) |
| Playlist Information | Number of playlists the video is a part of |
| | Average position in playlist |
| | Average number of videos on all the playlists the video is a part of |
| Time based covariates | Time from upload time to scrape time |
| | Year of upload (2006 to 2019) |
| | Time between uploads: Given and preceding influencer video in master list |
| | Time between uploads: Given and succeeding influencer video in master list |
| | Rank of video among all videos of the influencer in the master list |
| | Day fixed effects (7) in EST |
| | Time of day fixed effects in intervals of 4 hours from 00:00 hours EST (6) |
| Captions Indicator | Indicator of whether video has closed captions |
| URLs and Hashtags | Total number of URLs in description |
| | Indicator of Hashtag in description |

*Table 3: Structured features*

These include fixed effects for the influencer (channel) that account for time invariants effects associated with the influencer. In addition, we have features for video length, number of tags, features for playlist information, time-based features and whether video captions are available. We create two additional structured features from the *complete* description (as it is not supplied as input to the deep learning model). They comprise total number of URLs in description and an indicator for hashtag in description as these elements may lead viewers away from the video.

## 5. Model

This section outlines our two main modeling approaches: (a) the Combined Model, which integrates outputs from individually trained models on *raw* data from each modality (text, audio, and images); and (b) the Multimodal Model, which jointly trains *embeddings* from each modality in an end-to-end framework. Table 4 summarizes the approaches and their trade-offs.[7] Section

[7] We are unable to train *raw* data (along with structured features) in an end-to-end framework due to the extremely high computational (and monetary) cost associated with such an undertaking which is typically accomplished by researchers at large corporations. Hence, we use two approaches to combine information from all modalities, as each approach has its distinct advantage.

5.1 details the individual models, 5.2 describes the Combined Model, and 5.3 contrasts it with the Multimodal Model.

| Model | Approach | Strengths | Limitations | Input Data |
|---|---|---|---|---|
| Combined Model | Combines predictions from individual models trained separately on each modality of *raw* unstructured data | • Fine-tunes tailor-made models for each data modality<br>• Captures within-modality interactions during training<br>• Leverages information-rich raw data | • Loss of cross-modal information due to interactions being captured only after individual model training<br>• Slower computation | • 30 second segments from the beginning, middle and end (Results in 7.1.2) |
| Multimodal Model | Trains a single model using *embeddings* from each modality of raw unstructured data | • Captures both within-modality and cross-modal interactions during training<br>• Faster computation | • Tailor-made models are only used for generating embeddings as input and are not fine-tuned.<br>• Loss of information due to reliance on embeddings and their averaging | • 30 second segments from the beginning, middle and end (Results in 7.1.3)<br>• 30 second segments across the *entire video* (Results in 8.1) |

*Table 4: Summary of Modelling Approaches*

### 5.1 Individual Models

For text data, we use Bidirectional Encoder Representation from Transformers (BERT Base), a high performing NLP model (Devlin et al., 2018). This seminal model has been widely used in the machine learning literature for single sequence tasks like classification and regression, and its use is also emerging in marketing research (Puranam et al., 2021). BERT is pre-trained on Book Corpus (800M words) and English Wikipedia (2,500M words) to predict words and sentences over four days using four Tensor Processing Units. It employs a self-attention mechanism with scaled dot-product attention as its core computation method that generates relative attention weights for each word-piece. We fine-tune BERT on our data sample for each engagement measure, and provide details on how we operationalize the model and obtain attention weights in Online Appendix B.

For audio data, we use the pre-trained YAMNet model which converts raw audio signals into Mel spectrograms (capturing acoustic characteristics) and processes them through a MobileNet v1 architecture (Pilakal & Ellis, 2020). YAMNet is well-regarded in the machine learning community for its ability to identify sound classes, as it is pre-trained on the AudioSet data which contains over two million 10-sec YouTube audio segments (Gemmeke et al., 2017). We customize YAMNet by adding a Bidirectional LSTM (Bi-LSTM) layer to capture the sequential relationship between sound moments. Additionally, we incorporate the additive attention mechanism from neural machine translation literature (Bahdanau et al., 2014) to help the Bi-LSTM layer capture relative attention weights between sound moments. We fine-tune the Audio model on our data sample for each engagement measure and provide details on how we operationalize the model and obtain attention weights in Online Appendix C.

For video images we use a combination of VGG-16 and Bi-LSTM (and for thumbnail images we only use VGG-16). VGG-16 is a high-performing, seminal image model pre-trained on 1.2M images from the ImageNet dataset (Simonyan & Zisserman, 2014) that has also gained popularity in business research (Hartmann et al., 2021; Zhang et al., 2021). To analyze video image data, each model processes a single video frame as input, and we combine the outputs using a Bi-LSTM, an architecture known for its effectiveness at capturing sequential information from video frames (cf. Yue-Hei Ng et al. (2015)). We fine-tune the Video Image Model on our data sample for each engagement measure and identify salient areas in images using Grad-CAM (Gradient weighted Class Activation Mapping) (cf. Selvaraju et al., 2017), which employs gradient-based attention. We slightly modify the approach to allow the generation of both positive and negative gradients. Positive (negative) gradients correspond to regions that are positively (negatively) associated with continuous outcomes and the predicted class of the binary outcome. Details on the operationalization of the Video Image Model and gradient generation are in Online Appendix D.

While negative gradients in the Video Image Model identify areas that are negatively associated with an outcome, negative attention weights in the Text and Audio models identify areas with no association. We avoid using approaches like additive or scaled dot-product attention in the Video Image Model to find salient areas because it would be computationally expensive over a 135x240 pixel frame for 30 frames. The gradient-based approach achieves the same objective more efficiently.

### 5.2 Combined Model

We use equation (1) to combine information from unstructured data and structured features. This is also visually shown in Figure 4. Information from unstructured data is incorporated using the predicted outcome values $\hat{Y}_{it}$ for video $t$ by influencer $i$, obtained by inputting 30-second segments from each part of the video (beginning, middle or end) from each source of unstructured data (listed earlier in Table 1) into the corresponding individual model described in Section 5.1. The structured features, $X_{it}$, (listed earlier in Table 3) are supplied as an additional input to the equation. We test different models, $g$, to combine information from all these sources of data and capture any potential interactions between them. Equation (1) is shown below,

$$Y_{it} = g\left(\begin{array}{c} X_{it}, \hat{Y}_{it_{Title}}, \hat{Y}_{it_{Description\ (first\ 160\ c)}}, \\ \hat{Y}_{it_{Captions-Transcript\ (beginning)}}, \hat{Y}_{it_{Captions-Transcript\ (middle)}}, \hat{Y}_{it_{Captions-Transcript\ (end)}}, \\ \hat{Y}_{it_{Audio\ (beginning)}}, \hat{Y}_{it_{Audio\ (middle)}}, \hat{Y}_{it_{Audio\ (end)}}, \\ \hat{Y}_{it_{Thumbnail}}, \hat{Y}_{it_{Video\ Frames\ (beginning)}}, \hat{Y}_{it_{Video\ Frames\ (middle)}}, \hat{Y}_{it_{Video\ Frames\ (end)}} \end{array}\right) + \epsilon_{it} \qquad (1)$$

where $Y_{it}$ is the observed outcome for video $t$ by influencer $i$, $g$ is the combined model used and $\epsilon_{it}$ is the error term. We test the performance of seven different combined models that are known to predict well out-of-sample in the machine learning literature. They comprise four linear

models – OLS, Ridge Regression, LASSO and Elastic Net, and three non-linear models – Deep Neural Net, Random Forests and XGBoost.

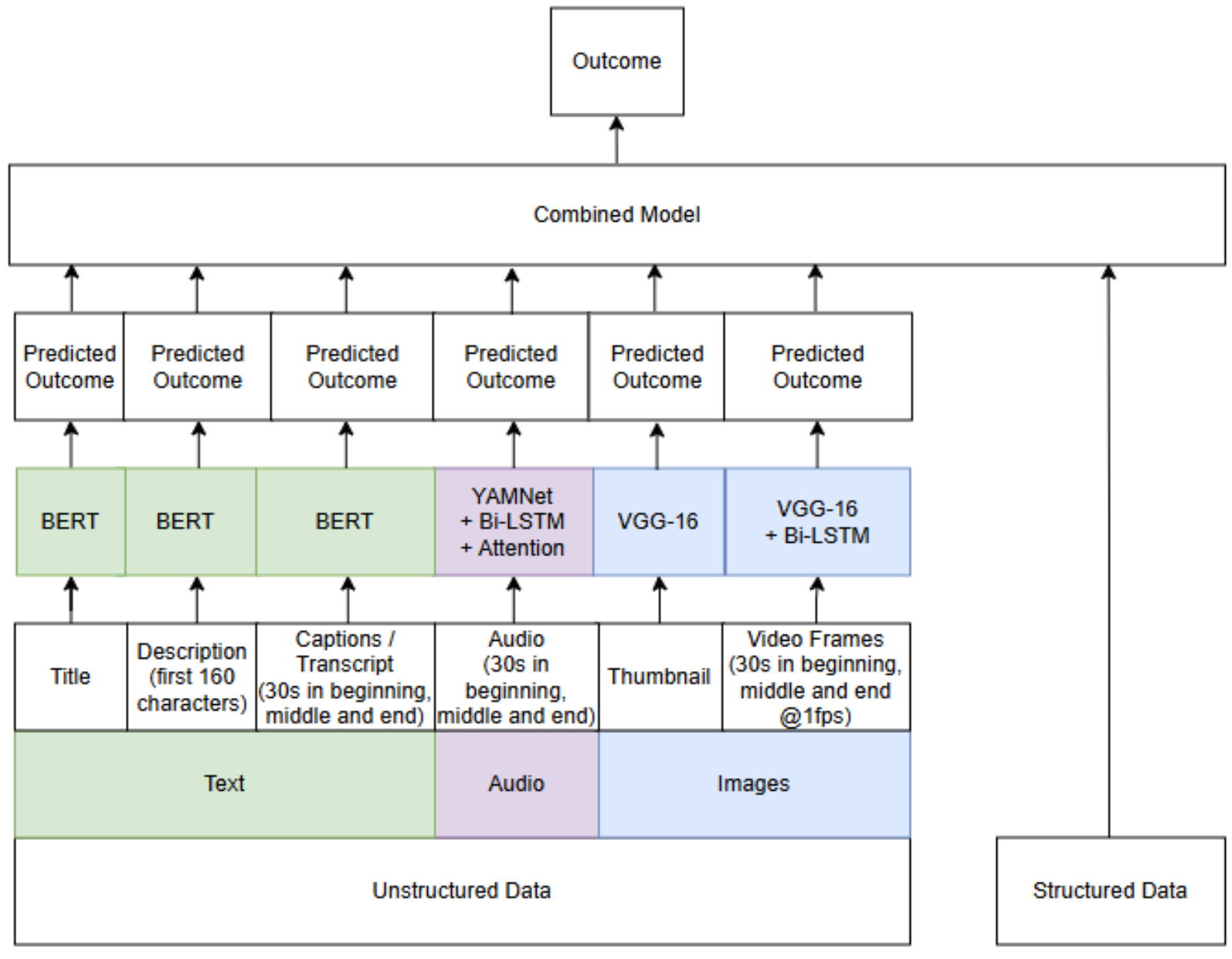

*Figure 4: Combined Model*

## 5.3 Multimodal Model

We show the architecture of our Multimodal model in Figure 5. It is based on Ghosal et al. (2018) and Wu et al. (2014), and has been adapted for our setting.

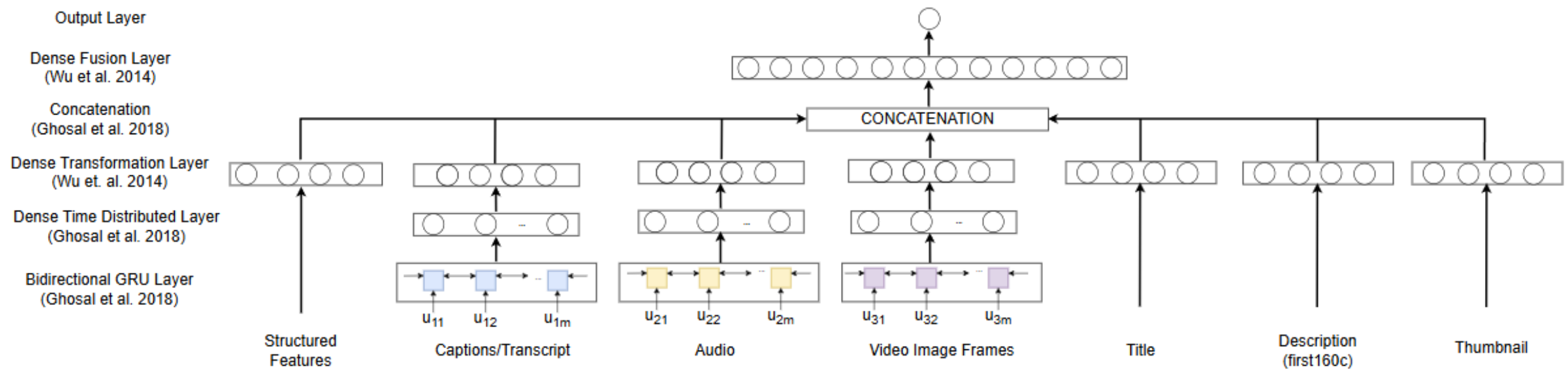

*Figure 5: Multimodal Model*

We first obtain embeddings (using the same pre-trained models in Section 5.1) for text (captions/transcript), audio and video image frames from each 30 second clip in the beginning, middle and end of a video (details in Online Appendix E.1). We also use embeddings from each 30 second clip over the entire duration of a video for validation (Section 8.1). We average the embedding over the length of the clip for each modality and pass it as input to Bi-Directional GRU Layers that capture the within-modality interaction between each clip. This is followed by a sequence of dense time-distributed layers and dense transformation layers that further process the data. In addition, we also obtain embeddings from the title, description (first 160c) and thumbnail of each video, which along with the structured features (listed earlier in Table 3), are supplied as input and processed through independent dense transformation layers. We then concatenate the output of the preceding seven layers and capture the cross-modal interactions between modalities using a dense fusion layer (see Online Appendix E.1 for details).

## 6. Interpretation Approach: Implementation

### 6.1 Theory-based Features

Having outlined our models, we turn to our process for identifying features to use within our framework (Box I in Figures 2 and 3). Now, existing literature in social media, advertising and influencers has studied the role of features on sentiment of engagement but has not differentiated between features linked to verbal versus non-verbal sentiment. As our goal is to uncover this distinction in the context of YouTube influencers, we need a systematic approach to identify the set of features to investigate from the vast pool of features in video data.

First, we focus only on features that can be identified locally within text, audio, or video images, so that they can be used within our interpretation framework (as discussed in Section 3.2). For example, a person present in an image is a locally identifiable feature, whereas overall image brightness, which affects the entire frame, is not (Dzyabura et al., 2023; Li et al., 2019; Zhang et al., 2021). Second, we consulted with executives at Google/YouTube and two prominent influencers to discuss local features of interest for these stakeholders. Third, we used theory from relevant literature including influencer marketing, social media, news media and advertising, to identify local features that have been linked with the sentiment of engagement, as it can act as a proxy for sales (as discussed in Section 4.2.2). Through this systematic approach, we identify a broad set of nine features with theory-based expectations for their impact on sentiment of engagement, summarized in Table 5.

| **Unstruct-ured data** | **Features in unstruct-ured data** | **Theory** | **Representative Source** | **Expected change in sentiment of engagement** |
|---|---|---|---|---|
| Text: Captions/ Transcript | Presence of brand names (1) | Brand mentions may lower entertainment value and decrease persuasiveness of message.<br>Brand mentions may increase informational value and increase viewers' trust in the brand. | Tellis et al. (2019); Teixeira et al. (2010)<br>Leung et al. (2022) | Ambiguous |
| | Presence of emotional words (2) | Emotional words are tied to a brand's personality, evoke high arousal and increase message persuasiveness.<br>Emotional words are associated with disingenuous persuasion or insincerity and signal fakeness. | Lee et al. (2018); Berger and Milkman (2012)<br>Bakir and McStay (2018); Guo et al. (2019) | Ambiguous |
| Audio | Duration of music (3) | Music in ad videos reduces irritation towards the ad. | Pelsmacker and Van den Bergh (1999) | Positive |
| | Rapid speech (4) | Rapid speech is associated with being more knowledgeable and an increase in watch time.<br>Change in pace of voice grabs viewer attention. | Peterson et al. (1995); Guo et al. (2014)<br>Beck (2015); Jennings (2021) | Ambiguous |
| Video Images | Size of humans (5) | Human images increase desire to socialize, engage and communicate. | Xiao and Ding (2014); Hartmann et al. (2021); To and Patrick (2021) | Positive |
| | Size of packaged goods (6) | Packshots (which focus on a product and not humans) decrease desire to socialize, engage and communicate. | Hartmann et al. (2021) | Negative |
| | Size of animals (7) | Ads featuring animals are more likeable. | Biel and Bridgwater (1990); Pelsmacker and Van den Bergh (1999) | Positive |
| | Emotional expressio-ns of joy (8) and surprise (9) | Joy and surprise in internet videos ads concentrate attention and retain viewers for longer time periods.<br>Joyous faces are associated with decrease in retweets but are not associated with likes on Twitter or Instagram.<br>Surprise exhibited by instructors in educational videos has no dominant directional effect on video watch time, as it depends on contextual factors and topics discussed | Teixeira et al. (2012)<br>Li and Xie (2020)<br>Zhou et al. (2021) | Ambiguous |

*Table 5: Theory-based features and expectations*

As shown in the table, existing literature in influencer marketing, social media, news media and advertising, provide either ambiguous or deterministic expectations for the relationship between video features and sentiment of engagement on YouTube. Furthermore, since the differential association of these features with sentiment of verbal and non-verbal engagement has not been explored in prior work, uncovering these distinctions can offer new insights into consumer behavior on YouTube. We generate the features listed in Table 5 using our models and other APIs, with details and descriptive statistics provided in Online Appendix F.

### 6.2 Interpreting Relationship with Attention (Step 1)

After estimating attention weights from the Text model for each word-piece in captions/transcript, we implement Step 1 of the interpretation approach by analyzing the following equation 12 times—across 4 outcomes and 3 video parts (beginning, middle and end):[8]

$$\log(AttentionWeight_{itk}) = \alpha_i + \gamma X_{it} + \sum_{k=1}^{n_b} \beta_{1k}(BIT_{itk}) + \sum_{k=1}^{n_e} \beta_{2k}(EIT_{itk}) + \beta_3(TP_{itk}) + \beta_4(NOT_{it}) + \epsilon_{itk} \quad (2)$$

where $AttentionWeight_{itk}$ is the estimated weight for each token $k$ (word-piece created from raw text by the model) in video $t$ made by influencer $i$. Since it is exponentially distributed, we take its log. $\alpha_i$ is influencer fixed effects and $X_{it}$ is the vector of structured features listed earlier in Table 3. $BIT_{itk}$ is 'brand indicator in token' indicating whether token $k$ is a brand name, $EIT_{itk}$ is 'emotion indicator in token' indicating whether token $k$ is an emotional word, $TP_{itk}$ controls for token position, $NOT_{it}$ controls for number of tokens in the text, $n_b$ is total number of brand names, $n_e$ is total number of emotional words, and $\epsilon_{itk}$ is the error. We use a unique coefficient for each brand and emotional word to model potential heterogeneity in effects. We use a covariate for the token position ($TP_{itk}$) to control for any potential influence of the position of the word-piece in the text, and we control for the number of tokens ($NOT_{it}$) because attention weights in text are relative to each other and sum up to one, i.e. more the number of tokens, lower will be the attention directed to it.

[8] Note that we cannot simultaneously control for presence of features in all parts of the video in the same equation in Step 1 because the number of observations (e.g., tokens $k$) in each part of the video may be different (see Tables H1 & H3 in Online Appendix H for difference in number of observations).

For audio data, after estimating attention weights for each sound moment, we implement Step 1 using a similar approach for each outcome and video part:

$$\log(AttentionWeight_{itk}) = \alpha_i + \gamma X_{it} + \beta_1(CI(Human)_{itk}) + \beta_2(CI(Music)_{itk}) + \beta_3(CI(Human)_{itk} x\, CI(Music)_{itk}) + \beta_4(CI(Animal)_{itk}) + \beta_5(CI(Other)_{itk}) + \beta_6(Location_{itk}) + \epsilon_{itk} \quad (3)$$

where $AttentionWeight_{itk}$ is the estimated weight for sound moment $k$ in video $t$ made by influencer $i$, $CI(Human)_{itk}$ is the Category Indicator for human sounds in moment $k$, and $CI(Human)_{itk}$ x $CI(Music)_{itk}$ corresponds to moments when both human and music sounds occur together, $Location_{itk}$ controls for location of the moment within the 30 second audio clip and other terms are similar to those in Equation 2. We use a covariate for location so that we can control for any potential influence of the position of the moment. Note that we do not include a covariate for number of audio moments, as each clip has the same length of 30 seconds.

For video image data, after estimating gradient values, we implement Step 1 using a similar approach for each outcome and video part:

$$MeanGradientValues_{itk} = \alpha_i + \gamma X_{it} + \sum_{k=1}^{8} \beta_{1k} SizeObject(k)_{it} + \beta_{21} Joy(Face)_{itk} + \beta_{22} Surprise(Face)_{it} + \beta_{23} Joy(Face)_{it}\ x\ SizeObject(Face)_{it} + \beta_{24} Surprise(Face)_{it}\ x\ SizeObject(Face)_{it} + \epsilon_{itk} \quad (4)$$

where $MeanGradientValues_{itk}$ is the mean gradient values across the area (pixels) occupied by all items of object category $k$ across 30 frames in video $t$ made by influencer $i$, and $k$ = 1 to 8 corresponds to each object category:{humans, faces (of humans), animals, brand logos, packaged goods, clothes & accessories, home & kitchen and other objects}. $SizeObject(k)_{it}$ is the mean across 30 frames of the percentage of the image occupied by all objects of category $k$ in video $t$ made by influencer $i$. Hence, the coefficient $\beta_{1k}$ can be interpreted as the effect of a one percent increase in size of the object of category $k$ on average across 30 seconds of the video. Note that features for size of humans and size of faces (of humans) are not highly correlated (variance inflation factor ≤ 2.5 across each video part). $Joy(Face)_{it}$ and $Surprise(Face)_{it}$ indicate the mean (across 30 frames) of the level of surprise or joy in each face {-2: very unlikely, -1: unlikely, 0: possible, 1: likely, 2: very likely}. The two interaction terms capture the interaction between the emotion registered and the size of faces, and other terms are similar to those in Equation 2.

**6.3 Interpreting Relationship with Outcome (Step 2)**

After obtaining predicted outcomes from all models, we implement Step 2 of the interpretation approach, where information from all modalities (text, audio, and video images) is combined into a single equation. This allows us to examine the association of a feature from each modality while controlling for associations of features from other modalities. We analyze the following equation 36 times—across 3 models, 4 outcomes and 3 video parts (beginning, middle and end):

$$PredictedOutcome_{it} = \alpha_i + \gamma X_{it} + \sum_{p\ in\ part}\{\sum_{k=1}^{n_b}\beta_{1pk}(BITX_{it}) + \sum_{k=1}^{n_e}\beta_{2pk}(EITX_{it}) + \beta_{3p}(NOT_{it}) + \beta_{4p}(Sum\ of\ CI(Human)_{it}) + \beta_{5p}(Sum\ of\ CI(Music)_{it}) + \beta_{6p}(Sum\ of\ CI(Human)_{it}\ x\ Sum\ of\ CI(Music)_{it}) + \beta_{7p}(Sum\ of\ CI(Animal)_{it}) + \beta_{8p}\ (Sum\ of\ CI(Other)_{it}) + \sum_{k=1}^{8}\beta_{9pk}SizeObject(k)_{it} + \beta_{10p}Joy(Face)_{it} + \beta_{11p}Surprise(Face)_{it} + \beta_{12p}(Joy(Face)_{it}\ x\ SizeObject(Face)_{it}) + \beta_{13p}(Surprise(Face)_{it}\ x\ SizeObject(Face)_{it})\} + \epsilon_{it} \quad (5)$$

where $PredictedOutcome_{it}$ is the predicted outcome from a model for a part of video $t$ made by influencer $i$, $\alpha_i$ is influencer fixed effects and $X_{it}$ is the same vector of structured features used earlier in Equation 2. $BITX_{it}$ is a 'brand indicator in text' indicating whether the text (captions/transcript) in video $t$ by influencer $i$ has a brand, $EITX_{it}$ is 'emotion indicator in text' indicating whether the text has an emotional word, and $NOT_{it}$ controls for number of tokens in text. As done in Equation 2, we use a unique coefficient for each brand and emotional word to model potential heterogeneity in effects. $Sum\ of\ CI(Human)_{it}$ corresponds to the duration of human sounds (or sum of the Category Indicator for human sounds) across the 30 seconds in video $t$ made by influencer $i$, and $Sum\ of\ CI(Human)_{it}\ x\ CI(Music)_{it}$ finds the total duration when human and music sounds overlap. As we have controlled for the number of tokens (word-pieces), an increase in $Sum\ of\ CI(Human)_{it}$ can be interpreted as slower speech whereas a decrease in $Sum\ of\ CI(Human)_{it}$ can be interpreted as rapid speech. The other variables mirror those used in Equation 4. Thus, we control for features across all sources of unstructured data (that were used in the three equations of Step 1). We also sum over the *parts* = {beginning, middle, end} in Equation 5 to control for the presence of words, sounds and objects in different parts of the video.

The coefficients from Step 1 (Equation 2-4) and Step 2 (Equation 5) help identify whether a feature is attributed attention (Step 1) and associated with the predicted outcome (Step 2).

# 7. Results

## 7.1 Prediction Results

We randomly divide our sample of 1620 videos into a 60% training sample (972 videos), a 20% validation sample (324 videos) and a 20% holdout sample (324 videos). We present results using data from the beginning, middle and end 30 seconds of all videos (as shown earlier in Table 4).

### 7.1.1 Individual Models

We train the model on the training sample, tune the number of steps of Adam gradient descent on the validation sample, and then compute predictive performance on the holdout sample. Our parameter choices during model training are guided by the standard values used in the BERT, YAMNet and VGG-16 models. Importantly, we implement a form of bootstrapping, by repeating model training, validation and prediction 50 times (25 times) for every covariate-outcome pair in the Text and Audio models (Image Model) to mitigate concerns of model brittleness. We carry out our analysis using one NVIDIA RTX A6000 GPU (48GB RAM) and 128GB CPU RAM, that takes less than 700 hours to run all models, with the maximum time taken to analyze the Video Image model. For computational ease, we restrict the bootstrap iterations to 25 for the Video Image Model.[9] Bootstrapping allows us to mitigate potential concerns about the models converging to different (local) optima in every instance of model training due to sensitivity to starting values of *hyperparameter* weights randomly chosen by the deep learning models.

We compare the predictive performance of our models with other standard benchmarks used in the marketing literature to demonstrate that our *interpretable* models (with attention mechanisms) do not compromise on predictive ability. We show the detailed results of predictive performance in Online Appendix G and summarize the key results here. We find that our Text model (BERT) has better prediction error than benchmarks such as LSTM, CNN (Liu et al., 2019), CNN-LSTM (Chakraborty et al., 2022) and CNN-Bi-LSTM. This demonstrates the benefit of a model that captures contextual word embeddings, has hierarchical layers and a self-attention mechanism. The Audio model has better prediction error than benchmarks that do not use transfer learning. This demonstrates that addition of transfer learning not only helps with interpretability but also contributes towards the predictive ability of the Audio model. The Image model (VGG-16) has better prediction error than a conventional 4-layer CNN on thumbnail images, thus demonstrating the benefit of transfer learning and a deeper architecture. Overall, we demonstrate that our interpretable models perform better than, or at least as well as, other standard predictive models and thus we do not compromise on predictive ability.

---

[9] For 25 bootstrap iterations, it takes around 36 hours to run the Video Image Model and complete the gradient analysis for each covariate-outcome pair (e.g. beginning 30 sec-sentiment of verbal engagement). It takes a total of 432 hours {36 hours x 4 outcomes x 3 video parts} to analyze all pairs. Hence, for computational ease, we restrict the number of iterations to 25 for the Video Image model but go up till 50 iterations for the Text and Audio models.

Table 6 summarizes the prediction errors from our models, for each component of unstructured data, and for each of the four outcome measures of engagement. Our models predict all the three continuous outcomes with a RMSE ranging from 0.69 to 1.03 and the binary outcome (verbal engagement sentiment) with accuracies ranging from 63% to 72%. For example, *captions/transcript in the beginning 30s* can be used to predict verbal engagement level with an average RMSE range of $\pm e^{0.92} = \pm\, 2.5\ \frac{\#comments+1}{\#views}$. Importantly, the sample standard deviation for the error and accuracy values across all the bootstrap iterations range from 0.01 to 0.03 across all our covariate-outcome pairs. The low value of the standard deviation demonstrates that our model iterations are quite stable. Of all the sources of unstructured data, the title of the video has the best out-of-sample predictive performance, which suggests that *title* is able to better discriminate between engagement values of influencer videos.

| | | Verbal Engagement | | Non-Verbal Engagement | |
|---|---|---|---|---|---|
| Model | Unstructured data | Sentiment (VS) | Level (VL) | Sentiment (NVS) | Level (NVL) |
| Text Model (BERT) | Title | 0.72 (0.02) | 0.83 (0.03) | 0.84 (0.03) | 0.69 (0.02) |
| | Description (first 160c) | 0.70 (0.03) | 0.87 (0.02) | 0.94 (0.03) | 0.72 (0.02) |
| | Captions/transcript (begin 30s) | 0.71 (0.03) | 0.92 (0.02) | 0.98 (0.02) | 0.75 (0.02) |
| | Captions/transcript (middle 30s) | 0.69 (0.02) | 0.98 (0.02) | 1.03 (0.02) | 0.79 (0.02) |
| | Captions/transcript (end 30s) | 0.67 (0.02) | 0.98 (0.02) | 1.03 (0.02) | 0.78 (0.02) |
| Audio Model (YAMNet + Bi-LSTM + Attention) | Audio (begin 30s) | 0.64 (0.01) | 0.92 (0.02) | 1.01 (0.01) | 0.80 (0.01) |
| | Audio (middle 30s) | 0.63 (0.01) | 0.95 (0.01) | 1.02 (0.01) | 0.80 (0.01) |
| | Audio (end 30s) | 0.65 (0.01) | 0.96 (0.01) | 1.01 (0.01) | 0.79 (0.01) |
| Image Model (VGG-16) | Thumbnail | 0.68 (0.02) | 0.96 (0.01) | 1.00 (0.01) | 0.78 (0.01) |
| Video Image Model (VGG-16 + Bi-LSTM) | Video Frames (begin 30s @1fps) | 0.66 (0.02) | 0.92 (0.02) | 0.99 (0.02) | 0.77 (0.02) |
| | Video Frames (middle 30s @1fps) | 0.66 (0.01) | 0.96 (0.02) | 0.99 (0.01) | 0.77 (0.01) |
| | Video Frames (end 30s @1fps) | 0.67 (0.02) | 0.94 (0.03) | 0.98 (0.03) | 0.75 (0.02) |
| Note: The number in brackets is the standard deviation in prediction error across all bootstrap iterations. | | | | | |

*Table 6: Model performance for each component of unstructured data on holdout sample (Accuracy for binary outcome - VS; RMSE for continuous outcomes – VL, NVS, NVL)*

We also find that the prediction error while using any of the unstructured data in captions/transcript, audio or video frames (in each part of the video) are close to each other while predicting all engagement measures. A systematic analysis reveals that no single model (Text,

Audio, or Video Image) predicts *significantly* better than the other model across all 12 results (4 outcomes x 3 parts of video) presented in Table 5. This demonstrates two important things. First, each of the three individual models perform comparably well when using unstructured data from different parts of the video. Second, it suggests that information contained in text, audio and video images is highly correlated. This is not unexpected, as the words spoken by the influencer are likely to be 'in sync' with the accompanying acoustics and imagery.

7.1.2 Combined Model

To estimate the Combined Model in Section 5.2, we first re-predict each of our individual models on the entire sample (training, validation and holdout) to estimate $\widehat{Y}_{it}$ for video $t$ by influencer $i$. We repeat this process for all bootstrap iterations and obtain the average of the predicted values. We then combine them with structured features $X_{it}$ (as shown earlier in Eq (1)) using different combined models, $g$. We train each combined model on the training sample, validate the hyperparameters on the validation sample, and obtain predictions on the holdout sample (see results in Table 7).

| | Verbal Engagement | | Non-Verbal Engagement | |
|---|---|---|---|---|
| | Sentiment (VS) | Level (VL) | Sentiment (NVS) | Level (NVL) |
| OLS | 0.74 | 0.78 | 0.77 | 0.63 |
| Ridge Regression | 0.74 | 0.73 | 0.71 | 0.56 |
| LASSO | 0.73 | 1.17 | 1.29 | 0.96 |
| Elastic Net | 0.74 | 1.12 | 1.20 | 0.91 |
| Deep Neural Net | 0.74 | 0.77 | 0.73 | 0.56 |
| Random Forests | 0.74 | 0.73 | 0.76 | 0.59 |
| XGBoost | 0.72 | 0.77 | 0.76 | 0.63 |

*Table 7: Performance of different Combined Models on holdout sample*
*(Accuracy for binary outcome - VS; RMSE for continuous outcomes – VL, NVS, NVL)*

We find that Ridge Regression, a linear model, has better performance on the holdout sample for all continuous outcomes (lowest RMSE) and the binary outcome (highest accuracy) as compared to OLS, LASSO, Elastic Net, Deep Neural Net, Random Forest and XGBoost. This suggests that linear effects are more salient than cross-modal interactions. This is not surprising given that cross-modal interactions are more likely to occur among the latent dimensions of unstructured data than among the predicted outcome values from the individual models (consistent with the limitations discussed earlier in Table 4).

As noted in Section 7.1.1, the predicted outcomes from the individual models ($\hat{Y}_{it}$) supplied as features to the Combined Model can be collinear with each other. However, all features with multicollinearity in the Ridge Regression model are regularized equally towards the null. Hence their relative importance can be captured by taking the magnitude of each estimated coefficient from the Ridge Regression model applied on the training sample and scaling it by the sum of the magnitude of all coefficient values. We show the relative importance (*marginal effect*) or percentage contribution of each feature while predicting each engagement measure in Table 8 (Panel A).[10]

Note that as the features used in the model ($\hat{Y}$ in Eq 1) can be collinear, the feature importance measures should be interpreted as conveying a combination of *between-influencer* (different influencers' videos) and *within-influencer* (same influencer's videos) importance of a feature. We find that influencer fixed effects are especially important for predicting measures of non-verbal engagement (39.5% for level and 31.8% for sentiment). This suggests that characteristics unique to an influencer are more important in predicting non-verbal reactions from viewers. Now, "incentive to click" data (title, description – first 160c, thumbnail) are more likely to drive a change in views than engagement, since they are visible in search results and prompt the viewer to click on a video. However, their power in explaining engagement with the video suggests a correlation between them and the content of the video. This is especially prominent for title and description (first 160 characters) which explain more variation in all the engagement measures than thumbnail images do.

[10] As we scale all the features by their $L^2$ norm before running the model, the coefficients are regularized to the same degree and hence we can make relative comparisons. Also, we sum up the coefficient values that lie within a class (e.g., sum up coefficient values of influencer fixed effects, sum up other structured features (in Table 3)) to get an overall idea of the contribution of a class of features in predicting each outcome.

| | . | | Verbal Engagement | | Non-Verbal Engagement | |
|---|---|---|---|---|---|---|
| | | | Sentiment | Level | Sentiment | Level |
| Panel A | Structured Features | Influencer Fixed Effects | 18.0% | 2.9% | 31.8% | 39.5% |
| | | Other features (in Table 3) | 36.8% | 2.4% | 7.7% | 10.5% |
| | | Total | 54.8% | 5.4% | 39.4% | 49.9% |
| | Unstructured data | | | | | |
| | Incentive to Click Data | Title | 9.4% | 25.4% | 15.5% | 9.9% |
| | | Description (first 160c) | 9.1% | 22.7% | 13.0% | 7.6% |
| | | Thumbnail | 0.1% | 2.6% | 1.9% | 2.2% |
| | | Total | 18.5% | 50.8% | 30.3% | 19.7% |
| | Captions / Transcript | Begin 30s | 5.7% | 13.7% | 9.5% | 7.1% |
| | | Middle 30s | 5.7% | 10.7% | 4.7% | 5.6% |
| | | End 30s | 5.7% | 7.4% | 4.4% | 5.5% |
| | | Total | 17.0% | 31.7% | 18.7% | 18.2% |
| | Audio | Begin 30s | 1.1% | 3.0% | 1.0% | 1.7% |
| | | Middle 30s | 0.8% | 1.0% | 1.2% | 1.3% |
| | | End 30s | 0.7% | 1.7% | 1.7% | 1.9% |
| | | Total | 2.6% | 5.7% | 3.8% | 4.9% |
| | Video Frames | Begin 30s | 4.1% | 3.0% | 2.4% | 2.4% |
| | | Middle 30s | 1.0% | 1.2% | 2.0% | 2.1% |
| | | End 30s | 2.0% | 2.4% | 3.3% | 2.6% |
| | | Total | 7.1% | 6.5% | 7.7% | 7.1% |
| Panel B | | Total Begin 30s | 10.9% | 19.7% | 12.9% | 11.2% |
| | | Total Middle 30s | 7.4% | 12.8% | 7.9% | 9.1% |
| | | Total End 30s | 8.4% | 11.4% | 9.4% | 10.0% |

*Table 8: Importance of features based on the Combined Model*

Importantly, we find that text (captions/transcript) explains more than *twice* the variation in all engagement measures than audio or video frames across each part of the video (beginning, middle or end). This suggests that words spoken by the influencer are more influential in capturing variation in engagement than audio-visual features. However, the difference in importance between audio and video frames is minimal. Overall, we can conclude that "what is said" (words spoken) is more important than "how it is said" (acoustics or video imagery) to distinguish between engagement levels of different videos (by the same or different influencer). Note that this conclusion is possible because the Text, Audio and Video Image model predict comparatively well when using captions/transcript, audio and video frames respectively (across each part of the video) as discussed earlier in 7.1.1. Hence the results inferred from Table 8 should not be attributed to better predictive ability of an individual model as compared to another, but to the stronger marginal impact of the modality of text (captured by captions/transcript data).

Our key finding on the higher importance of text is consistent with findings in the advertising literature. Tellis (2003) mentions that the strength of the argument is more important than stylistic elements in explaining ad effectiveness. MacInnis and Jaworski (1989) find that comprehension and cognitive responses to verbal content drive persuasion, while presentation elements like imagery and music mainly influence attention and affect. Similarly, the advertising adage of Ogilvy states, "*What you say is more important than how you say it: the information you give is more important to the consumer than the way you present it.*" (Ogilvy, 1983).

Next, we analyze the relative importance of unstructured data in each part of the video in Table 8 (Panel B). We find that, on average, unstructured data in the beginning of videos explain more variation in all the engagement measures than unstructured data in the middle or end of videos. This insight is informative as it provides evidence that viewers need not watch videos completely before making the decision to engage with it. This is consistent with the fact that many viewers indeed do not finish watching YouTube videos (Bump, 2021). Hence, the relatively higher importance of the beginning maybe stemming from its relatively higher contribution to the video viewing experience compared to the middle or end of the video.

#### 7.1.3 Multimodal Model

We train the model on the training sample, tune the number of steps of Adam gradient descent on the validation sample, and then compute predictive performance on the holdout sample. Our parameter choices during model training are guided by the standard values used in the multimodal architecture of Ghosal et al. (2018). As done in Section 7.1.1, we repeat model training, validation and prediction 50 times and obtain the average of the prediction errors to mitigate concerns of model brittleness. We show the results from embeddings in the beginning, middle and end 30 seconds in Table 9. We find that the prediction error from the Multimodal Model is worse than the prediction error that was observed earlier using the Combined Model in Table 7. This demonstrates that the strengths of the Combined Model outweigh the strengths of the Multimodal Model (refer to Table 4 shown earlier in Section 5).

In order to measure relative importance of the modalities (features), we use Permutation Feature Importance (PFI) which is defined as the magnitude of the worsening in prediction error (*increase in RMSE and decrease in accuracy*) when a feature is randomly permuted between observations (Fisher et al., 2019; Molnar et al., 2021). For example, to test whether text (captions/transcript) is important, the embeddings of text are permuted between observations in the holdout sample, but the remaining features are held constant. The magnitude of the worsening in prediction error will inform us about the strength of association (or magnitude of the marginal effect) between text (captions/transcript) and the outcome that was captured during the training process. This approach is also consistent with the way we measured feature importance in the Combined Model by capturing relative marginal effects of features. The advantage of the PFI approach over conventional leave-one-feature-out approaches, is that PFI directly relies on interactions between all modalities that were captured during the training process to measure the model's reliance on a modality to predict the outcome (Fisher et al., 2019). On the other hand, in the leave-one-feature-out approach, a highly collinear feature can more easily compensate for the absence of signals from the main feature. [11]

[11] The more conventional leave-one-feature-out approach is *not* suitable when features are highly collinear as is the case in our setting between text, audio and images (discussed in Section 7.1.1). This is because if one feature is left-out during model training and prediction is carried out on the holdout sample, then the effect of the missing feature on the outcome will be compensated by the other collinear features. The leave-one-feature-out approach assigns more importance to features that are unique and not necessarily to features that have a stronger marginal effect. The PFI approach relies on interactions across all modalities that were captured during the training process and measures the magnitude of the marginal effect of a feature.

| | Verbal Engagement | | Non-Verbal Engagement | |
|---|---|---|---|---|
| | Sentiment (VS) | Level (VL) | Sentiment (VS) | Level (VL) |
| Prediction Error | 0.69 | 0.84 | 0.91 | 0.71 |
| Permuted Prediction Error: Text (Captions/Transcript) | 0.52 | 0.99 | 1.04 | 0.80 |
| Permuted Prediction Error: Audio | 0.69 | 0.88 | 0.94 | 0.73 |
| Permuted Prediction Error: Video Images (Video Frames) | 0.69 | 0.88 | 0.95 | 0.76 |
| Difference in Permuted Prediction Error: Text vs Audio | −0.16*** | 0.10*** | 0.10*** | 0.07*** |
| Difference in Permuted Prediction Error: Text vs Video Images | −0.17*** | 0.11*** | 0.08*** | 0.04*** |
| Note: ***p<0.001 | | | | |

*Table 9: Results of Multimodal Model (Embeddings from Begin, Middle and End 30s)*
*(Accuracy for binary outcome - VS; RMSE for continuous outcomes – VL, NVS, NVL)*

We focus on finding the relative importance between text (captions/transcript), audio and video images (video frames). We compare the difference in prediction errors between the permuted modalities across the 50 bootstrap iterations using a t-test and show the difference in means in Table 9. We find that text is significantly more important than audio or video images (*as there is an increase in RMSE or decrease in accuracy*) while predicting each of our engagement measures, consistent with the findings for the Combined Model in Table 8. Note that the difference in means between audio and video images is minimal (consistent with results of the Combined Model), suggesting that these modalities have comparable importance. Second, we also find that the beginning 30 seconds (across text, audio and video images) is significantly more important than the middle or end, also consistent with results of the Combined Model in Table 8 (see details in Online Appendix E.2). Overall, we are able to validate all conclusions drawn from the Combined Model, using an alternative approach with the Multimodal model. We also validate the superiority of text using data from the complete video in Section 7.3.

### 7.2 Interpretation Results

We use the individual deep learning models tailor-made for each modality of data for ex-post interpretation. We do so by re-predicting each of our individual models on the entire sample of 1620 videos to take advantage of complete information in our data. Note that our interpretation is carried out for each bootstrap iteration that is repeated 50 times (25 times) for every covariate-outcome pair in Text and Audio models (Video Image Model), to mitigate potential concerns of brittleness (as discussed earlier in Section 7.1). We examine the association between 9 features of interest (Table 5) present in 3 parts of long-form videos (beginning, middle and end) and 4 engagement measures, resulting in 108 possible hypotheses (9 x 3 x 4).

We assess whether the association is in the same direction (and significant) in at least 80% of bootstrap iterations across both Step 1 and Step 2, in order to identify robust relationships suitable for formal causal testing. The choice of the 80% threshold is validated in Section 7.3. A relationship selected in only one of the two steps is classified as spurious. After pruning 53 spurious associations, we identify a subset of 21 robust relationships, as summarized in Table 10. The remaining 34 associations (108 – 21 – 53) represent null effects, which were either infrequently significant in Steps 1 and 2, or not robust to sensitivity checks. We provide further details on our interpretation results and robustness checks in Online Appendix H.

| | Robust | | | | Spurious | | | | % Spurious |
|---|---|---|---|---|---|---|---|---|---|
| | Begin | Middle | End | Total | Begin | Middle | End | Total | |
| Text | 1 | 1 | | 2 | 3 | 3 | 4 | 10 | 83% |
| Audio | 3 | 1 | 1 | 5 | 3 | 6 | 9 | 18 | 78% |
| Video Images | 6 | 4 | 4 | 14 | 8 | 10 | 7 | 25 | 64% |
| Total | 10 | 6 | 5 | 21 | 14 | 19 | 20 | 53 | 72% |

*Table 10: Tabulation of robust and spurious associations*

The robust relationships were obtained while controlling various features across unstructured and structured data sources. However, our approach does not guarantee causality of the 21 relationships because of potential correlation with unobservables such as influencer or brand promotions of the video across various digital channels. Despite this limitation, our method substantially reduces the effort needed for future causal work (e.g., field experiments) by pruning 72% of (frequently) significant associations that are spurious and shortlisting 28% of relationships for formal causal testing. Of the 21 relationships identified, 10 correspond to beginning 30 seconds of the video, 6 to the middle 30 seconds and 5 to the end 30 seconds. This reduction in identified relationships at the middle and end of the video as compared to the beginning, suggests that features at a video's onset are salient and contribute more towards viewer engagement, consistent with our prediction results in Section 7.1.

We summarize the 10 robust results for features in the beginning 30 seconds in Table 11. The highlighted cells in grey are linked to theory (discussed earlier in Table 5). The direction of association between features and outcomes is indicated by a + or – sign. Note that to interpret results of the Text Model, we use ridge regression to capture heterogeneity in effects across brand names and hence do not report the effect size (but only the direction of the effect) since estimates of ridge regression are biased (see details in Online Appendix H).

| | | Verbal Engagement | | Non-Verbal Engagement | | Expected change in sentiment before interpreting model (Table 5) |
|---|---|---|---|---|---|---|
| Unstructured Data | Features | Senti-ment | Engage-ment Level | Senti-ment | Engage-ment Level | |
| Text: Captions/ Transcript | Brand names | – | | | | Ambiguous |
| Audio | Music | +30.6% | –2.0% | | –0.9% | Positive |
| Video Images | Human | | | +0.4% | +0.4% | Positive |
| | Animal | | | +1.0% | +0.9% | Positive |
| | Packaged goods | | | –0.7% | –0.5% | Negative |

*Table 11: Robust interpretation results for the beginning 30 seconds*

We describe the results in Table 11. We find that brand mentions in the beginning are associated with a decrease in the sentiment of verbal engagement. Notably, a tabulation shows that the negative association is dominated by brands in the electronics and digital categories, where related ads are typically more informative and less entertaining, reflecting the generally functional rather than hedonic nature of these products. This suggests that a decrease in entertainment value resulting from a discussion of commercial content in the beginning may lead viewers to respond unfavorably or neutrally in their comments (Teixeira et al., 2010; Tellis et al., 2019), as also noted earlier in Table 5. This aligns with previous research showing a decrease in the sharing of commercial content on YouTube (Tellis et al., 2019). We find no strong evidence of a relationship between emotional words and sentiment of engagement.

We also find that an increase in the duration of music (without simultaneous speech) by one moment (about half a second) in the beginning is associated with an increase in the odds ratio of sentiment of verbal engagement by 30.6% which aligns with our general expectation in Table 5, where we highlighted the positive affective influence of music in advertising literature (Pelsmacker & Van den Bergh, 1999). We also find a 2.0% decrease in verbal engagement level which suggests a decrease in negative comments, which contributes to the increase in verbal engagement sentiment. While we also find a 0.9% decrease in non-verbal engagement level, there is no significant association with non-verbal engagement sentiment. Note that we find no robust association between *rapid* speech in any part of the video and our engagement measures, consistent with the expectation of linguists (Beck, 2015; Jennings, 2021).

For video images, we find that an increase in size of human images (packaged goods) by 1% in the beginning is associated with a 0.4% increase (0.7% decrease) in sentiment of non-verbal engagement. These findings are consistent with Hartmann et al. (2021) who found that consumer selfies (human images) receive more likes than brand selfies and packshots (packaged goods) on Instagram. The desire to socialize with other humans likely drives these associations (Hartmann et al., 2021; To & Patrick, 2021; Xiao & Ding, 2014). We also find that a 1% increase in the size of animals at the beginning is associated with a 1.0% increase in sentiment of non-verbal engagement. This is consistent with prior research in advertising where ads featuring animals have been found to be more likeable and have a low irritation score (Biel & Bridgwater, 1990; Pelsmacker & Van den Bergh, 1999). We do not find frequent significant associations between facial expressions of joy or surprise and our engagement measures, which aligns with results in related domains (Li & Xie, 2020; Zhou et al., 2021). This likely occurs because inferences of emotion, whether made by humans or AI, based on facial expressions in images rely on common heuristics and may not reflect the true emotion being experienced (Barrett et al., 2019).

Importantly, our key results show distinct associations between video features and sentiment of verbal versus non-verbal engagement, a difference not pinpointed in prior research. During the critical period of video onset (in the first 30 seconds), our *text and audio* features are linked to sentiment of verbal engagement, while our *video image* features are linked to sentiment of non-verbal engagement. This suggests that viewers often express their sentiment towards *auditory stimuli*—brand mentions and music—via *verbal engagement* in comments below the video. In contrast, viewers often express their sentiment towards *visual stimuli*—video images of humans, animals and packaged goods—via *non-verbal engagement* such as clicking the thumbs-up (like) or thumbs-down (dislike) button. This can be explained by prior research in sensory marketing where different sensory modalities have been found to elicit distinct emotional and cognitive responses, (Krishna, 2012), which, in turn, can map onto different types of engagement behaviors. Our findings offer interesting hypotheses that can be formally tested in future research via causal experiments.

### 7.3 Validation

Next, we conduct systematic analyses to validate our modelling approaches and results.

#### 7.3.1 Validating Prediction Approach

We validate whether the main conclusions from the prediction approach hold true using data from the complete video and not just the beginning, middle and end 30 seconds. We accomplish this efficiently using the multimodal approach shown earlier in Figure 4, where we create embeddings from each successive 30-second clip of a video. We find that text continues to be significantly more important than audio or video images in predicting engagement even when we use information from the entire video and not just the beginning, middle or end (please see Online Appendix I for details).

#### 7.3.1 Validating Interpretation Approach

We validate our interpretation approach through three systematic analyses. First, we re-estimate Equation 5 on a random 80% sample of the 1,620 videos and find that all key findings from Table 11 remain consistent, with similar effect sizes and significance, confirming that our results are robust to the sample size and that our transfer learning methods are effective with moderate-sized samples. Second, we carry out simulations which show that our method accurately recovers the true data-generating process while pruning spurious relationships using an 80% threshold in both Steps 1 and 2 (see Online Appendix J for details). Third, we apply benchmark feature selection methods, such as LASSO and Elastic Net, to the simulated data and demonstrate that our approach yields better results (see Online Appendix J for details).

## 8. Conclusion

This paper advances the limited body of work explaining the relative importance of unstructured data modalities in predicting video engagement on social media. While deep learning models leveraging unstructured data excel in out-of-sample predictions, they often struggle with interpretability, particularly in business settings where human validation is difficult because true drivers of business outcomes are not well understood. We address this challenge by developing an “interpretable deep learning framework” that not only predicts effectively but also allows interpretation of the captured relationships between video features of interest and measures of engagement. Interpretation is accomplished using a novel two-step approach that prunes several spurious relationships. By shortlisting a subset of relationships, our approach reduces the effort required for future causal work. We apply our framework to long-form YouTube influencer videos, a growing but understudied area in influencer marketing, and uncover novel associations between video features and engagement. The engagement measures we study are conditioned on views, and thus capture unique constructs, overcoming a significant limitation of past social media research which has typically studied various engagement measures synonymously.

**8.1 Managerial Implications**

The findings from our research can be useful for practitioners in many ways. First, our prediction-based results demonstrate that influencers' choice of words helps distinguish engagement values more than video imagery or acoustics. This insight empowers influencers by providing them starting points to prioritize their efforts while designing videos and assessing changes via A/B testing. Changing their choice of words could be easier and more cost effective for influencers than changing video imagery (e.g., shooting location and use of visual effects) or acoustics (e.g., microphone quality and background music). Second, as mentions of electronics and digital brands are more often associated with a decrease in sentiment of verbal engagement, brands in these categories may find it useful to investigate whether the influencers they partner with are sacrificing entertainment value for an increase in informativeness while discussing their brands. Similarly, packaged good brands may investigate the impact of having their images displayed later in the video than in the beginning where it is associated with a decrease in sentiment of non-verbal engagement. Finally, our approach also allows us to visualize the level of salience attributed to various features in video data, providing a valuable tool for agencies and influencers to identify a broader set of features for A/B testing (details in Online Appendix K).

### 8.2 Limitations

This work has some limitations. First, our sample consists of top-performing influencers across Facebook, Twitter, YouTube and Instagram who use brand endorsements. As a result, findings may not generalize to influencers who are not high performers, on average, across all platforms. However, many influencers in our sample do not have large followings on YouTube, as reflected in their small to moderate subscriber counts (see Section 4), which helps mitigate this concern to some extent. Second, while YouTube is a major platform for long-form video content, our findings may not generalize to platforms where such formats are less common or to newer formats like YouTube Shorts. Third, while our two-step process prunes many spurious relationships, it does not guarantee causality of all the identified relationships due to potential confounding with unobserved factors. By narrowing the analysis to 21 "feature – video part – outcome" triads of interest (out of 108 possibilities), our approach substantially reduces the effort needed to test hypotheses in future causal research, such as field experiments. In addition, future research could explore interactions between auditory and visual features, and assess their causal impact on engagement. Finally, our results represent average associations and effect sizes which may vary by video type (e.g., product reviews, unboxing videos) or viewing device (e.g., mobile, desktop, tablet). We hope that future work can examine these dimensions more closely.

**Declaration of Competing Interest**

The authors declare that they have no known competing financial interests or personal relationships that could have appeared to influence the work reported in this paper.

### Declaration of generative AI and AI-assisted technologies in the writing process

During the preparation of this work the author(s) used chatGPT for copy editing. After using this tool/service, the author(s) reviewed and edited the content as needed and take(s) full responsibility for the content of the publication.

### References

Alain, T. (2023). *Influencer engagement: everything you need to know*. Retrieved January 31 from https://www.upfluence.com/influencer-marketing/influencer-engagement-everything-you-need-to-know

Bahdanau, D., Cho, K., & Bengio, Y. (2014). Neural machine translation by jointly learning to align and translate. *arXiv preprint arXiv:1409.0473*.

Bakir, V., & McStay, A. (2018). Fake news and the economy of emotions: Problems, causes, solutions. *Digital journalism*, *6*(2), 154-175.

Barrett, L. F., Adolphs, R., Marsella, S., Martinez, A. M., & Pollak, S. D. (2019). Emotional expressions reconsidered: Challenges to inferring emotion from human facial movements. *Psychological science in the public interest*, *20*(1), 1-68.

Beck, J. (2015). *The Linguistics of 'YouTube Voice'*. Retrieved December 7 from https://www.theatlantic.com/technology/archive/2015/12/the-linguistics-of-youtube-voice/418962/

Berger, J., & Milkman, K. L. (2012). What makes online content viral? *Journal of Marketing Research*, *49*(2), 192-205.

Biel, A. L., & Bridgwater, C. A. (1990). Attributes of likable television commericals. *Journal of Advertising Research*, *30*(3), 38-44.

Bump, P. (2021). *Why People Click Out of YouTube Videos [New Data]*. Retrieved April 12 from https://blog.hubspot.com/marketing/why-people-click-out-of-youtube-videos

Burnap, A., Hauser, J. R., & Timoshenko, A. (2023). Product aesthetic design: A machine learning augmentation. *Marketing Science*.

Chakraborty, I., Kim, M., & Sudhir, K. (2022). Attribute sentiment scoring with online text reviews: Accounting for language structure and missing attributes. *Journal of Marketing Research*, *59*(3), 600-622.

Chen, L., Yan, Y., & Smith, A. N. (2022). What drives digital engagement with sponsored videos? An investigation of video influencers' authenticity management strategies. *Journal of the Academy of Marketing Science*, 1-24.

Cheng, M., & Zhang, S. (2024). Reputation burning: Analyzing the impact of brand sponsorship on social influencers. *Management Science*.

Cournoyer, B. (2014, March 19). *YouTube SEO Best Practices: Titles and Descriptions*. https://www.brainshark.com/ideas-blog/2014/March/youtube-seo-best-practices-titles-descriptions

Covington, P., Adams, J., & Sargin, E. (2016). Deep neural networks for youtube recommendations. Proceedings of the 10th ACM conference on recommender systems,

Devlin, J., Chang, M.-W., Lee, K., & Toutanova, K. (2018). Bert: Pre-training of deep bidirectional transformers for language understanding. *arXiv preprint arXiv:1810.04805*.

Dew, R., Ansari, A., & Toubia, O. (2022). Letting logos speak: Leveraging multiview representation learning for data-driven branding and logo design. *Marketing Science*, *41*(2), 401-425.

Digital Marketing Institute. (2024). *20 Surprising Influencer Marketing Statistics*. https://digitalmarketinginstitute.com/blog/20-influencer-marketing-statistics-that-will-surprise-you

Draelos, R. L., & Carin, L. (2020). Use HiResCAM instead of Grad-CAM for faithful explanations of convolutional neural networks. *arXiv preprint arXiv:2011.08891*.

Dwoskin, J. (2021). *There's A Big Difference Between Likes and Comments*. Retrieved June 5 from https://www.stampede.social/there-s-a-big-difference-between-likes-and-comments

Dzyabura, D., El Kihal, S., Hauser, J. R., & Ibragimov, M. (2023). Leveraging the power of images in managing product return rates. *Marketing Science*.

Ershov, D., He, Y., & Seiler, S. (2025). Frontiers: How Much Influencer Marketing Is Undisclosed? Evidence from Twitter. *Marketing Science*.

Fields, A. (2025). *Influencer Marketing Pricing: What's It Cost in 2025?* Retrieved April 4 from https://www.webfx.com/blog/social-media/influencer-marketing-pricing/

Fisher, A., Rudin, C., & Dominici, F. (2019). All Models are Wrong, but Many are Useful: Learning a Variable's Importance by Studying an Entire Class of Prediction Models Simultaneously. *J. Mach. Learn. Res.*, *20*(177), 1-81.

Fong, H., Kumar, V., & Sudhir, K. (2025). A theory-based explainable deep learning architecture for music emotion. *Marketing Science*, *44*(1), 196-219.

FTC. (2020). *Disclosures 101 for Social Media Influencers*. https://www.ftc.gov/system/files/documents/plain-language/1001a-influencer-guide-508_1.pdf

Gemmeke, J. F., Ellis, D. P., Freedman, D., Jansen, A., Lawrence, W., Moore, R. C., Plakal, M., & Ritter, M. (2017). Audio set: An ontology and human-labeled dataset for audio events. *2017 IEEE International Conference on Acoustics, Speech and Signal Processing*, 776-780.

Ghosal, D., Akhtar, M. S., Chauhan, D., Poria, S., Ekbal, A., & Bhattacharyya, P. (2018). Contextual inter-modal attention for multi-modal sentiment analysis. proceedings of the 2018 conference on empirical methods in natural language processing,

Goh, K.-Y., Heng, C.-S., & Lin, Z. (2013). Social media brand community and consumer behavior: Quantifying the relative impact of user-and marketer-generated content. *Information Systems Research*, *24*(1), 88-107.

Google. (2020). *Add tags to videos*. https://support.google.com/youtube/answer/146402?hl=en

Gu, X., Zhang, X., & Kannan, P. (2024). Influencer mix strategies in livestream commerce: impact on product sales. *Journal of Marketing*, *88*(4), 64-83.

Guo, C., Cao, J., Zhang, X., Shu, K., & Yu, M. (2019). Exploiting emotions for fake news detection on social media. *arXiv preprint arXiv:1903.01728*.

Guo, P. J., Kim, J., & Rubin, R. (2014). How video production affects student engagement: An empirical study of MOOC videos. *Proceedings of the first ACM conference on Learning@ scale conference*, 41-50.

Hartmann, J., Heitmann, M., Schamp, C., & Netzer, O. (2021). The power of brand selfies. *Journal of Marketing Research*, *58*(6), 1159-1177.

He, J., Li, B., & Wang, X. S. (2023). Image features and demand in the sharing economy: A study of Airbnb. *International Journal of Research in Marketing*, *40*(4), 760-780.

Huang, Y., & Morozov, I. (2025). The promotional effects of live streams by Twitch influencers. *Marketing Science*.

Hughes, C., Swaminathan, V., & Brooks, G. (2019). Driving Brand Engagement Through Online Social Influencers: An Empirical Investigation of Sponsored Blogging Campaigns. *Journal of Marketing*.

Hwang, S., Liu, X., & Srinivasan, K. (2022). Voice Analytics of Online Influencers. *Working Paper, Available at SSRN 3773825*.

Influencer Marketing Hub. (2022). *The State of Influencer Marketing*. https://influencermarketinghub.com/ebooks/Influencer_Marketing_Benchmark_Report_2022.pdf

Jain, S., & Wallace, B. C. (2019). Attention is not explanation. *arXiv preprint arXiv:1902.10186*.

Jennings, R. (2021). *How should an influencer sound?* Retrieved July 13 from https://www.vox.com/the-goods/2021/7/13/22570476/youtube-voice-tiktok-influencer-sound

Krishna, A. (2012). An integrative review of sensory marketing: Engaging the senses to affect perception, judgment and behavior. *Journal of Consumer Psychology*, *22*(3), 332-351.

Lanz, A., Goldenberg, J., Shapira, D., & Stahl, F. (2019). Climb or Jump: Status-Based Seeding in User-Generated Content Networks. *Journal of Marketing Research*, *56*(3), 361-378.

Lanz, A., Goldenberg, J., Shapira, D., & Stahl, F. (2024). Buying Future Endorsements from Prospective Influencers on User-Generated Content Platforms. *Journal of Marketing Research*, 00222437231207323.

Lee, D., Hosanagar, K., & Nair, H. S. (2018). Advertising content and consumer engagement on social media: Evidence from Facebook. *Management Science*, *64*(11), 5105-5131.

Lee, D. D., Cheng, Z. Z., Mao, C., & Manzoor, E. (2024). Guided Diverse Concept Miner (GDCM): Uncovering Relevant Constructs for Managerial Insights from Text. *Information Systems Research*.

Leung, F. F., Gu, F. F., Li, Y., Zhang, J. Z., & Palmatier, R. W. (2022). EXPRESS: Influencer Marketing Effectiveness. *Journal of Marketing*, 00222429221102889.

Li, X., Shi, M., & Wang, X. S. (2019). Video mining: Measuring visual information using automatic methods. *International Journal of Research in Marketing*, *36*(2), 216-231.

Li, Y., & Xie, Y. (2020). Is a picture worth a thousand words? An empirical study of image content and social media engagement. *Journal of Marketing Research*, *57*(1), 1-19.

Liu, L., Dzyabura, D., & Mizik, N. (2020). Visual listening in: Extracting brand image portrayed on social media. *Marketing Science*, *39*(4), 669-686.

Liu, X., Lee, D., & Srinivasan, K. (2019). Large-scale cross-category analysis of consumer review content on sales conversion leveraging deep learning. *Journal of Marketing Research*, *56*(6), 918-943.

Lu, S., Xiao, L., & Ding, M. (2016). A video-based automated recommender (VAR) system for garments. *Marketing Science*, *35*(3), 484-510.

MacInnis, D. J., & Jaworski, B. J. (1989). Information processing from advertisements: Toward an integrative framework. *Journal of Marketing*, *53*(4), 1-23.

Mitchell, A. A. (1986). The effect of verbal and visual components of advertisements on brand attitudes and attitude toward the advertisement. *Journal of Consumer Research*, *13*(1), 12-24.

Molnar, C., Freiesleben, T., König, G., Casalicchio, G., Wright, M. N., & Bischl, B. (2021). Relating the partial dependence plot and permutation feature importance to the data generating process. *arXiv preprint arXiv:2109.01433*.

O'Connor, C. (2017, September 26). *Forbes Top Influencers: Meet The 30 Social Media Stars Of Fashion, Parenting And Pets (Yes, Pets)*. https://www.forbes.com/sites/clareoconnor/2017/09/26/forbes-top-influencers-fashion-pets-parenting/

Ogilvy, D. (1983). *Confessions of an advertising man*. Atheneum New York.

Overgoor, G., Rand, W., van Dolen, W., & Mazloom, M. (2022). Simplicity is not key: Understanding firm-generated social media images and consumer liking. *International Journal of Research in Marketing*, *39*(3), 639-655.

Parsons, J. (2017, August 24). *How Long Until Watching a YouTube Video Counts as a View?* https://growtraffic.com/blog/2017/08/youtube-video-counts-view

Pelsmacker, P. D., & Van den Bergh, J. (1999). Advertising content and irritation: a study of 226 TV commercials. *Journal of international consumer marketing*, *10*(4), 5-27.

Peterson, R. A., Cannito, M. P., & Brown, S. P. (1995). An exploratory investigation of voice characteristics and selling effectiveness. *Journal of Personal Selling & Sales Management*, *15*(1), 1-15.

Pieters, R., & Wedel, M. (2004). Attention capture and transfer in advertising: Brand, pictorial, and text-size effects. *Journal of Marketing*, *68*(2), 36-50.

Pilakal, M., & Ellis, D. (2020). *YAMNet*. https://github.com/tensorflow/models/tree/master/research/audioset/yamnet

Puranam, D., Kadiyali, V., & Narayan, V. (2021). The impact of increase in minimum wages on consumer perceptions of service: A transformer model of online restaurant reviews. *Marketing Science*, *40*(5), 985-1004.

Quuu Blog. (2024). *Predictive Analytics for Influencer Marketing Success in 2024*. https://blog.quuu.co/predictive-analytics-for-influencer-marketing-success-in-2024/

Selvaraju, R. R., Cogswell, M., Das, A., Vedantam, R., Parikh, D., & Batra, D. (2017). Grad-cam: Visual explanations from deep networks via gradient-based localization. Proceedings of the IEEE international conference on computer vision,

Shaikh, A. (2024). *Most Popular Social Media Platforms: What's Working in 2024?* https://www.socialchamp.io/blog/most-popular-social-media-platforms/

Simonyan, K., & Zisserman, A. (2014). Very deep convolutional networks for large-scale image recognition. *arXiv preprint arXiv:1409.1556*.

Statista. (2023). *Global Influencer Marketing Size from 2020 to 2025*. https://www.statista.com/statistics/1328195/global-influencer-market-value/

Teixeira, T., Wedel, M., & Pieters, R. (2010). Moment-to-moment optimal branding in TV commercials: Preventing avoidance by pulsing. *Marketing Science*, *29*(5), 783-804.

Teixeira, T., Wedel, M., & Pieters, R. (2012). Emotion-induced engagement in internet video advertisements. *Journal of Marketing Research*, *49*(2), 144-159.

Tellis, G. J. (2003). *Effective advertising: Understanding when, how, and why advertising works*. Sage Publications.

Tellis, G. J., MacInnis, D. J., Tirunillai, S., & Zhang, Y. (2019). What drives virality (sharing) of online digital content? The critical role of information, emotion, and brand prominence. *Journal of Marketing*, *83*(4), 1-20.

Tian, Z., Dew, R., & Iyengar, R. (2023). EXPRESS: Mega or Micro? Influencer Selection Using Follower Elasticity. *Journal of Marketing Research*, 00222437231210267.

To, R. N., & Patrick, V. M. (2021). How the eyes connect to the heart: The influence of eye gaze direction on advertising effectiveness. *Journal of Consumer Research*, *48*(1), 123-146.

Vashishth, S., Upadhyay, S., Tomar, G. S., & Faruqui, M. (2019). Attention interpretability across nlp tasks. *arXiv preprint arXiv:1909.11218*.

Wiegreffe, S., & Pinter, Y. (2019). Attention is not not explanation. *arXiv:1908.04626*.

Wu, Z., Jiang, Y.-G., Wang, J., Pu, J., & Xue, X. (2014). Exploring inter-feature and inter-class relationships with deep neural networks for video classification. Proceedings of the 22nd ACM international conference on Multimedia,

Xiao, L., & Ding, M. (2014). Just the faces: Exploring the effects of facial features in print advertising. *Marketing Science*, *33*(3), 338-352.

Yang, J., Zhang, J., & Zhang, Y. (2025). Engagement that sells: Influencer video advertising on TikTok. *Marketing Science*, *44*(2), 247-267.

Yue-Hei Ng, J., Hausknecht, M., Vijayanarasimhan, S., Vinyals, O., Monga, R., & Toderici, G. (2015). Beyond short snippets: Deep networks for video classification. *Proceedings of the IEEE conference on computer vision and pattern recognition*, 4694-4702.

Zhang, M., & Luo, L. (2023). Can consumer-posted photos serve as a leading indicator of restaurant survival? Evidence from yelp. *Management Science*.

Zhang, S., Lee, D., Singh, P. V., & Srinivasan, K. (2021). What makes a good image? Airbnb demand analytics leveraging interpretable image features. *Management Science*.

Zhou, M., Chen, G. H., Ferreira, P., & Smith, M. D. (2021). Consumer Behavior in the Online Classroom: Using Video Analytics and Machine Learning to Understand the Consumption of Video Courseware. *Journal of Marketing Research*, *58*(6), 1079-1100.

## Online Appendices

## Unboxing Engagement in YouTube Influencer Videos: An Attention-Based Approach

## Online Appendix A – Theory of Model Attention

Three popular attention mechanisms introduced in seminal papers in the deep learning literature are additive attention, scaled dot-product attention and gradient-based attention. We explain how spurious associations can emerge in each attention mechanism and our approach to prune such associations.

### A.1 Additive Attention

Introduced by Bahdanau et al. (2014), additive attention improves neural machine translation by allowing models to focus on different parts of the input, thereby enhancing performance over fixed-size context vectors. It functions similarly to a gating-unit that either allows or blocks the passage of information (see Vashishth et al. (2019) for details) by adding hidden state vectors and passing them through a feedforward neural network to generate attention scores. The mechanism has since been widely adopted in sequence-to-sequence models but is known to allow spurious associations.

Vashishth et al. (2019) investigate interpretability of attention mechanisms in single-sequence tasks such as sentiment analysis on the IMDb dataset. They find that the originally learned model attention weights are both meaningful (i.e., correspond to words that are truly linked with sentiment) and associated with the correct prediction 79.5% of the time (see Figure A1).

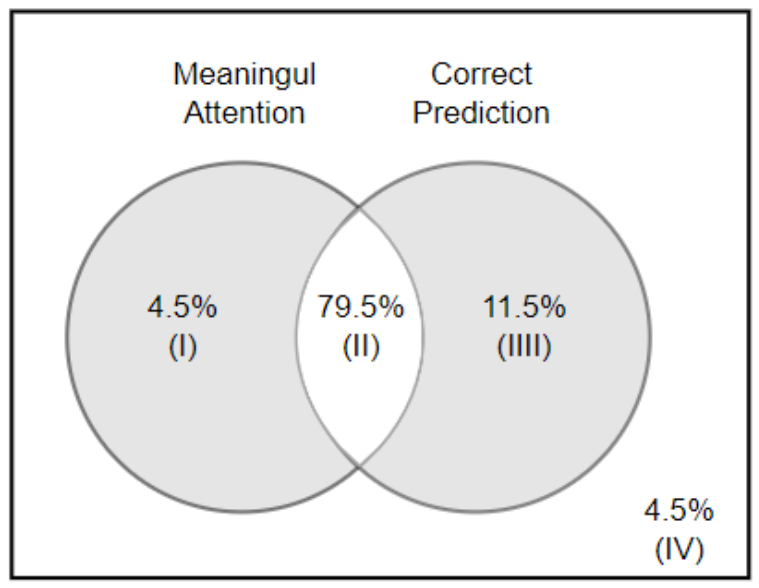

*Figure A1: Additive Attention: Results of Sentiment Analysis on IMDb dataset (Vashishth et al., 2019)*

However, there are instances when the model learns meaningful attention weights but still makes incorrect predictions (4.5% of the time). Additionally, there are cases when the model learns meaningless attention weights but the predictions are correct (11.5% of the time). These are spurious associations which we highlight in grey in Figure A1. The remaining instances (4.5% of the time) involve both meaningless attention weights and incorrect predictions.

Thus, Figure A1 demonstrates that models using the additive attention mechanism on single-sequence tasks most often identify relationships that align with both meaningfulness and correctness (corresponding to the ground truth underlying the data) but can also pick up spurious associations. The approach to test for meaningfulness in Vashishth et al. (2019) rely on human validation of whether the words (with high attention weights) in sentences convey sentiment, a common approach in the machine learning literature. Such an approach is not easily replicable in business settings such as ours where the ground truth linking influencer video features and engagement measures is often unknown. To address this challenge, we investigate whether deep learning models attribute attention to the focal feature of interest and also whether that focal feature is correlated with the predicted outcome (see Figure A2).

As shown in Figure A2, there can be situations where the focal feature is attributed attention but the feature is *not* correlated with the predicted outcome, resulting in Type A spuriousness. This can occur when (a) the focal feature is meaningless (but is attributed attention) leading to incorrect predictions (Category IV in Figure A1), or (b) the focal feature is meaningful (and is attributed attention), yet the predictions are incorrect (Category I in Figure A1). Similarly, there can be situations when the focal feature is *not* attributed attention but the feature is correlated with the predicted outcome, resulting in Type B spuriousness. This can happen when (a) the presence of the focal feature is correlated with the presence of other meaningless features (to which attention is attributed) that are associated with correct predictions (Category III in Figure A1), or (iv) the presence of the focal feature is correlated with the presence of other meaningful features (to which attention is attributed) that are associated with correct predictions (Category II in Figure A1). Thus, theoretically, both Type A and Type B spuriousness are possible.

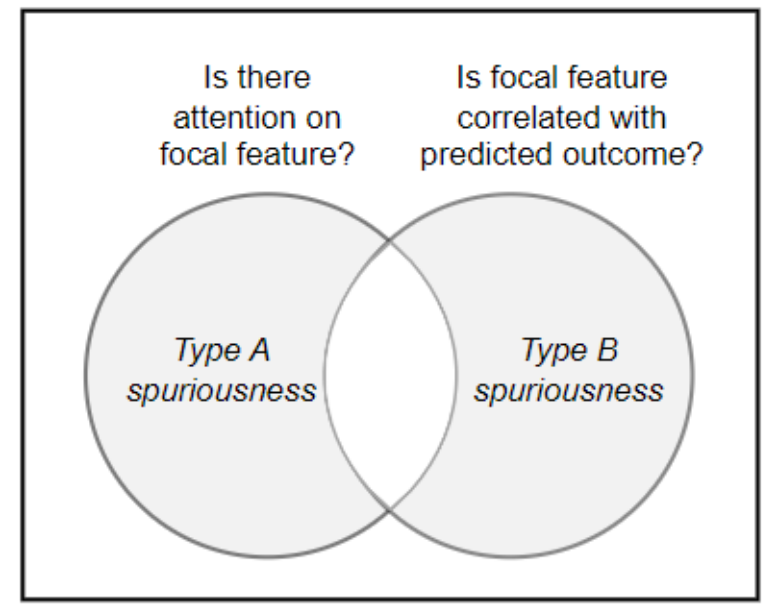

*Figure A2: Types of spurious associations*

### A.2 Scaled Dot-Product Attention

Scaled Dot-Product attention, introduced by Vaswani et al. (2017), has become the core component of self-attention and cross-attention mechanisms in modern architectures like transformers. It takes the dot product of hidden state vectors followed by scaling, simultaneously computing attention scores for all tokens (word-pieces) relative to a given token. This method is better suited than additive attention for capturing long range dependencies, improving computational efficiency and enabling parallel processing.[1]

When attention weights in encoder-based transformer models (e.g., BERT, RoBERTa, etc.) are randomly permuted (giving rise to meaningless features), the rate of decrease in prediction ability is slower (see Figure 4b in Vashishth et al. (2019)) compared to encoder-decoder-based transformer models used for generation of unstructured data (e.g., GPT). This can be problematic for encoder-based transformer models when used for single sequence tasks such as classification or regression (see Figure 4a in Vashishth et al. (2019)). In such cases, the model might produce correct predictions while attributing attention to meaningless features. As discussed earlier, if the presence of the focal feature is correlated with the presence of other (meaningless or meaningful) features that are both given attention and associated with the predicted outcome, this can result in Type B spurious associations. While Type A spurious associations may also occur, the likelihood of Type B spurious associations using scaled dot-product attention on single sequence tasks is higher as it has been documented in prior work.

### A.3 Gradient-based Attention

A generalized approach for gradient-based attention named Grad-CAM (Gradient-weighted Class Activation Mapping) was introduced by Selvaraju et al. (2017) to interpret CNNs (Convolutional Neural Networks) by highlighting image regions that influence predictions. Despite the rise of transformer-based methods, CNNs remain popular for image analysis due to their established effectiveness and relatively lower computational demands. Grad-CAM computes gradients with respect to the last convolutional layer, averages them over each feature map channel to determine importance, and uses these values as weights to produce saliency scores that highlight image regions influencing the outcome. While the averaging process in Grad-CAM helps reduces noise in the gradients (Selvaraju et al., 2017), it also has a downside: it can expand the attention map to include areas not actually used by the model for prediction (see Figure 1 in Draelos and Carin (2020)). However, it better identifies pixels associated with the focal object as compared to other approaches such as HiResCAM (Draelos & Carin, 2020)[2].

[1] Additive attention can capture more complexity (than scaled dot-product attention) due to its use of a non-linear feedforward neural network, making it preferable when capturing complexity is more important.

Since Grad-CAM can expand the attention map, it may attribute spurious attention to features not actually used by the model for prediction, leading to Type A spurious associations. While Type B spurious associations may also occur, the likelihood of Type A spurious associations using gradient-based attention is higher as it has been documented in prior work.

[2] Draelos and Carin (2020) propose a modified framework, HiResCAM, that does not do averaging like Grad-CAM but instead multiplies the gradients element-wise with the feature map. This approach better discriminates the pixels used for prediction (e.g., using a nearby car to predict presence of a bus in an image), but does a poorer job of identifying all pixels associated with the focal object. In other words, HiResCAM excels at pinpointing the pixels used for predicting an object even if those pixels lie outside the object. However, this is not ideal for our setting as our framework (shown in Figure 3, Section 3.3) aims to investigate whether a focal image feature, rather than other correlated features, is associated with the outcome (engagement). Hence, we use Grad-CAM and not HiResCAM.

## Online Appendix B – Operationalization of Text Model

We provide an overview of the implementation of our Text Model – the BERT Framework (Devlin et al., 2018) and explain the architecture of the encoders and attention mechanism (used in the BERT framework).

### B.1 BERT Framework

The BERT model converts a sentence into word-piece tokens[3] as done by state-of-the-art machine translation models (Wu et al., 2016). Furthermore, the beginning of each sentence is appended by the 'CLS' (classification) token and the end of each sentence is appended by the 'SEP' (separation token). For example, the sentence 'Good Morning! I am a YouTuber.' will be converted into the (word-pieces or) tokens ['[CLS]', 'good', 'morning', '!', 'i', 'am', 'a', 'youtube', '##r', '.', '[SEP]']. A 768-dimensional initial embedding learnt for each token during the pre-training phase is passed as input to the model, and is represented by the vector $x_m$ in Figure B1, where m is the number of tokens in the longest sentence. This vector $x_m$ has pre-learned contextual embeddings that will aid the model in capturing relationships with our four outcome variables.

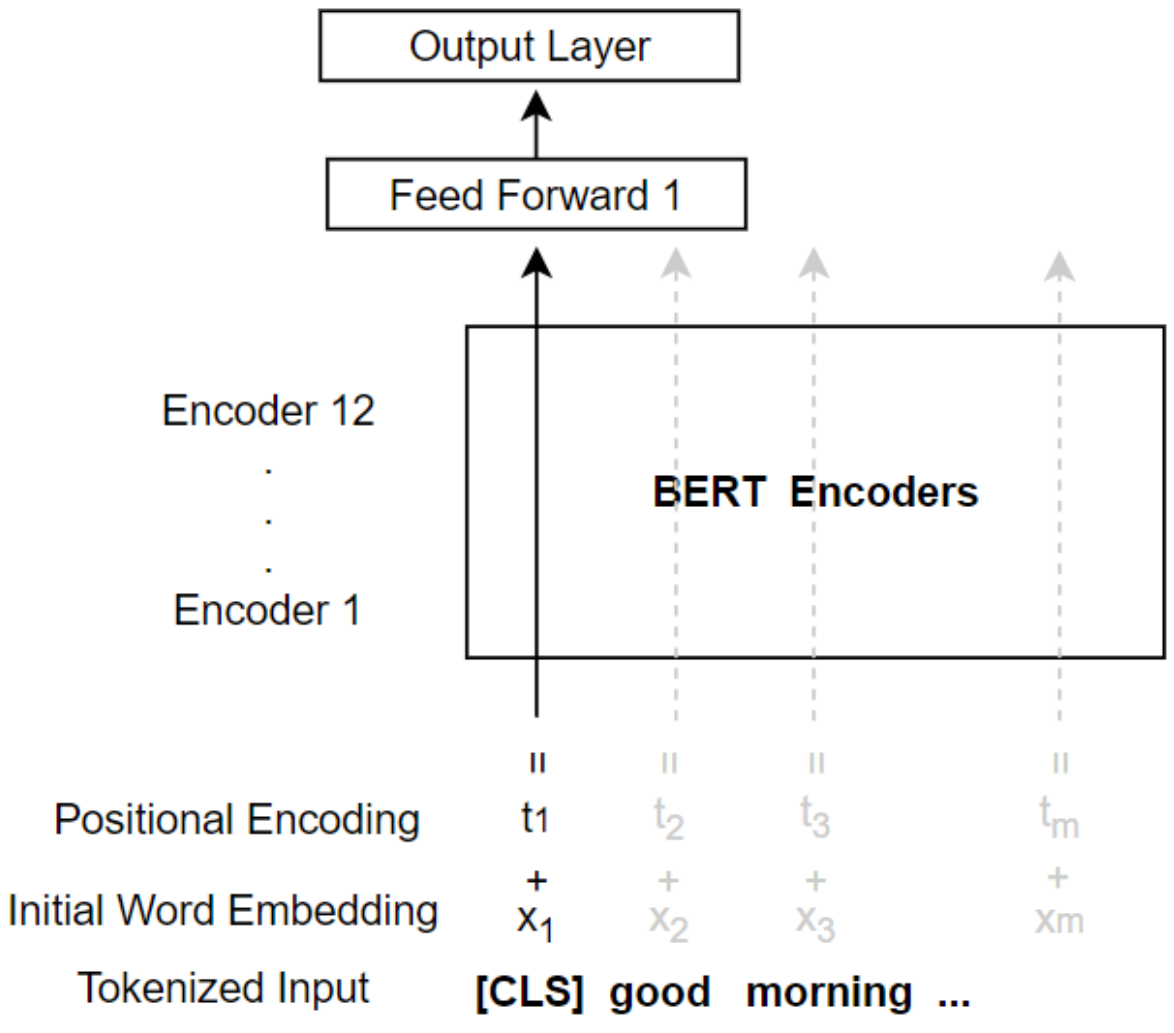

*Figure B1: BERT Framework*

The token embedding is combined with a positional encoder $t_m$ that codes the position of the token in the sentence using sine and cosine functions (see Devlin et al. (2018) for details).

[3] We use the BERT-base-uncased model (that converts all words to lower case and removes accent markers) as compared to the cased model, because the uncased model is known to typically perform better unless the goal is to study case specific contexts such as 'named entity recognition' and 'part-of-speech tagging.'

This is passed through a set of 12 encoders arranged sequentially. The output of the 'CLS' token is passed through the feed forward 1 layer that is initialized with pre-trained weights from the next sentence prediction task, and has a *tanh* activation function. We follow this up with an output layer that connects with either of our three continuous outcomes or binary outcome. We then fine-tune the entire model with all hierarchical layers over our data sample.

The Encoders contain the self-attention heads (explained further ahead) which help the model capture the relative importance between word-pieces while forming an association with the outcome of interest. By virtue of being pre-trained to capture contextual usage of words, the model is able to make better decisions on assigning relative attention (importance) weights to different word-pieces (tokens) in our sample. Word-pieces that receive more attention play a more important role (in either a positive or negative direction) in predicting the outcome of interest. We analyze these attention weights during ex-post interpretation.

### B.2 BERT Encoders

BERT Encoders comprise a set of 12 sequentially arranged identical encoders, and we illustrate the architecture of one encoder in Figure B2. [4]

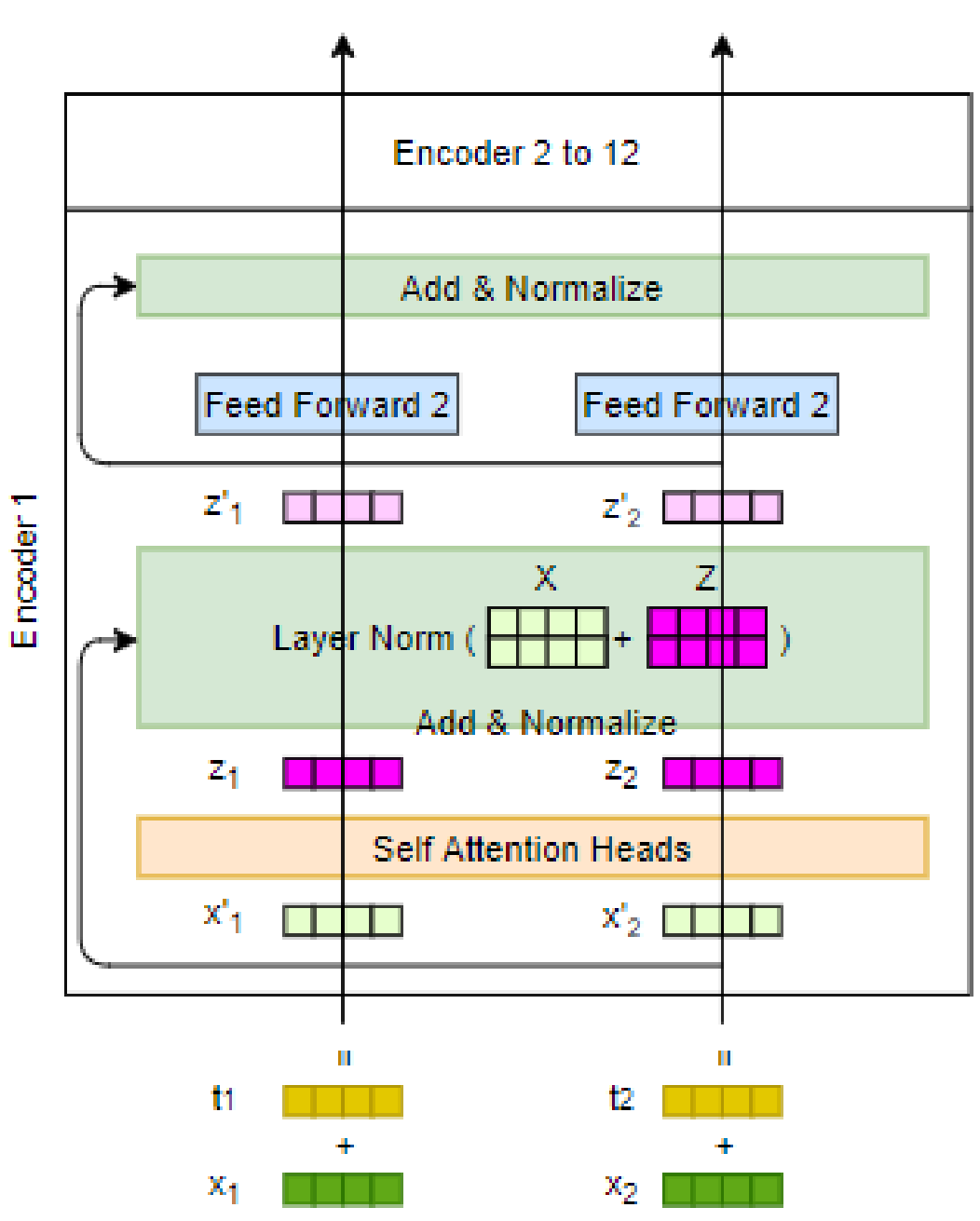

*Figure B2: Encoders*

[4] Our figures are inspired by the work of Jay Alammar (see Alammar (2018) for more details).

We use an example with a sentence that has only two tokens, and this can be extended to any example that has a maximum of 512 tokens, which is the maximum limit of the pre-trained BERT model. The combined vector of the initial token embedding $(x_1, x_2)$ and positional encoding $(t_1, t_2)$ results in the vectors $(x_1', x_2')$ that are passed through self-attention heads which incorporate information of other relevant tokens into the focal tokens. The architecture of the self-attention head is explained further ahead. The outputs of the self-attention head $(z_1, z_2)$ are then added with the original input $(x_1', x_2')$ using a residual connection (shown with a curved arrow) and normalized (using mean and variance). The outputs $(z'_1, z'_2)$ are passed through identical feed forward networks that have a GELU (Gaussian Error Linear Unit) activation function, i.e. $gelu(x) = 0.5x\left(1 + erf\left(\frac{x}{\sqrt{2}}\right)\right)$. The gelu activation combines the advantages of the ReLU (Rectified Linear Unit) non-linearity (i.e., $relu(x) = max(0, x)$) with dropout regularization. The outputs of the feed forward network are added with the inputs $(z'_1, z'_2)$ using a residual connection and normalized again before being fed to the next encoder in sequence. In addition, each sub-layer is first followed by a dropout probability of 0.1 before being added and normalized.

### B.3 BERT Attention Mechanism: Scaled Dot-Product Attention

Next, we explain the scaled dot-product attention mechanism within the self-attention heads illustrated in Figure B3.

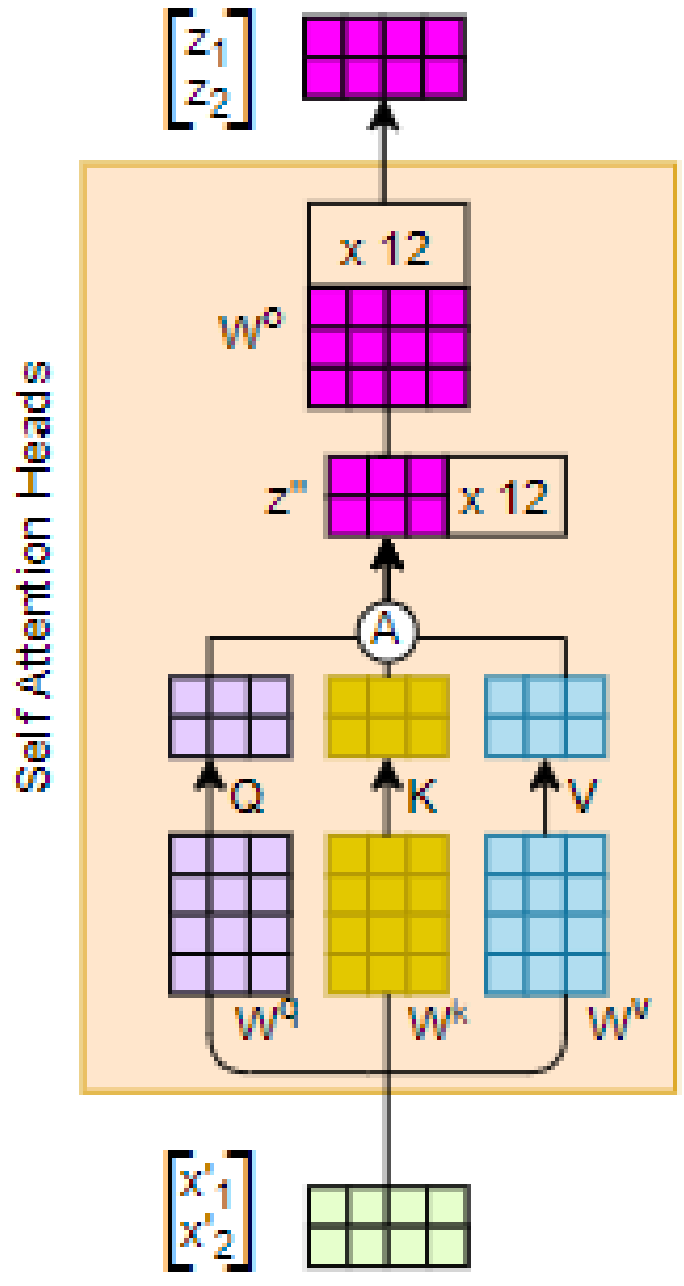

*Figure B3: Self-Attention Heads*

There are 12 self-attention heads that capture the contextual information of each token in relation to all other tokens used in the text. In other words, the self-attention heads allow the model to identify and weigh all other tokens in the text that are important when learning the vector representation of the focal token. We use them to measure the importance (strength) of association between the tokens in the text and the outcome of interest.

The inputs $(x_1', x_2')$ are concatenated and multiplied with three weight matrices, $W^q, W^k$ and $W^v$ (that are fine-tuned during model training) to get three vectors – $Q$ (Query), $K$(Key) and $V$(Value). These three vectors are combined by taking a dot product using an attention function (A):

$$A(Q, K, V) = z_0'' = softmax\left(\frac{Q.K^T}{\sqrt{d_k}}\right).V$$

where, $d_k$, the dimension of the Key vector, is 64 and is equal to the dimensions of the other two vectors $d_q$ and $d_v$; and $softmax(x) = \frac{e^{x_i}}{\sum_{i=1}^{m} e^{x_i}}$. The scaling (division) by $\sqrt{d_k}$ is performed to ensure stable gradients. The computation of $z_0''$ is for one attention head, and this is carried out in parallel for 11 additional attention heads to give us 12 vectors, $z_0''$ ... $z_{12}''$, which are concatenated to produce $z''$. This is multiplied with a weight vector $W^O$ (which is fine-tuned during model training) to produce output $(z_1, z_2)$. The use of 11 additional attention heads allows the model to capture more complex contextual information.

In order to capture the estimated attention weights, we average the output across all the attention heads in the last encoder of the BERT model, which results in an attention vector of dimension <n, $k$, $k$> where n is the number of videos used in the analysis, and <$k$,$k$> corresponds to $k$ weights for $k$ tokens, where $k$ equals the maximum number of tokens (word-pieces) for a covariate type – title, description (first 160 characters) or captions/transcript (beginning, middle or end). As mentioned earlier, the first token for each example is the 'CLS' or classification token. We are interested in the attention weights corresponding to this token because the output from this token goes to the output layer (as shown earlier in Figure B1). Thus, we get at an attention weight vector of dimension <n, $k$>, where each observation has $k$ weights corresponding to the 'CLS' token. We exclude 'CLS','SEP' and any token used for padding short sentences), during ex-post interpretation.

## Online Appendix C – Operationalization of Audio Model

We provide an overview of the implementation of our Audio Model – YAMNet +Bi-LSTM+ Attention Mechanism, and explain the architecture of MobileNet v1 and Bi-LSTM with attention mechanism.

### C.1 YAMNet +Bi-LSTM+ Additive Attention Mechanism

We analyze audio data using the pre-trained YAMNet model (Pilakal & Ellis, 2020), and customize it with an additional Bidirectional LSTM (Bi-LSTM) layer and an attention mechanism as shown in the framework in Figure C1.

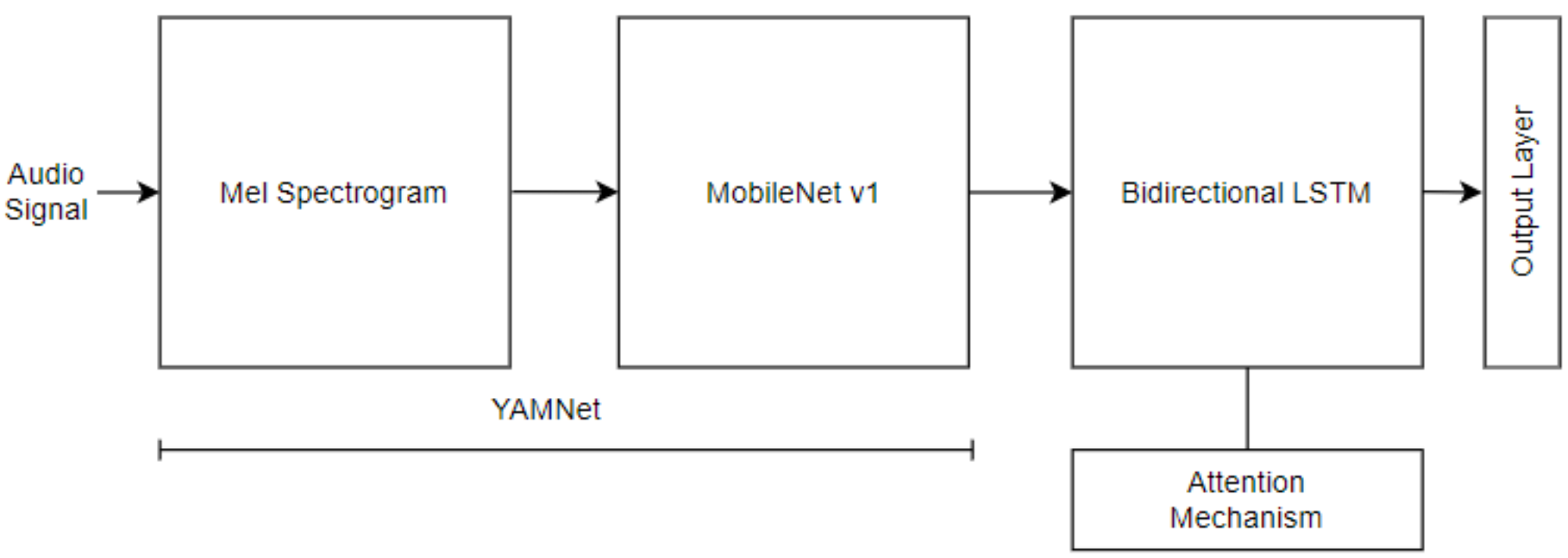

*Figure C1: Audio Model Framework*

Each audio signal is a 30 second clip, which we resample at 16,000 Hz and mono sound (for consistency), and this results in 480,000 data points for each clip. Note that for a few influencer videos that are shorter than 30 seconds, we append it with moments of silence to make the audio length consistent across our sample. To summarize the large number of data points, the YAMNet model first generates a Mel spectrogram that spans the frequency range of 125 to 7500Hz (note that the 2000-5000 Hz range is most sensitive to human hearing (Widex, 2016)) over which the YAMNet model has been pre-trained. The spectrogram uses the pre-trained Short-Term Fourier Transform window length of 25ms with a hop size of 10ms that results in a 2998 x 64 (time steps x frequency) vector corresponding to 30 seconds of each audio clip. This corresponds to 64 equally spaced Mel bins on the log scale, such that sounds of equal distance on the scale also sound equally spaced to the human ear. The model then passes each segment of 960ms from the spectrogram output, i.e., 96 frames of 10ms each with overlapping patches (to avoid losing information at the edges of each patch) as input to the MobileNet v1 architecture.

The size of the overlap or hop size is 490ms, which results in a total of 60 moments for each 30 second audio clip.

The MobileNet v1 processes the spectrogram through multiple mobile convolutions and returns audio class predictions for each of the 60 moments in the clip. This comprise a total of 521 different audio classes such as speech, music, animal, etc. (corresponding to each 960 ms segment) over which the model has been pre-trained. Pilakal and Ellis (2020) remove 6 audio classes (viz. gendered versions of *speech* and *singing*; *battle cry*; and *funny music*) from the original set of 527 audio classes to avoid potentially offensive mislabeling.

We then pass the <521x60 > dimensional vector as input to the Bi-Directional LSTM layer with an attention mechanism. We make this layer bidirectional to allow it to capture the interdependence between sequential audio moments from both directions. For example, the interdependence between the sound of a musical instrument at 5 seconds and the beginning of human speech at 15 seconds can be captured by the model bidirectionally. We adapt the attention mechanism used for neural machine translation by Bahdanau et al. (2014) (explained further ahead) to help the Bi-LSTM layer capture the relative importance between sound moments in order to form an association with an outcome. These measures of relative importance (attention) can be understood similarly as the attention weights in the Text model, and are analyzed during ex-post interpretation. We pass the output of the Bi-LSTM (with attention mechanism) through an output layer which connects with either of our three continuous outcomes or binary outcome. We then fine-tune the 'Bi-LSTM with attention mechanism' over our data sample.

### C.2 MobileNet v1 architecture

The MobileNet v1 architecture is illustrated in detail in Table C1 (Howard et al., 2017). Each row describes Stage $i$ with input dimension $[\widehat{H}_i, \widehat{W}_i]$ (resolution), output channels $\hat{C}_i$ and $\hat{L}_i$ layers (depth). Stage 1 has a regular convolution operation, whereas Stage 2 to 10 have the Mobile Convolution which is the main building block of the architecture. It is represented as "MConv, $k$ x $k$, s" where $k$ x $k = 3$ x 3 is the size of the kernel and $s = \{1,2\}$ is the stride. MConv divides the regular convolution operation into two steps – depth wise separable convolutions and point wise convolution, thus increasing the speed of computation (see Howard et al. (2017) for details). Stage 11 has a Global Average Pooling Layer that averages the inputs along its height and width and passes its output to Stage 12 which is a Dense output layer with 521 logistic functions that give the per class probability score corresponding to the 960 ms input segment. As mentioned earlier, we use a hop size of 490 ms so that we get an even number of 60 time step predictions corresponding to the 30 seconds of input. The resulting output vector has a dimension of 521x60 (audio classes x time steps) for each 30 second clip.

| **Stage** $i$ | **Operator** $\hat{F}_i$ | **Input Resolution** ($\hat{H}_i$ x $\hat{W}_i$) | **Output Channels** $\hat{C}_i$ | **Depth** $\hat{L}_i$ (Layers) | **Pre-trained Weights** |
|---|---|---|---|---|---|
| 1 | Conv, k3x3, s2 | 96 x 64 | 32 | 1 | Yes |
| 2 | MConv, k3x3, s1 | 48 x 32 | 64 | 1 | |
| 3 | MConv, k3x3, s2 | 48 x 32 | 64 | 1 | |
| 4 | MConv, k3x3, s1 | 24 x 16 | 128 | 1 | |
| 5 | MConv, k3x3, s2 | 24 x 16 | 128 | 1 | |
| 6 | MConv, k3x3, s1 | 12 x 8 | 256 | 1 | |
| 7 | MConv, k3x3, s2 | 12 x 8 | 256 | 1 | |
| 8 | MConv, k3x3, s1 | 6 x 4 | 512 | 5 | |
| 9 | MConv, k3x3, s2 | 6 x 4 | 512 | 1 | |
| 10 | MConv, k3x3, s2 | 3 x 2 | 1024 | 1 | |
| 11 | Global Average Pooling | 3 x 2 | 1024 | 1 | |
| 12 | Dense | 1 x 1 | 521 | 1 | |

*Table C1: MobileNet-v1 architecture*

### C.3 Bi-LSTM with Additive Attention

The output from MobileNet v1 is passed as input to the Bi-LSTM with (additive) attention mechanism, shown in Figure C2.

We use two layers of LSTM cells – the first layer is a 32-unit Bidirectional LSTM layer and the second layer is a 64-unit (unidirectional) LSTM layer. They are separated by an additive attention mechanism as shown in the figure. Each audio segment $x_m$ <521,1>, where $m$ is the total number of moments (time steps), is passed as input to each cell of the Bidirectional LSTM layer. This layer is made bidirectional to allow it to capture the interdependence between sequential audio segments from both directions. The sequential nature of LSTM cells in a layer allow the model to capture dependencies between audio segments that are separated from each other (see the LSTM paper by Gers et al. (1999) for more details). We adopt the additive attention mechanism used for neural machine translation by Bahdanau et al. (2014) to help the Bi-LSTM model focus on more important parts of the input. The mechanism weighs the output activations ($a^{<t>} = [\vec{a}^{<t>}, \overleftarrow{a}^{<t>}], t = 1$ to $m$) from each cell of the pre-attention Bi-LSTM layer before passing the contextual output, $c^{<t>}$, to the post-attention LSTM layer above it. In addition, each cell of the attention mechanism takes as input the output activation $s(t-1)$ from each preceding cell of the post-attention LSTM layer which allows it to factor in the cumulative information learnt by the model till that time step (see Bahdanau et al. (2014) for more details on the attention mechanism). The output of the last cell in the post-attention LSTM layer is passed to an output layer which has a linear activation function for the three continuous outcomes and a sigmoid activation function for the binary outcome. The context vector $c^{<m>}$ from the last cell of the attention mechanism allows measurement of the relative weights placed by the model along the time dimension of the input in order to form an association with the outcome of interest. Audio moments that have higher weight are more important while forming an association between the audio clip and the outcome.

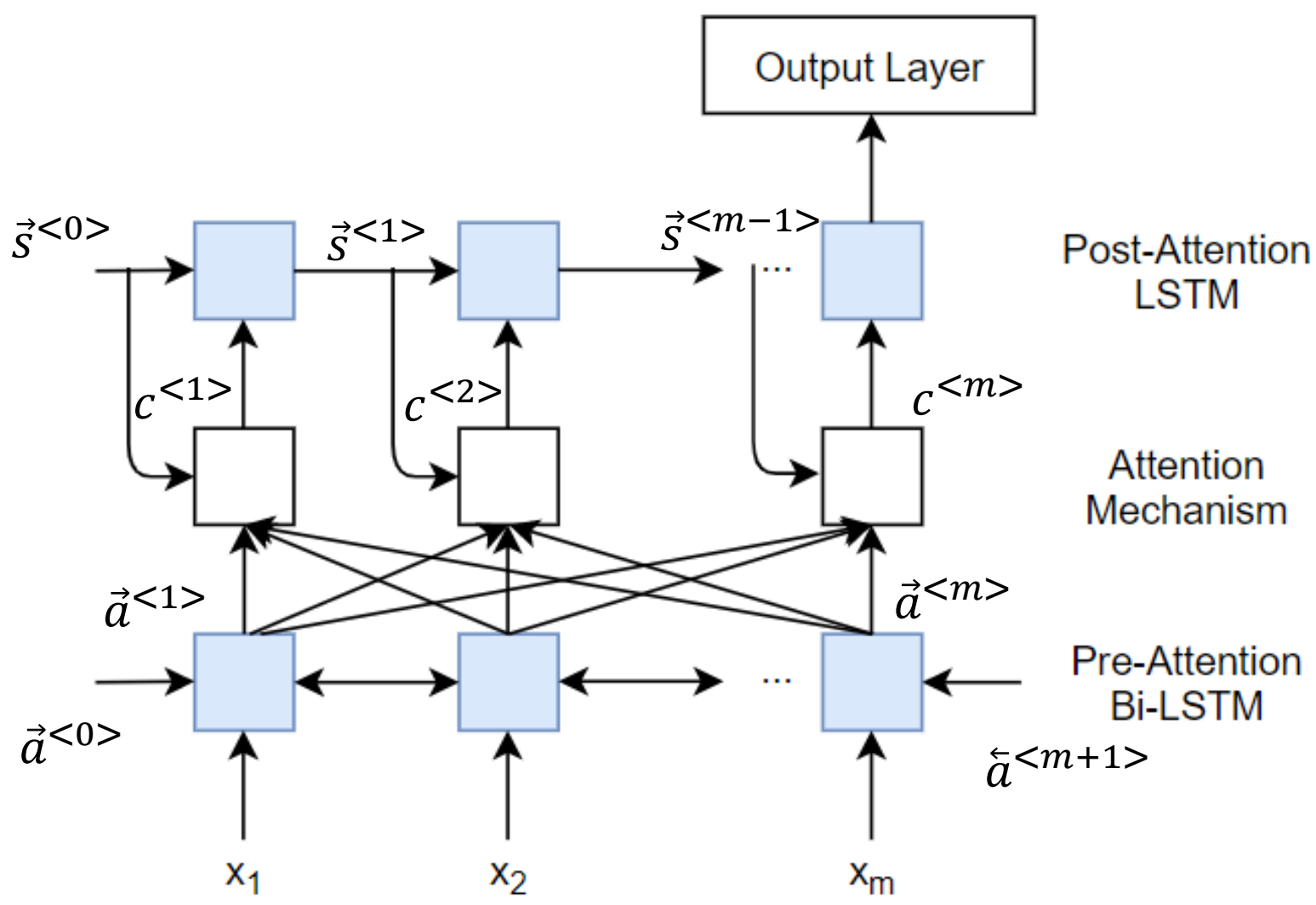

*Figure C2: Bi-LSTM with Additive Attention*

**Online Appendix D – Operationalization of Video Image Model**

We provide an overview of the implementation of our Video Image Model – VGG-16 +Bi-LSTM. We then explain the architecture of VGG-16, the architecture to combine information from all video frames, and our approach to find gradient-based attention (saliency).

**D.1 VGG-16 + Bi-LSTM**

We use VGG-16 to analyze thumbnail images and a combination of VGG-16 and Bi-LSTM to analyze video frames.

First, we pass thumbnail images as input to one VGG-16, and then finetune its final layers to capture relationship with each outcome. We provide details on the VGG-16 architecture in Section D.2.

Next, we illustrate in Figure D1 our framework to analyze video frames. We pass each video image frame $i = 1$ to $m$, where $m$ has a value of 30 frames, through a VGG-16 architecture. Note that for a few influencer videos that are shorter than 30 seconds, we append it with black frames to make the video image length consistent across our sample. Our sampling rate of one frame per second (30 frames in 30 seconds) in conjunction with the size of our data sample (1620 videos) ensures that our model is feasible to analyze.

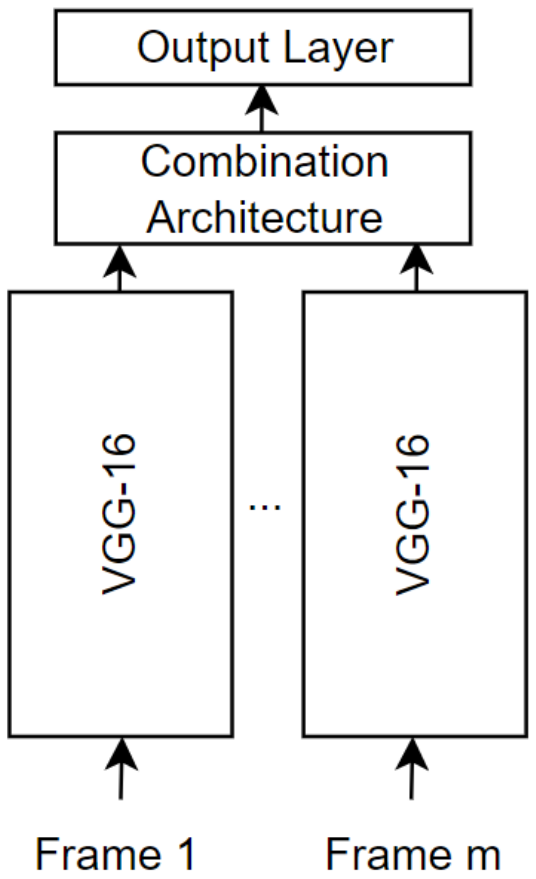

*Figure D1: Framework to analyze Video Frames*

In the last step, we combine the outputs from each VGG-16 model. Our combination architecture comprises the Bi-LSTM, known to be one of the best performing architectures at capturing sequential information from video frames (c.f. Yue-Hei Ng et al. (2015)). The output of the combination architecture is passed through an output layer which connects with either of our three continuous outcomes or binary outcome.

### D.2 VGG-16 Architecture

VGG-16 has 16 layers whose parameters can be learned (Simonyan & Zisserman, 2014). The architecture of VGG-16, customized to our input dimension of 135 x 240 x 3 (where 3 corresponds to pixel intensities for Red, Green and Blue channels), is shown in Table D1. Each row describes Stage $i$ with input dimension $[\hat{H}_i, \hat{W}_i]$ (resolution), output channels $\hat{C}_i$ and $\hat{L}_i$ layers (depth).

| **Stage** $i$ | **Operator** $\hat{F}_i$ | **Input Resolution** ($\hat{H}_i$ x $\hat{W}_i$) | **Output Channels** $\hat{C}_i$ | **Depth** $\hat{L}_i$ (Layers) | **Pre-trained ImageNet weights** |
|---|---|---|---|---|---|
| 1 | Conv, k3x3, s1 | 135 x 240 | 64 | 2 | Yes |
| 2 | Max Pooling, k2x2, s2 | 135 x 240 | 64 | 1 | |
| 3 | Conv, k3x3, s1 | 67 x 120 | 128 | 2 | |
| 4 | Max Pooling, k2x2, s2 | 67 x 120 | 128 | 1 | |
| 5 | Conv, k3x3, s1 | 33 x 60 | 256 | 3 | |
| 6 | Max Pooling, k2x2, s2 | 33 x 60 | 256 | 1 | |
| 7 | Conv, k3x3, s1 | 16 x 30 | 512 | 3 | |
| 8 | Max Pooling, k2x2, s2 | 16 x 30 | 512 | 1 | |
| 9 | Conv, k3x3, s1 | 8 x 15 | 512 | 3 | |
| 10 | Max Pooling, k2x2, s2 | 8 x 15 | 512 | 1 | |
| 11 | Global Average Pooling | 4 x 7 | 512 | 1 | No |
| 12 | Dense | 1 x 1 | 1 | 1 | |

*Table D1: VGG-16 architecture*

Stages 1, 3, 5, 7 and 9 have convolution operations, whereas Stages 2, 4, 6, 8 and 10 have the max pooling operation, where $k$ x $k = 3$ x 3 is the size of the kernel and $s = \{1,2\}$ is the stride. Stage 11 has a Global Average Pooling Layer that averages the inputs along its height and width and passes its output to Stage 12 which is a Dense layer.

### D.3 Combination Architecture

To analyze video frames ($i = 1$ to $m$, where m has a maximum value of 30) we use the Bi-LSTM architecture that captures sequential information across different video frames. This is illustrated in Figure D2. Each VGG-16 architecture takes a unique video frame as input and provides the output from Stage 10 to the Global Average Pooling (GAP) Layer. This is followed by Dense Middle Layers (that use ReLU activation for continuous outcomes and sigmoid activation for the binary outcome), which is followed by a single Bi-LSTM layer with 256 memory cells, and finally a Dense output layer (that uses linear activation for continuous outcomes and softmax activation for the binary outcome). We use the pre-trained ImageNet weights (Krizhevsky et al., 2012) from Stage 1 to 10 in each VGG-16 architecture and then we finetune the weights of the top layers.

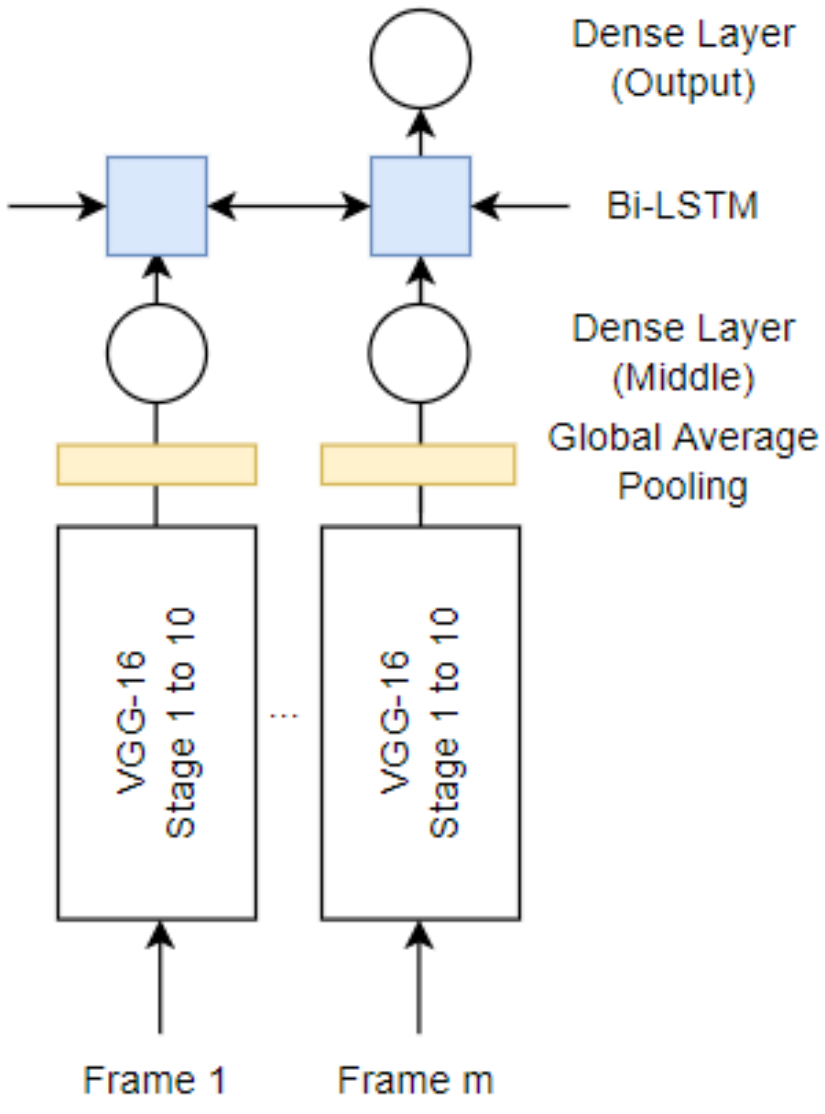

*Figure D2: Bi-LSTM*

### D.4 Gradient-based attention (saliency)

As our VGG-16 model uses pretrained weights from the ImageNet classification task, the lower layers of the model are well trained to detect basic attributes of objects in images. This helps the model differentiate between various objects in the images of our video sample during the process of finetuning. After the end of model training, we ex-post identify the salient parts of images that are associated with an outcome through gradient-based activation maps (cf. Selvaraju et al., 2017). We find gradients by taking the derivative between the predicted continuous outcome (or class of predicted binary outcome) and the output of the activation layer after the last convolution layer in each VGG-16 architecture that processes one video frame. However, unlike Selvaraju et al. (2017), we do not apply the ReLU (Rectified Linear Unit) activation on the gradient values as we would like to retain negative gradient values for interpretation. Hence our approach is a modified gradient-based activation map that is suitable for our setting. Areas of the image with positive (negative) gradients correspond to regions that are positively (negatively) associated with continuous outcomes and the predicted class of the binary outcome.

## Online Appendix E – Details on Multimodal Approach

### E.1 Embeddings and Model Design

We use the state-of-the-art individual models discussed in Section 5.1 to generate embeddings from unstructured data. This makes our multimodal approach comparable with our combined model (equation (1)) that uses the predictions from the individual models. Specifically, we generate embeddings from text data using the second-to-last encoder of the pre-trained BERT model (Devlin et al., 2018; Xiao, 2018), embeddings from audio data using the predictions from the pre-trained YAMNet model (Pilakal & Ellis, 2020), and embeddings from video image data using the global average pooling layer of the pre-trained VGG-16 model (Simonyan & Zisserman, 2014). Specifically, these embeddings are generated in two steps: (a) we first generate an embedding from each 30-second segment, and (b) average the embedding over the length of each 30 second segment to get a unique vector for each segment. Hence, for each 30-second segment, we get a text embedding (averaged from the CLS to SEP token) of dimension (n,1,768), audio embedding (averaged over each patch of 960 ms with hop size of 490ms) of dimension (n,1,521) and image embedding (averaged over each frame at one frame per second) of dimension (n,1,512), where n is the total number of videos in our sample.

Our multimodal architecture (shown earlier in Figure 5, Section 5.3) is inspired from the work of Ghosal et al. (2018) and Wu et al. (2014). We borrow the relevant aspects of their architecture and parameter choices to our setting. Our architecture takes as input the text embeddings $u_{1m}$ of captions/transcript, audio embeddings $u_{2m}$ and image embeddings of video frames $u_{3m}$, which are of dimensions (n,m,768), (n,m,521) and (n,m,512) respectively, where n is the number of videos in the sample and m is the number of 30-sec segments. In addition, we supply as input the set of structured features $X_{it}$ (listed earlier in Table 3, Section 4.3), text embeddings for title (n,1,768), text embeddings for description (first 160 characters) (n,1,768) and image embeddings for thumbnail (n,1,512). Note that we scale each set of embeddings by their respective $L^2$ norm (before supplying them as input to our multimodal model) so that their relative influence is comparable. We also do this same scaling for the structured features.

We pass the set of embeddings $u_{1m}$, $u_{2m}$ and $u_{3m}$ through separate Bi-Direction GRU layers which capture the sequential interdependence (within-modality interaction) between different 30-sec clips of each modality (Ghosal et al., 2018). This is followed by a sequence of dense time-distributed layers to further capture the sequential dependence from the output of the preceding layer (Ghosal et al., 2018). We follow this up with a dense transformation layer to further process the information from the preceding layer (Wu et al., 2014). Similarly, the input from the structured features and the incentive to click data (title, description and thumbnail) are passed through separate dense transformation layers to process the data (Wu et al., 2014). Hence, we have a total of seven dense transformation layers and each of them are followed by a global average pooling layer and the outputs of the seven layers are then concatenated (Ghosal et al., 2018). This is then supplied as input to a dense fusion layer that captures cross-modal interactions between all the preceding layers (Wu et al., 2014), which is then finally connected to an output layer.

**E.2 Results of Multimodal Model for Relative Importance of Position in Video**

Similar to our approach in Section 7.1.3 and Table 9, we test whether embeddings from the beginning 30 seconds (across text (captions/transcript), audio and video images (video frames)) are significantly more important in predicting engagement than embeddings from the middle and end 30 seconds. In order to do so, we permute the embeddings from the beginning 30 seconds in the holdout sample while holding all other features constant, and measure the change in prediction error. We repeat this process for the middle and end 30s. Subsequently, we compare the difference in prediction errors between the permuted positional embeddings across 50 bootstrap iterations using a t-test and show the difference in means in Table E1.

| | Verbal Engagement | | Non-Verbal Engagement | |
|---|---|---|---|---|
| | Sentiment (VS) | Level (VL) | Sentiment (NVS) | Level (NVL) |
| Permuted Prediction Error: Begin 30s | 0.57 | 0.96 | 0.97 | 0.77 |
| Permuted Prediction Error: Middle 30s | 0.65 | 0.86 | 0.93 | 0.73 |
| Permuted Prediction Error: End 30s | 0.67 | 0.86 | 0.95 | 0.72 |
| Difference in Permuted Prediction Error: Begin 30s vs Middle 30s | (-)0.08*** | 0.10*** | 0.04*** | 0.04*** |
| Difference in Permuted Prediction Error: Begin 30s vs End 30s | (-)0.09*** | 0.10*** | 0.02*** | 0.04*** |
| Note: ***p<0.001 | | | | |

*Table E1: Results of Multimodal Model for Position (Embeddings from Begin, Middle and End 30s) (Accuracy for binary outcome - VS; RMSE for continuous outcomes – VL, NVS, NVL)*

We find that the beginning 30s is significantly more important than middle or end 30s (*as there is an increase in RMSE or decrease in accuracy*) while predicting each of our engagement measures, consistent with the findings for the Combined Model in Table 8, Section 7.1.2. This demonstrates that our findings hold true even when we account for interactions between modalities during the training process. Note that the difference in means between middle 30s and end 30s is minimal, and hence we do not report those results.

## Online Appendix F – Feature Generation of Theory-based Features

We generate theory-based features with the help of trusted databases, reliable transfer learned models and accurate APIs, which is more efficient compared to employing human coders.

### F.1 Presence of Text Features

Viewers generally prefer ads with high entertainment value and low informational content (Woltman Elpers et al., 2003), and similarly, influencer videos are watched for entertainment (Landsberg, 2021). Brand mentions (in captions/transcript) may lower entertainment value and increase informational content, which could reduce the persuasiveness of the message and decrease (positive) sentiment (Teixeira et al., 2010; Tellis et al., 2019). However, an alternate view suggests that increase in informational value may complement entertainment, educate viewers and build trust, potentially increasing (positive) sentiment (Leung et al., 2022).

The use of emotional words has been associated with higher engagement, such as shares, comments, and likes, due to their connection to brand personality and their ability to evoke high arousal (Berger & Milkman, 2012; Lee et al., 2018). However, emotional language can also be perceived as insincere, particularly in fake news (Bakir & McStay, 2018; Guo et al., 2019).

Given these contrasting set of findings in prior literature on the effect of brand mentions and emotional words, their association with the sentiment for influencer videos on YouTube remains unclear.

We identify brand names and emotion words in captions/transcript in 30 seconds of the beginning, middle and end of a video as follows. We apply *regular expressions* on textual data while relying on a master list of words. The master list of brand names comprise the Top 100 Global brands in 2019 that were obtained from BrandZ, Fortune100 and Interbrand. We then add more than 32,000 brands (with US offices) from the Winmo database to this list. This is further combined with brand names identified by applying Google's Vision API – Brand Logo Detection on video frames (one fps in 30 seconds of beginning, middle and end) in our sample. From this combined list, we remove more than 800 generic brand names such as 'Slice,' 'Basic,' 'Promise,' etc. that are likely to be used in non-brand related contexts. The master list of emotional words is obtained from the list of 2,469 emotional words identified in the LIWC dictionary (Berger & Milkman, 2012; Pennebaker et al., 2015). We identify brand and emotion mentions within 305,554 word-pieces (that are generated by the BERT model) in captions/transcript across all parts (beginning, middle and end) of the 1620 videos in our sample.

We find that 28% (80%) of videos have a brand (emotion) mentioned at least once (in any part of the video).

F.2 Duration of Audio Features

Music in ads can reduce irritation and enhance brand memory, especially when used without voiceovers (Alexomanolaki et al., 2007; Pelsmacker & Van den Bergh, 1999). Thus, longer durations of music without voiceover may be associated with increased (positive) sentiment.

Fast speakers are often perceived as more knowledgeable and effective (Peterson et al., 1995), and faster speech rates in educational videos have been linked to increased watch time (Guo et al., 2014). However, anecdotal evidence from linguists suggests that changes in pace, rather than rapid speech, are what typically capture attention (Beck, 2015; Jennings, 2021). The effect of speech rate on sentiment for influencer videos on YouTube is therefore uncertain.

We identify music and human speech in 30 seconds of the beginning, middle and end of a video as follows. The YAMNet model (Mel Spectrogram + MobileNet v1) finds the predicted probability of each moment[5] of a 30 second audio clip (which has 60 moments) belonging to a sound class. Our model can efficiently accomplish this identification for 291,600 moments (1620 videos x 3 parts x 60 moments in a part). We combine the identified sound classes into different categories based on the AudioSet ontology (Gemmeke et al., 2017) – *Human* (86%), *Music* (83%), *Animal* (17%), and *Others* which include sounds of silence, things, ambiguous sounds, background sounds and natural sounds. The percentage in brackets indicates the percentage of videos that contain a sound of that category (with probability> 0.5) in any part of the video. Note that a moment of sound can be classified into multiple categories if sounds from more than one category occur together.

F.3 Size of Video Image Features

Human faces are generally preferred in print ads and are linked to increased engagement on social media (Li & Xie, 2020; Xiao & Ding, 2014; Yang et al., 2025), possibly due to human desire for social interaction (To & Patrick, 2021). In influencer videos, the influencer is often demonstrating something to a viewer with the help of their hands which can further aid in engaging the viewer. Hence, we can expect the presence and size of the whole human (influencer) in the video, including their hands, to enhance (positive) sentiment. Conversely, images of brand packaging, unlike human faces, tend to decrease engagement, as they are less likely to elicit a desire to interact (Hartmann et al., 2021). Thus, larger images of packaged goods in videos may lead to reduced (positive) sentiment. However, like humans, animals in ads are often more likable and have lower irritation scores, suggesting that larger images of animals may increase (positive) sentiment (Biel & Bridgwater, 1990; Pelsmacker & Van den Bergh, 1999).

[5] Each moment is 960ms long, and the subsequent moment begins after a hop of 490ms. Hence, each 30 second audio clip encompasses 60 moments (details in Online Appendix C).

Facial expressions like joy and surprise in ads can hold attention and retain viewers (Teixeira et al., 2012). However, their impact on social media engagement is mixed, with happy faces sometimes linked to fewer retweets and no effect on likes (Li & Xie, 2020). Similarly, surprise in educational videos shows mixed effects on watch time depending on context (Zhou et al., 2021), making the impact of facial expressions on sentiment in influencer videos on YouTube ambiguous.

We focus on the following features: size of humans, size of face of humans, size of packaged goods, size of animals, and human facial expressions of joy and surprise. In addition, we also study the role of brand logos, and control for the size of everyday objects such as clothes & accessories and home & kitchen items. We study these features in 30 frames that lie in 30 seconds of the beginning, middle and end of a video. We identify these features using the Cloud Vision API from Google, that has been pre-trained on millions of images and whose high accuracy has been validated in prior academic and industry research (FileStack, 2019; Li & Xie, 2020; Szegedy et al., 2016). We can efficiently implement this API over 145,800 frames (1620 video x 3 parts of video x 30 frames per part). The API returns the vertices of the identified object which allows us to create a rectangular bounding box to define its area. We divide the identified objects into eight different categories – *Humans* (91%), *Faces* (86%), *Animals* (27%), *Brand Logos* (16%), *Packaged Goods* (28%), *Clothes & Accessories* (86%), *Home & Kitchen* (58%), and *Other Objects*. The percentage in brackets indicate the percentage of videos that contain an object of that category (in any part of the video). The API also returns the level of surprise or joy in each face that was detected: {-2: very unlikely, -1: unlikely, 0: possible, 1: likely, 2: very likely}.

## Online Appendix G – Comparison of Model Performance

We compare the predictive performance of our individual models for Text, Audio and Video Images with standard benchmarks used in the marketing literature to demonstrate that our models do not compromise on predictive ability.[6] To make this comparison, we choose the two outcome measures of verbal engagement: VL (continuous outcome) and VS (binary outcome). For the unstructured covariates, we focus on the beginning 30 seconds of video data (Tables G1, G2 and G3) in addition to title (Table G1), description (first 160 characters) (Table G1) and thumbnails (Table G3).

We first compare the predictive performance of the Text Model (BERT) with four standard models in Table G1. These standard benchmarks include an LSTM (with a 300 dimensional Glove word vector embedding), CNN model (Liu et al., 2019), CNN-LSTM (Chakraborty et al., 2022) and CNN-Bi-LSTM. As can be seen in Table G1, BERT performs better than the benchmarks while predicting both VL (lowest RMSE) and VS (highest accuracy). This is because of primarily three reasons: (1) BERT learns contextual embeddings (e.g., the embedding for the word 'bark' will change based on the context in which it used – a dog's bark or tree's outer layer) as compared to benchmarks methods which use fixed word embeddings such as Glove and word2vec (2) The entire BERT model with hierarchical layers is pre-trained whereas conventional models only initialize the first layer with a word embedding (3) BERT uses a self-attention mechanism that avoids loss of information as compared to a sequential process (such as in an LSTM) that can lead to loss of information.

| Outcome | Covariate | LSTM | CNN | CNN-LSTM | CNN-Bi-LSTM | BERT |
|---|---|---|---|---|---|---|
| Verbal Engagement Level (VL) | Title | 0.99 | 0.91 | 0.90 | 0.88 | 0.83 |
| | Description (first 160c) | 0.98 | 0.93 | 0.89 | 0.88 | 0.87 |
| | Captions/transcript (beginning 30s) | 0.97 | 1.04 | 0.94 | 0.93 | 0.92 |
| Verbal Engagement Sentiment (VS) | Title | 0.67 | 0.70 | 0.70 | 0.70 | 0.72 |
| | Description (first 160c) | 0.50 | 0.69 | 0.69 | 0.69 | 0.70 |
| | Captions/transcript (beginning 30s) | 0.67 | 0.70 | 0.70 | 0.70 | 0.71 |

*Table G1: Comparison of Text Model predictive performance on holdout sample for measures of verbal engagement (RMSE for VL; Accuracy for VS)*

[6] We run all the standard benchmark models at least three times and average their prediction errors. We run our main models for Text and Audio (and Video Images) 50 (and 25) times and average their prediction errors.

We now compare the model performance of the Audio model with other variants that are devoid of transfer learning and attention, and present the results in Table G2. First, we find that the addition of the transfer learned MobileNet v1 to a model that uses only the 'Mel Spectrogram + Bi-LSTM' results in improved RMSE when predicting VL and improved accuracy when predicting VS. Addition of the attention mechanism to this results in our Audio model. However, this does not improve the prediction error but also does not deteriorate it. Overall, our Audio model performs the best in out-of-sample prediction, thus demonstrating that transfer learning (via MobileNet v1) contributes towards the predictive ability of the model. Also, capturing relative attention weights (via the attention mechanism) while contributing towards interpretability does not deteriorate predictive ability.

| Outcome | Covariate | Mel Spectrogram + Bi-LSTM | Mel Spectrogram + MobileNet v1 + Bi-LSTM | Audio Model: Mel Spectrogram + MobileNet v1 + Bi-LSTM + Attention |
|---|---|---|---|---|
| Verbal Engagement Level (VL) | Audio (first 30s) | 0.95 | 0.92 | 0.92 |
| Verbal Engagement Sentiment (VS) | Audio (first 30s) | 0.59 | 0.64 | 0.64 |

*Table G2: Comparison of Audio Model performance on holdout sample*
*(RMSE for VL; Accuracy for VS)*

Next, in Table G3, we compare the performance of the Image Model (VGG-16) with a simple 4-layer CNN model using thumbnail data. We see a substantial improvement in both RMSE and accuracy when using VGG-16, thus demonstrating the benefits of both transfer learning and a deeper architecture.

| Outcome | Covariate | 4-layer CNN | VGG-16 |
|---|---|---|---|
| Verbal Engagement Level (VL) | Thumbnail | 2.53 | 0.96 |
| Verbal Engagement Sentiment (VS) | Thumbnail | 0.55 | 0.68 |

*Table G3: Comparison of Image Model performance on holdout sample*
*(RMSE for VL; Accuracy for VS)*

**Online Appendix H – Detailed Interpretation Results**

**H.1 Interpretation Results for Text Model**

We estimate Equation 2 (Step 1) and Equation 5 (Step 2) using Ridge Regression to capture heterogeneity in effects across brand names and emotional words. Additional reasons for our choice of Ridge Regression are as follows: (a) Some brands or emotional words may only be used once in our data and hence OLS cannot be used, (b) number of predictors $n_p$ > n for Equation 5 while interpreting the Text model which makes Ridge Regression suitable, and (c) a limitation of other penalized regression methods such as LASSO is that it will cap variable selection at n variables (and not $n_p$) and hence we may miss out on capturing important predictors, (d) LASSO or Elastic Net may miss out on selecting some brands or emotional words if their effect is collinear with other brands or emotional words, (e) Ridge Regression shrinks the non-important predictors towards 0, thus allowing us to identify the relatively more important predictors. The ridge parameter is chosen from a wide array of values with the help of the validation sample, and is then applied to estimate Equation 2 over the full video sample.

Table H1 presents the analysis results for brand and emotion mentions, showing the percentage of total brand names and emotional words whose median coefficient value (across 50 bootstrap iterations) has a positive effect (+) on predicted attention weights (Equation 2) and a positive (+) and/or negative effect (−) on predicted outcome (Equation 5). When calculating this percentage, we ignore brand/emotional words whose value is less than 5% of the magnitude of the maximum predictive effect on the respective outcome in Equation 2 or 5. This 5% *bar* (with robustness discussed ahead) allows us to ignore non-important predictors whose coefficients have been shrunk toward 0 by the Ridge Regression model.

We highlight in grey the cells representing the dominant directional effect of brand/emotional words on predicted outcome. For these cells, we also show (in the row below) the average across all brand/emotional words of the average frequency (over 50 bootstrap iterations) with which a brand/emotional word (that meets the 5% bar) consistently has a positive effect on attention weight and maintains the dominant directional effect on predicted outcome. We highlight in green the cells where the effect on attention weight (Step 1) and predicted outcome (Step 2) is in the same direction (+/−) at least 80% of the time. Our choice of the 80% threshold (as compared to a less conservative 50% even chance threshold) lends more confidence that these results are less likely to be spurious. We validate this threshold in our simulations (Section 7.3). Conversely, we highlight in orange the cells where the effect is the same direction in *only one* of the steps (Step 1 or Step 2) at least 80% of the time, indicating spurious relationships.

| | | | Sentiment of Engagement | | | |
|---|---|---|---|---|---|---|
| | (Eq 3) Attention Weights (AW) (Step 1) | (Eq 5) Predicted Outcome (PO) (Step 2) | Brand names | | Emotional words | |
| | | | Verbal (VS) | Non-Verbal (NVS) | Verbal (VS) | Non-Verbal (NVS) |
| Beginning | + | + | 21.6% | 38.6% | 8.3% | 29.0% |
| | % of time AW +, PO + | | | 76%, 92% | | 77%, 90% |
| | + | - | 42.5% | 29.4% | 23.3% | 19.5% |
| | % of time AW +, PO - | | 85%, 93% | | 79.5%, 93% | |
| Middle | + | + | 20.0% | 26.4% | 11.4% | 22.1% |
| | % of time AW +, PO + | | | 77%, 86% | | 76%, 86% |
| | + | - | 36.4% | 21.8% | 20.3% | 16.5% |
| | % of time AW +, PO - | | 82%, 93% | | 78%, 95% | |
| End | + | + | 18.4% | 33.3% | 9.4% | 20.3% |
| | % of time AW +, PO + | | | 76%, 86% | | 77%, 87% |
| | + | - | 40.2% | 26.4% | 20.5% | 16.9% |
| | % of time AW +, PO - | | 79%, 94% | | 79%, 94% | |

Sample Size: N = 114,536 tokens in beginning 30s, 107,411 tokens in middle 30s, 83,607 tokens in end 30s (Step 1); N = 1620 videos (Step 2).
Bar: 5% (% of max coefficient value used to ignore non-important predictors that have been shrunk towards 0 by Ridge Regression)
Highlighted grey cells: Dominant directional effect of brand/emotional words on predicted outcome.
Highlighted green (orange) cells: Average across all brand/emotional words of the average frequency with which a brand/emotional word has the same directional effect is at least 80% across 50 iterations in both (only one of) Step 1 and Step 2.

*Table H1: Interpretation results of the Text model*

Our findings indicate that most brand mentions in the beginning or middle of the video are often assigned high importance (Step 1) and often have a negative association with predicted VS (Step 2). A graphical visualization of these results is shown in Figure H1. The X axis shows the median value of brand coefficients (across 50 bootstrap iterations) from Step 1 of the analysis that correspond to a change in the attention weights. The Y axis shows the median value of brand coefficients (across 50 bootstrap iterations) from Step 2 of the analysis that correspond to a change in the predicted outcome. Our quadrants of interest are Quadrant 1 and 4 that have a positive X axis corresponding to an increase in attention directed to the brand/emotion. The brands/emotions in Quadrant 1 have a positive Y axis corresponding to an increase in the predicted outcome variable, whereas those in Quadrant 4 have a negative Y axis corresponding to a decrease in the predicted outcome variable. We label the data points for the brands/emotions in these quadrants whose coefficient value is at least 30% (*bar* value) of the magnitude of the maximum coefficient value on each axis to avoid cluttering the figures with labels (Note that the results in Table H1 used a *bar* of 5%). The values in brackets below the name of a brand in Q1 (Q4) correspond to the % of time (across 50 bootstrap iterations) the brand coefficient is positive (positive) on the X axis and positive (negative) on the Y axis. As can be seen in the graph, brands are more often present in Q4 as compared to Q1.

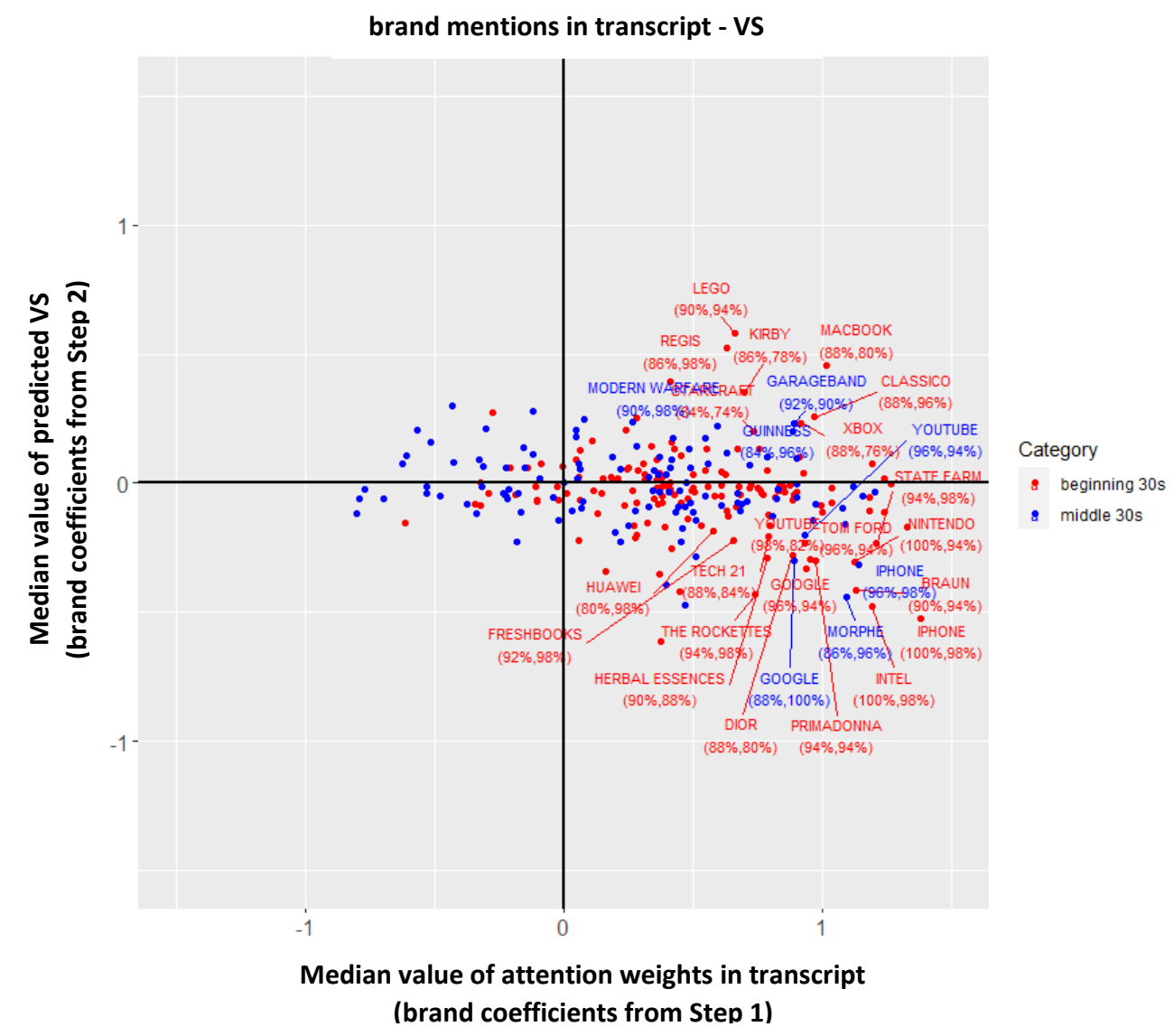

*Figure H1: Interpretation results of Text Model for brand mentions*

The results for the Text model corresponding to the highlighted cells in green in Table H1 are robust to changes in the values of the *bar*. This is because the difference between the percentage of brands that have a negative effect (Quadrant 4) versus a positive effect (Quadrant 1) on predicted VS continue to hold true across a range of values of the *bar*. We show this in Figure H2.

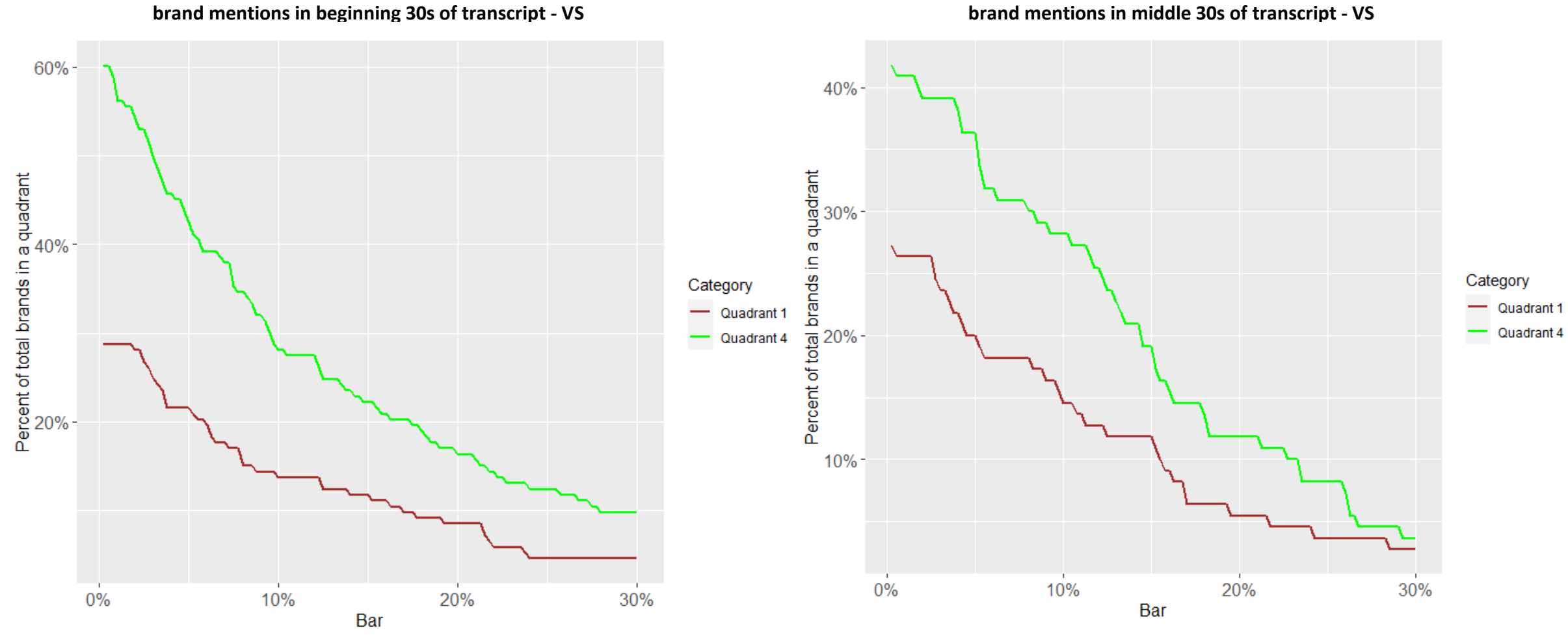

*Figure H2: Robustness of interpretation results of Text Model for brand mentions in beginning 30 seconds and middle 30 seconds*

As can be seen, the green and brown lines do not intersect which demonstrates that the differential effect is in the same direction as we move along the respective X and Y axis of Figure H2. In other words, we demonstrate how our findings for the beginning 30s and middle 30s are robust to brands that have a small or a large negative association with VS. Additionally, we find no robust association between use of brand names or emotional words and measures of VL and NVL.

Overall, we identify two robust relationships (green cells) at the intersection of Steps 1 and 2 that are less likely to be spurious and exclude 10 Type B spurious relationships (orange cells) from Step 2. The presence of Type B but not Type A spurious relationships above the 80% threshold aligns with the theoretical discussion on scaled dot-product attention in Online Appendix A.2.

**H.2 Interpretation Results for Audio Model**

We estimate Equation 3 (Step 1) and Equation 5 (Step 2) using OLS for continuous outcomes and logistic regression for the binary outcome across the entire video sample. Here, we are able to use OLS instead of logistic regression because in these instances we do not model the heterogeneity across brands or emotional words. Instead, we use covariates for 'number of brand mentions' and 'number of emotional word mentions' because textual features in this case are not features of interest and are only used as controls. More concretely, we replace $\sum_{k=1}^{n_b} \beta_{1pk}(BITX_{it}) + \sum_{k=1}^{n_e} \beta_{2pk}(EITX_{it})$ in Equation 5 with $\beta_{1p}(Number\ of\ brand\ mentions_{it}) + \beta_{2p}(Number\ of\ emotion\ mentions_{it})$.

Table H2 presents the results, showing the median value of estimated coefficients, $\hat{\beta}$, across 50 bootstrap iterations. The coefficients indicate the percent change in the (non-log-transformed) outcome when a sound moment is present (Equation 3) or when sound duration increases by one moment (Equation 5). The rows below each coefficient show the percentage of time (across 50 iterations) the corresponding p value is statistically significant. As before, we highlight in green (orange) the cells that are significant at least 80% of the time in both (only one of) Step 1 and Step 2 (and we validate the 80% threshold in Section 7.3).

| | | Verbal Engagement | | | | | | Non-verbal Engagement | | | | | |
|---|---|---|---|---|---|---|---|---|---|---|---|---|---|
| | | Sentiment (VS) | | | Level (VL) | | | Sentiment (NVS) | | | Level (NVL) | | |
| | | Begin | Middle | End | Begin | Middle | End | Begin | Middle | End | Begin | Middle | End |
| Eq (3) Attention Weights (Step 1) | Human $\hat{\beta}_1$ | -3.9% | 17.8% | -0.4% | 2.6% | 9.5% | 0.0% | 0.0% | 24.7% | 1.0% | 0.0% | 2.5% | 9.4% |
| | p <= 0.05 | 49% | 55% | 58% | 60% | 86% | 50% | 56% | 82% | 50% | 52% | 78% | 68% |
| | Music $\hat{\beta}_2$ | 248.3% | 78.8% | -0.6% | 13.9% | 4.7% | 0.0% | 0.0% | 15.9% | 0.1% | 0.5% | 3.3% | 1.3% |
| | p <= 0.05 | 96% | 63% | 52% | 92% | 54% | 74% | 82% | 60% | 80% | 82% | 56% | 64% |
| | H x M $\hat{\beta}_3$ | -54.4% | -46.0% | 3.9% | -5.3% | -7.1% | 0.0% | 0.0% | -5.8% | 0.0% | 0.0% | -1.1% | 0.0% |
| | p <= 0.05 | 90% | 65% | 64% | 78% | 82% | 62% | 68% | 60% | 72% | 60% | 58% | 42% |
| | Animal $\hat{\beta}_4$ | 354.3% | 1325.2% | 111.7% | 0.0% | 0.0% | 0.0% | 0.0% | 52.5% | 0.0% | 0.0% | 1.1% | 10.1% |
| | p <= 0.05 | 90% | 80% | 92% | 46% | 20% | 78% | 52% | 68% | 72% | 38% | 44% | 68% |
| Eq (5) Predicted Outcome (Step 2) | Human $\hat{\beta}_1$ | -2.5% | -6.5% | -4.6% | -0.2% | 0.1% | -0.4% | 0.2% | 0.3% | 0.3% | -0.1% | -0.1% | -0.4% |
| | p <= 0.05 | 29% | 69% | 46% | 38% | 48% | 82% | 78% | 90% | 74% | 40% | 52% | 90% |
| | Music $\hat{\beta}_2$ | 30.6% | 20.5% | 16.4% | -2.0% | -1.4% | -1.5% | -0.3% | 0.0% | -0.2% | -0.9% | -0.7% | -1.1% |
| | p <= 0.05 | 100% | 96% | 96% | 100% | 100% | 100% | 78% | 50% | 70% | 100% | 100% | 100% |
| | H x M $\hat{\beta}_3$ | -18.0% | -7.6% | -15.1% | -0.4% | 0.2% | 0.7% | 0.1% | 0.1% | 0.0% | -0.1% | 0.1% | 0.4% |
| | p <= 0.05 | 82% | 61% | 64% | 52% | 52% | 82% | 58% | 54% | 24% | 34% | 66% | 54% |
| | Animal $\hat{\beta}_4$ | 51.5% | 72.6% | 98.0% | -0.1% | 0.6% | 1.6% | 0.8% | 1.2% | 1.2% | -0.5% | 0.7% | 4.3% |
| | p <= 0.05 | 59% | 57% | 84% | 2% | 38% | 82% | 90% | 96% | 76% | 20% | 74% | 94% |

Sample Size: N = 97,200 moments in beginning 30s, middle 30s or end 30s (Step 1); N= 1620 videos (Step 2)
Highlighted green (*orange*) cells: p value is significant ≤ 0.05 at least 80% of the time across 50 iterations in both (*only one of*) Step 1 and Step 2 and coefficient in Step 1 is positive (because negative attention weight indicates non-important coefficient).

*Table H2: Interpretation results of the Audio Model*

In summary, we identify a subset of 5 robust relationships (green cells) at the intersection of Steps 1 and 2 that are less likely to be spurious. We exclude 18 spurious relationships (orange cells): 4 Type A relationships from Step 1 and 14 Type B relationships from Step 2. The presence of both Type A and Type B spurious relationships above the 80% threshold aligns with the theoretical discussion on additive attention in Online Appendix A.1.

### H.3 Interpretation Results for Video Image Model

We estimate Equation 4 (Step 1) and Equation 5 (Step 2) using OLS for continuous outcomes and logistic regression for the binary outcome on the entire video sample (as done in Section H.2). The results are presented in Table H3, which shows the median value of estimated coefficients, $\hat{\beta}$, across 25 bootstrap iterations. The coefficients reflect the percent change in the (non-log-transformed) outcome when size of an object increases by one percent on average across 30 video frames. The rows below each coefficient show the percentage of time (across 25 iterations) the corresponding p value is statistically significant. As before, we highlight in green (orange) the cells that are significant at least 80% of the time in both (only one of the) steps (and we validate the 80% threshold in Section 7.3). We do not find frequent significant associations for everyday objects such as clothes & accessories and home & kitchen items, so their coefficient values are not reported in Table H3.

| | | Verbal Engagement | | | | | | Non-Verbal Engagement | | | | | |
|---|---|---|---|---|---|---|---|---|---|---|---|---|---|
| | | Sentiment (VS) | | | Level (VL) | | | Sentiment (NVS) | | | Level (NVL) | | |
| | | Begin | Middle | End | Begin | Middle | End | Begin | Middle | End | Begin | Middle | End |
| Eq (4) Mean Gradients (Step 1) | Humans $\hat{\beta}_{11}$ | 0.0% | 0.0% | 0.1% | 0.8% | 0.8% | 0.9% | 0.8% | 0.8% | 0.9% | 0.9% | 0.8% | 0.9% |
| | p <= 0.05 | 100% | 100% | 96% | 100% | 100% | 100% | 100% | 100% | 100% | 100% | 100% | 100% |
| | Faces $\hat{\beta}_{12}$ | -0.2% | -0.1% | -0.2% | 0.8% | 0.5% | 0.3% | 0.6% | 0.4% | 0.3% | 1.0% | 0.6% | 0.2% |
| | p <= 0.05 | 56% | 12% | 60% | 100% | 96% | 56% | 100% | 100% | 60% | 100% | 100% | 48% |
| | Animals $\hat{\beta}_{13}$ | 0.2% | 0.3% | 0.3% | 1.1% | 1.1% | 1.2% | 1.2% | 1.3% | 1.4% | 1.2% | 1.2% | 1.3% |
| | p <= 0.05 | 100% | 100% | 100% | 100% | 100% | 100% | 100% | 100% | 100% | 100% | 100% | 100% |
| | B Logos $\hat{\beta}_{14}$ | 0.0% | -0.1% | 0.0% | 0.1% | -0.2% | 0.2% | -0.1% | -0.2% | 0.0% | 0.0% | -0.1% | 0.1% |
| | p <= 0.05 | 0% | 0% | 0% | 0% | 0% | 0% | 0% | 0% | 0% | 0% | 0% | 0% |
| | P Goods $\hat{\beta}_{15}$ | 0.1% | -0.1% | 0.0% | 0.7% | 0.7% | 0.8% | 0.7% | 0.7% | 0.6% | 0.7% | 0.5% | 0.7% |
| | p <= 0.05 | 0% | 0% | 0% | 100% | 96% | 100% | 100% | 100% | 96% | 100% | 92% | 96% |
| Eq (5) Predicted Outcome (Step 2) | Humans $\hat{\beta}_{11}$ | 0.0% | 1.5% | 1.8% | 0.3% | 0.2% | 0.0% | 0.4% | 0.2% | 0.4% | 0.4% | 0.3% | 0.1% |
| | p <= 0.05 | 8% | 76% | 60% | 64% | 52% | 44% | 92% | 96% | 92% | 96% | 96% | 48% |
| | Faces $\hat{\beta}_{12}$ | 7.4% | 3.4% | 3.3% | 0.9% | 0.0% | 0.6% | 0.4% | 0.5% | -0.8% | 0.9% | 0.9% | 0.4% |
| | p <= 0.05 | 48% | 16% | 12% | 44% | 48% | 24% | 0% | 52% | 32% | 68% | 96% | 28% |
| | Animals $\hat{\beta}_{13}$ | 5.3% | 4.1% | 13.3% | 0.6% | 0.0% | 0.0% | 1.0% | 0.6% | 1.7% | 0.9% | 0.4% | 0.6% |
| | p <= 0.05 | 16% | 4% | 100% | 56% | 16% | 36% | 100% | 88% | 100% | 96% | 36% | 88% |
| | B Logos $\hat{\beta}_{14}$ | 8.6% | 22.8% | 1.6% | -0.6% | 0.0% | 0.8% | 0.4% | -0.4% | -3.8% | 0.3% | -1.2% | 0.1% |
| | p <= 0.05 | 8% | 8% | 0% | 0% | 0% | 4% | 0% | 0% | 72% | 0% | 12% | 4% |
| | P Goods $\hat{\beta}_{15}$ | -1.6% | -3.1% | -3.6% | -0.2% | 0.0% | 0.0% | -0.7% | 0.0% | -0.3% | -0.5% | -0.2% | 0.2% |
| | p <= 0.05 | 8% | 0% | 8% | 4% | 28% | 24% | 88% | 8% | 12% | 84% | 0% | 12% |

Sample Size: N = 5868 objects in beginning 30s, 5659 objects in middle 30s and 5519 objects in end 30s (Step 1); N = 1620 videos (Step 2)
Highlighted green (*orange*) cells: p value is significant ≤ 0.05 at least 80% of the time across 25 iterations in both (*only one of*) Step 1 and Step 2

*Table H3: Interpretation results of the Video Image Model*

Overall, we identify 14 robust relationships (green cells) at the intersection of Steps 1 and 2 that are less likely to be spurious and exclude 25 Type A spurious relationships (orange cells) from Step 1. The presence of Type A but not Type B spurious relationships above the 80% threshold aligns with the theoretical discussion on gradient-based attention in Online Appendix A.3.

## Online Appendix I – Analysis On Entire Video Duration

We use data from the complete video (and not just the beginning, middle and end 30 seconds) as input to the multimodal model shown earlier in Section 5.3. We accomplish this by creating embeddings from each successive 30-second clip of a video. For videos shorter than the longest video in our sample, we add padded embeddings with constant values corresponding to empty segments, so that all videos have the same number of time-steps. We pass embeddings from each time step (30-second clip) as input to the Bi-Directional GRU layers (in Section 7.1.3 we did it for three time steps - beginning, middle and end). We show the results in Table I1 for the difference in permuted prediction error at different video durations analyzed.

| Maximum duration of video analyzed (A) | Correspon d-ing video percentile duration (B) | No. of 30 sec time steps | Modality Comparison | Verbal Engagement | | Non-Verbal Engagement | |
|---|---|---|---|---|---|---|---|
| | | | | Sentiment (VS) | Level (VL) | Sentiment (NVS) | Level (NVL) |
| 5.5 min | 50%[#] | 11 | Text vs Audio | -0.13*** | 0.12*** | 0.14*** | 0.09*** |
| | | | Text vs Video Images | -0.13*** | 0.15*** | 0.14*** | 0.09*** |
| 18.5 min | 90%[#] | 37 | Text vs Audio | -0.01** | 0.06*** | 0.02*** | 0.02*** |
| | | | Text vs Video Images | -0.01** | 0.06*** | 0.02*** | 0.02*** |
| 25 min | 95%[#] | 50 | Text vs Audio | -0.01$^{n.s.}$ | 0.04*** | 0.01*** | 0.01*** |
| | | | Text vs Video Images | -0.01$^{n.s.}$ | 0.04*** | 0.01*** | 0.01*** |
| 271.6 min | 100%[##] | 544 | Text vs Audio | 0$^{n.s.}$ | 0.00001$^{n.s.}$ | 0.00001* | 0.00002** |
| | | | Text vs Video Images | 0$^{n.s.}$ | 0.00003*** | 0.00001** | 0.00003*** |

[#]Embeddings from full duration of B% of videos. Remaining (100-B) % of videos are truncated at A min.
[##]Embeddings from full duration of 100% of videos. Significance codes: ***p< 0.001, **p<0.01, *p<0.05, ^p<0.1, n.s. p ≥ 0.1 (not significant).

*Table I1: Difference in Permuted Prediction Error: Results of Multimodal Model (Embeddings from whole video) (Accuracy for binary outcome – VS; RMSE for continuous outcome – VL, NVS, NVL)*

We find that for video durations up to 18.5 min ($90^{th}$ percentile of duration in our sample), text is significantly more important than audio or video images for all outcomes. However, for longer durations we observe some non-significant effects which is primarily an artifact of the large variance in video duration across our sample.[7] Overall, text continues to be significantly more important than audio or video images in predicting engagement even when we use information from the entire video and not just the beginning, middle or end. Note that the difference in means between audio and video images is minimal and hence the effect is not always significant (similar to our findings in Sections 7.1.2 & 7.1.3), and hence we do not report those values.

[7] For video durations longer than 18.5 min, padded embeddings become a lot more dominant as they occupy a large majority of the time-steps. Hence, when observations are permuted to measure relative feature importance, the overall design of the matrix of covariates does not change enough (because values of padded embeddings remain constant). This results in a decrease in the magnitude of the 'difference in permuted prediction error' and hence we observe a few non-significant effects.

## Online Appendix J – Simulations and Comparison with Benchmarks

For the simulations, we create outcomes that vary based on specific features in text, audio or video images, followed by model training and ex-post interpretation using our approach. We then compare this with benchmark methods, testing whether LASSO and Elastic Net (0.5L1,0.5L2) can identify the data generating process without the need for a two-step process. Bootstrap iterations are conducted 50 times for the Text and Audio models and 25 times for the Video Image model across all approaches.

### J.1 Text Model – Simulation and Benchmarks

To simulate the outcome for the Text Model, we pick a covariate-outcome pair of interest, say "brand mentions in beginning 30 seconds of captions/transcript – NVS". For those observations (of video $t$ and influencer $i$ ) where a brand is *not* mentioned in the beginning 30 seconds, we generate a random normal distribution of log-NVS with mean and standard deviation values that approximate the values observed in our entire sample:

$$brand_absent_{it} \sim N(3.79,1.11)$$

For those observations (of video $t$ and influencer $i$ ) where a brand is mentioned in the beginning 30 seconds, we generate a random normal distribution of log-NVS whose mean is twice the mean and standard deviation is half the standard deviation of the above equation:

$$brand_present_{it} \sim N\left(2 * 3.79, \frac{1.11}{2}\right)$$

We can summarize our simulated outcome of log NVS, $Y_{simulated_{it}}$, for video $t$ by influencer $i$ , as follows:

$$Y_{simulated_{it}} = \begin{Bmatrix} N(3.79,1.11) \text{ } when\ brand\ is\ absent \\ N\left(2 * 3.79, \frac{1.11}{2}\right) \text{ } when\ brand\ is\ present \end{Bmatrix}$$

We follow this up with model training on our data followed by interpretation. The interpretation results are shown in Table J1 (analogous to Table H1). We find that 66.7% of brand names meeting the 5% *bar* (same bar as that in Table H1) are more often associated with an increase in predicted NVS, while only 0.7% of brand names show a decrease. Furthermore, (the average across all brands of) the average frequency with which a brand name (that meets the 5% bar) is associated with an increase in attention weights (Step 1) and predicted outcome (Step 2) is 91% and 97% respectively, both exceeding our 80% threshold.

| | | Text Outcome: NVS | |
|---|---|---|---|
| | | Brand names | Emotional words |
| Two-Step Approach Bar: 5% | Step 1 AW+, Step 2 PO+ | 66.7% | 25.6% |
| | % of time AW +, PO + | 91%, 97% | 78%, 90% |
| | Step 1 AW+, Step 2 PO- | 0.7% | 21.5% |
| LASSO | PO+ | 16.3% | 5.7% |
| | % of time PO+ | 67% | 69% |
| | PO- | 0.7% | 1.8% |
| Elastic Net (0.5L1,0.5L2) | PO+ | 19.6% | 8.0% |
| | % of time PO+ | 69% | 70% |
| | PO- | 0.7% | 2.0% |
| AW: Attention Weight; PO: Predicted Outcome; Bar: 5% (% of max coefficient value used to ignore non-important predictors that have been shrunk towards 0 by Ridge Regression); Highlighted feature name in grey: True data generating process; Highlighted green (*orange*) cells: Average across all brand/emotional words of the average frequency with which a brand/emotional word has the same directional effect is at least 80% across 50 iterations in both (*only one of*) Step 1 and Step 2. | | | |

*Table J1: Simulation and Benchmarks – Text model*

We visually illustrate the results of interpretation in Figure J1. As can be seen in the figure, brands are more often present in Quadrant 1 (increase in attention and an increase in predicted NVS). The results for the Text model are also robust to changes in the value of the *bar* used in Table J1. This is because the difference between the percentage of brands that have a positive effect (Quadrant 1) versus a negative effect (Quadrant 4) on predicted NVS continues to hold true across a range of values of the *bar*. We show this in Figure J2. As can be seen, the green and brown lines do not intersect which demonstrates that the differential effect is in the same direction as we move along the respective X and Y axis of Figure J2. Hence, our findings are robust to brands that have a small or a large effect on predicted NVS.

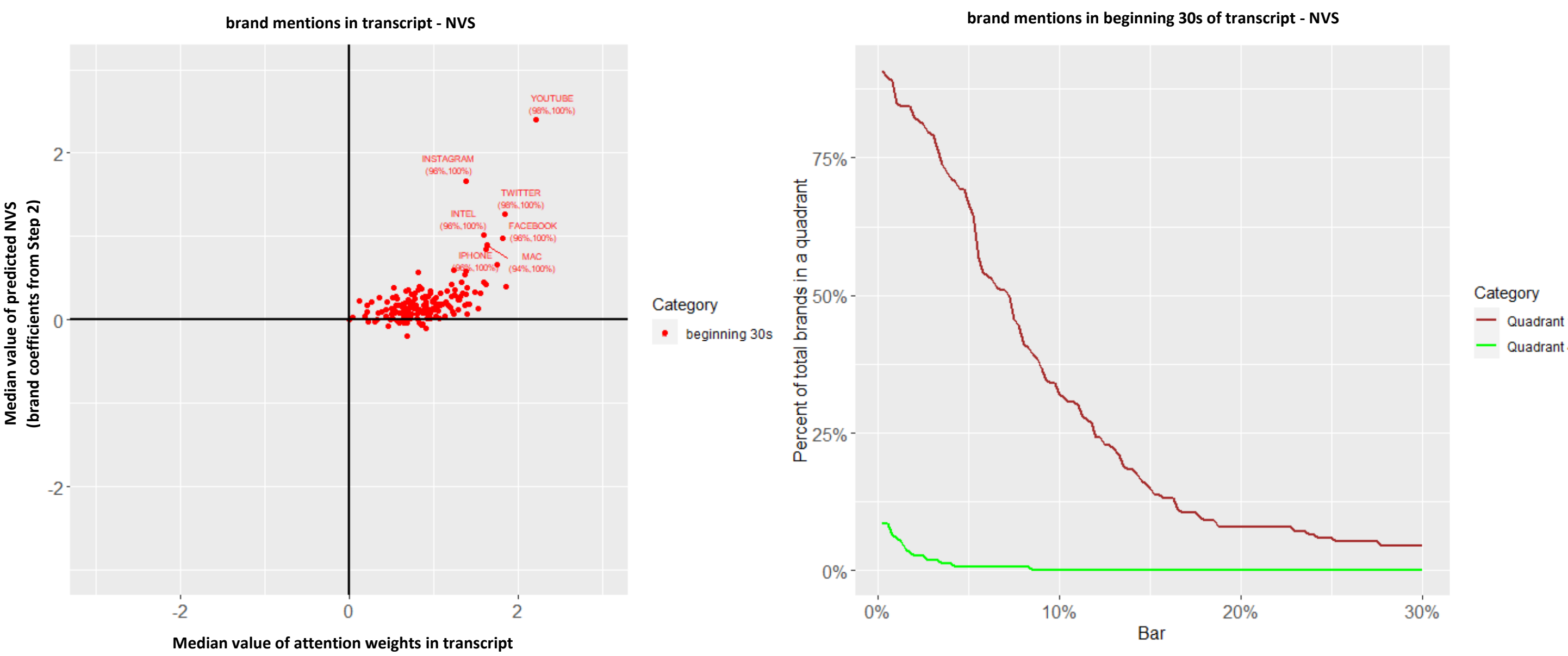

*Figure J1: Interpretation Results of Simulation – Text Model*

*Figure J2: Robustness of Interpretation Results of Simulation – Text Model*

Thus, our interpretation approach using text data is able to recover the true data generating process for brand mentions. For emotion mentions, 25.6% are more often associated with an increase in predicted NVS, while 21.5% show a decrease (as shown in Table J1). However, the average frequency of an increase in attention weights and predicted outcome is 78% and 90% respectively, with the former falling below the 80% threshold. Thus, our two-step approach effectively eliminates the Type B spurious association between emotional words and predicted NVS when using an 80% threshold.

Benchmark methods, LASSO and Elastic Net, also show that brand names are more often associated with an increase (16.3% brands in LASSO, 19.6% brands in Elastic Net) than a decrease in predicted NVS (0.7% brands in both) (as shown in Table J1). However, both methods also associate emotional words with an increase in predicted NVS (5.7% emotional words in LASSO and 8% emotional words in Elastic Net) more than a decrease (1.8% emotional words in LASSO and 2% emotional words in Elastic Net). The average across all brand/emotional words of the average frequency (over 50 bootstrap iterations) with which selected emotional words have a positive effect on predicted outcome is *higher* (69% in LASSO and 70% in Elastic Net) than for selected brand names (67% in LASSO and 69% in Elastic Net), which does not align with the true data generating process (since brand names and not emotional words affect the simulated outcome). This indicates that benchmark methods are not as reliable in eliminating spurious associations with emotional words.

**J.2 Audio and Video Image Model – Simulation and Benchmarks**

To simulate the outcome for the Audio model, we pick a covariate-outcome pair of interest, say "music duration in beginning 30 seconds – VL". We generate a random normal distribution for video $t$ by influencer $i$ with minimum and maximum values that approximate the values observed in our entire sample:

$$v_{it} \sim Uniform(-11.42, -2.18)$$

We then generate the simulated outcome of log VL, $Y_{simulated_{it}}$, for video $t$ by influencer $i$ as follows:

$$Y_{simulated_{it}} = v_{it} - 0.5 * Sum\ of\ CI(Music)_{it}$$

where, $Sum\ of\ CI(Music)_{it}$ is the duration of music sounds in the beginning 30 seconds in video $t$ by influencer $i$.

We follow this up with model training on our data followed by interpretation. The interpretation results, shown in Table J2 (analogous to Tables H2 & H3), include median values of estimated coefficients across 50 bootstrap iterations.

| | | Two-Step Approach | | LASSO | Elastic Net (0.5L1, 0.5L2) |
|---|---|---|---|---|---|
| | | Step 1 | Step 2 | | |
| | Feature | Sign of coefficient, % of time p<=0.05 | Coefficient, % of time p<=0.05 | Coefficient | Coefficient |
| Audio Outcome: VL | Human | -, 74% | 0.016, 58% | 0 | 0 |
| | Music | +, 92% | -0.47, 100% | -198.82 | -2.95 |
| | Human x Music | -, 84% | -0.03, 54% | 0 | -0.25 |
| | Animal | -, 72% | -0.13, 84% | 0 | 0 |
| Video Image Outcome: NVS | Human | +, 100% | 0.13, 100% | 0 | 1.7 |
| | Face | +, 100% | 0.09, 64% | 0 | 1.71 |
| | Animal | +, 100% | 0.05, 88% | 0 | 0 |
| | Brand Logo | -, 0% | -0.003, 0% | 0 | 0 |
| | Packaged Goods | +, 100% | 0.02, 4% | 0 | 0 |
| | Clothes & Acc | +, 100% | 0.04, 96% | 0 | 1.11 |
| | Home & Kitchen | +, 100% | 0.005, 8% | 0 | 0 |
| | Other Objects | +, 100% | 0.006, 4% | 0 | 0 |
| | Joy | -, 4% | -0.24, 8% | 0 | -0.39 |
| | Surprise | -, 100% | -0.60, 72% | 0 | -0.78 |
| | Joy x Face | -, 84% | 0.01, 0% | 0 | -1.31 |
| | Surprise x Face | +, 100% | -0.02, 36% | 0 | -1.74 |

Highlighted feature name in grey: True data generating process;
Highlighted green (*orange*) cells: p value is significant ≤ 0.05 at least 80% of the time across all bootstrap iterations in both (*only one of*) Step 1 and Step 2 and coefficient in Step 1 (for Audio Model) is positive (because negative attention weight indicates non-important coefficient).

*Table J2: Simulation and Benchmarks – Audio and Video Image model*

The median estimated coefficient for music duration in Step 2 is $-0.47$, close to the true value of $-0.5$, and is significant in 100% of iterations for Step 2 and 92% for Step 1. We also eliminate a Type B spurious association with animal sounds, frequently significant in Step 2 but not Step 1. This demonstrates that our interpretation approach using audio data can recover the true data generating process and remove spurious associations using an 80% threshold.

For the Video Image model, we first pick a covariate-outcome pair of interest, say "size of human images in beginning 30 seconds – NVS". We generate a random normal distribution for video $t$ by influencer $i$ with mean and standard deviation values that approximate the values observed in our entire sample:

$$w_{it} \sim Uniform(-0.92, 6.83)$$

We then generate the simulated outcome of log NVS, $Y_{simulated_{it}}$, for video $t$ by influencer $i$ as follows:

$$Y_{simulated_{it}} = w_{it} + 0.25 * SizeObject(Human)_{it}$$

where, $SizeObject(Human)_{it}$ is the mean across 30 frames of the percentage of the image occupied by all objects that are humans in video $t$ made by influencer $i$.

We follow this up with model training on our data followed by interpretation. The interpretation results, shown in Table J2, indicate that the median estimated coefficient for size of human images in Step 2 is 0.13, close to the true value of 0.25, and significant in 100% of the iterations for both Steps 1 and 2. Our estimation process also shortlists two spurious associations related to the size of animals and the size of clothes & accessories (highlighted green). However, we are able to remove seven Type A spurious associations that are frequently significant in Step 1 but not in Step 2 (highlighted orange). Thus, our interpretation approach using image data recovers the true data generating process while eliminating many spurious associations using an 80% threshold.

Table J2 also presents median estimated coefficients obtained from LASSO and Elastic Net for the Audio and Video Image models. LASSO recovers the true feature in the Audio model, but with a highly biased coefficient (-198.82), and fails to recover the true feature in the Video Image model. Elastic Net performs better, recovering the true features in both models with reduced bias, but the coefficient values are not closer to the ground truth as compared to those obtained from our two-step approach. Elastic Net also selects additional spurious features—one in the Audio model and six in the Video Image model—whereas our two-step approach selects no spurious features in the Audio model and only two in the Video Image model. Overall, our two-step approach is more effective in selecting the true feature (with smaller bias) and fewer spurious associations compared to benchmark methods.

## Online Appendix K – Visual Illustration of Salient Regions in Text, Audio and Video Images

We provide illustrations of how measures of attention returned by our deep learning models can be visualized to help influencers identify features for A/B testing.

First, we show an example of how text data can be visually interpreted. In Figure I1, we visualize the attentions weights on the captions/transcript (beginning 30s) from a video of a technology & business influencer. The words are tokenized into word-pieces in the figure as done by the model, and a darker background color indicates relatively higher attention weights. As can be seen in the figure, on average more attention is paid to the word 'iphone' than other words in the text. Note that the model assigns different attention weights to the word 'the' based on the context in which it is used (lower attention in the first line, but higher attention in the last line). While the model also assigns more attention to punctuation marks, such as the apostrophe, these associations may be spurious (as discussed in the manuscript) or may be confounded by word usage unique to the influencer which we control for during ex-post interpretation using influencer fixed effects ($\alpha_i$) in Equation 2. Our model predicts 'not positive' VS for this clip (consistent with the findings in Table H1) and this matches the observed value.

what ' s up guys lew here and this is the iphone 5 camera test if you haven ' t sub
##scribe ##d yet definitely click that button right now so you don ' t miss out on any of my
iphone 5 coverage or tests there ' s a button around you find it click it anyway ##s this
video here we ' re going to be looking at the rear facing camera as well as the newly
improved forward facing camera which now shoot 720 ##p making it a viable option for
shooting v ##log ##s and things like that stick around till the end of the video to get all

*Figure K1: Attention weights in captions/transcript (beginning 30s) of a video*

Next, we illustrate an example of how attention paid to audio moments in a video can be visually interpreted. We focus on the relationship between music and VS. In Figure K2, we show the beginning 30 seconds of the audio clip of a travel influencer using four sub plots. The first plot shows the variations in the amplitude of the 30 second audio wave (sampled at 16 KHz) followed by the spectrogram of the wave where brighter regions correspond to stronger (or louder) amplitudes. Next, we show the interim output of the Audio model with the top 10 sound classes at each moment in the audio, where the darker squares indicate higher probability of observing a sound of that class at that moment. The last plot displays the attention weights corresponding to each moment in the audio clip, where the darker squares indicate higher relative attention placed on that moment while forming an association with VS. As can be seen in the figure, relatively more attention is directed to moments where there is music but no simultaneous speech. The model predicts positive VS for this clip (consistent with the findings in

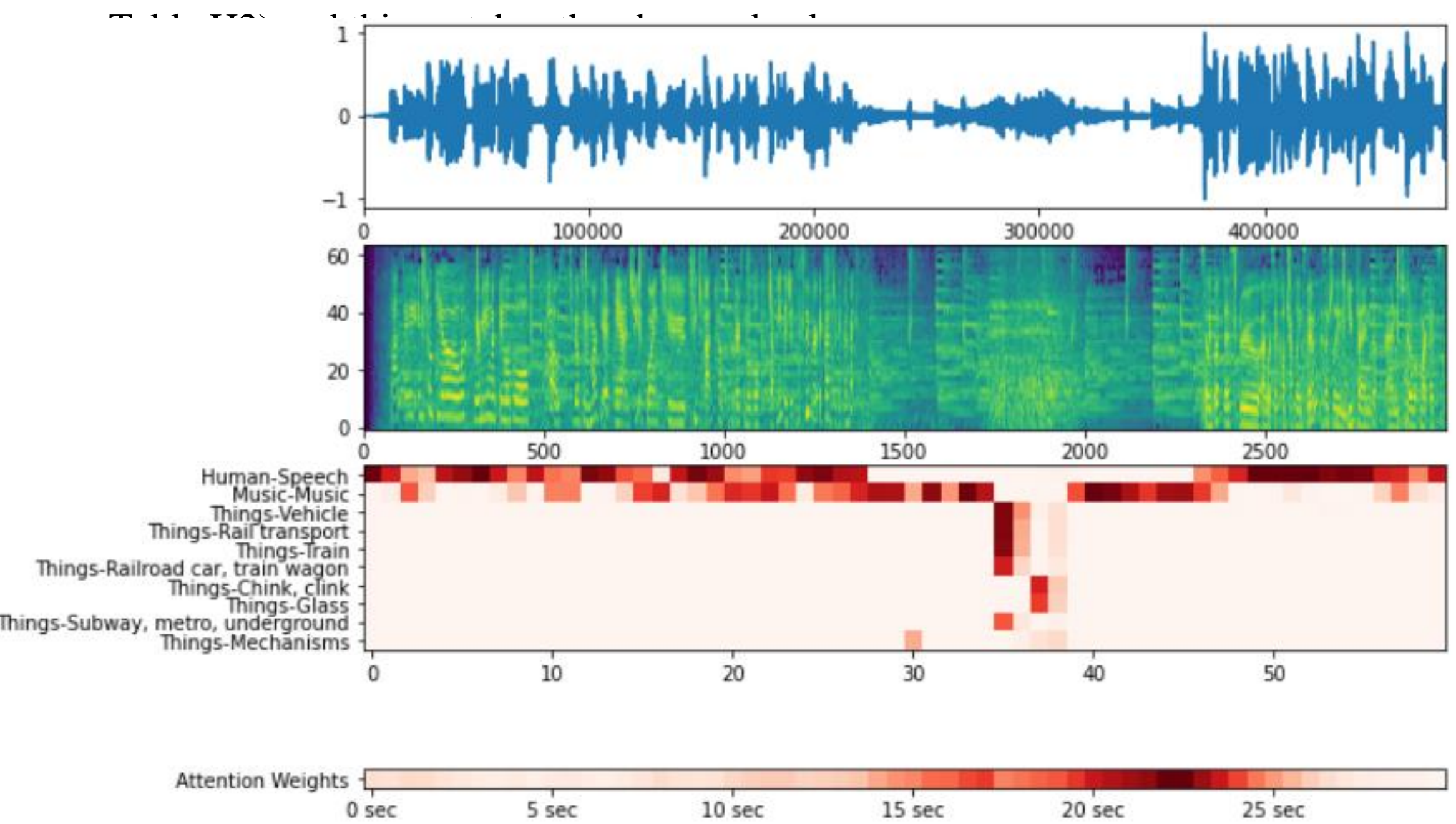

*Figure K2: Attention weights in an audio clip (beginning 30 sec) of a video*

Next, we illustrate how attention paid to image pixels on the video frames of a video can be visually interpreted. We show the first 15 frames @1fps for a video of a parenting influencer in Figure K3. The bottom of the figure shows the heat map (gradient values), where brighter (redder) regions are positively associated with NVS. We find that pixels associated with video images of persons have brighter heat maps, and more attention is often paid to the whole image of the person than just the face of the person (e.g., Frame 3, 5, 6 and 7). This is correlated with the area below the face of the person where the influencer is gesturing with their hands. We can also note the attention paid to video images of packaged goods in Frame 14 and 15. Also note that the model assigns high attention to all the pixels in Frames 1, 9 and 13, and hence they appear completely red. The predicted NVS for this example is 40 likes per dislike which is less than the median NVS of 54 likes per dislike in our sample. This can be expected given that the person is present in only around half the frames and the large size of packaged goods in the frames (which is consistent with the conclusions from Table H3).

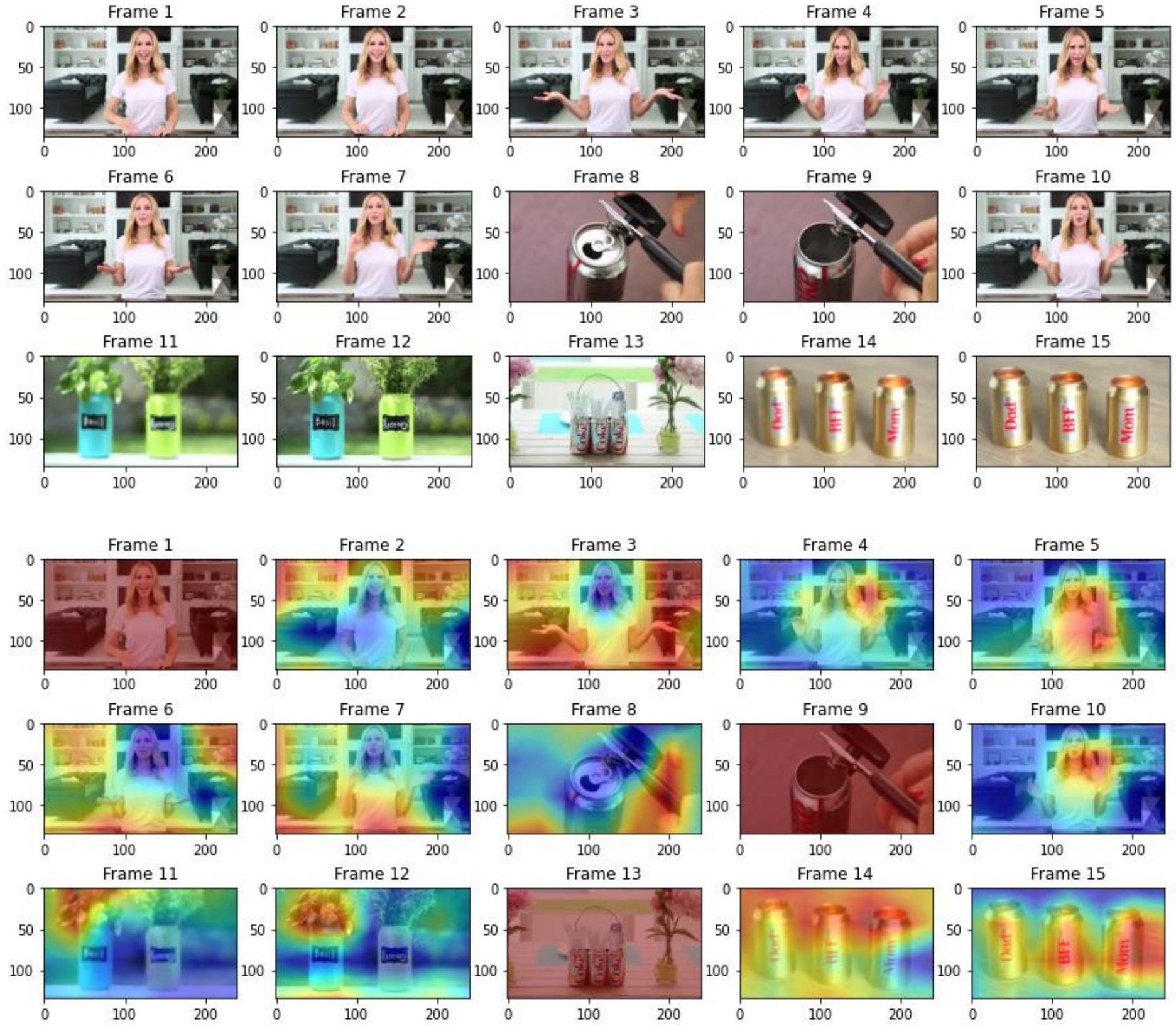

*Figure K3: Gradient heat map in video frames (first 15 seconds) of a video*

These visual illustrations for text, audio and video images can help influencers, agencies and brands identify salient words, sound elements and objects. They can experiment (through A/B testing) by making changes to these salient features and evaluate changes in engagement with the video.

## References to Online Appendices

Alammar, J. (2018, June 27). *The Illustrated Transformer*. http://jalammar.github.io/illustrated-transformer/

Bahdanau, D., Cho, K., & Bengio, Y. (2014). Neural machine translation by jointly learning to align and translate. *arXiv preprint arXiv:1409.0473*.

Berger, J., & Milkman, K. L. (2012). What makes online content viral? *Journal of Marketing Research*, *49*(2), 192-205.

Chakraborty, I., Kim, M., & Sudhir, K. (2022). Attribute sentiment scoring with online text reviews: Accounting for language structure and missing attributes. *Journal of Marketing Research*, *59*(3), 600-622.

Devlin, J., Chang, M.-W., Lee, K., & Toutanova, K. (2018). Bert: Pre-training of deep bidirectional transformers for language understanding. *arXiv preprint arXiv:1810.04805*.

Draelos, R. L., & Carin, L. (2020). Use HiResCAM instead of Grad-CAM for faithful explanations of convolutional neural networks. *arXiv preprint arXiv:2011.08891*.

FileStack. (2019). *Comparing Image Tagging Services: Google Vision, Microsoft Cognitive Services, Amazon Rekognition and Clarifai*. Retrieved April 25 from https://blog.filestack.com/thoughts-and-knowledge/comparing-google-vision-microsoft-cognitive-amazon-rekognition-clarifai/

Gemmeke, J. F., Ellis, D. P., Freedman, D., Jansen, A., Lawrence, W., Moore, R. C., Plakal, M., & Ritter, M. (2017). Audio set: An ontology and human-labeled dataset for audio events. *2017 IEEE International Conference on Acoustics, Speech and Signal Processing*, 776-780.

Gers, F. A., Schmidhuber, J., & Cummins, F. (1999). Learning to forget: Continual prediction with LSTM.

Ghosal, D., Akhtar, M. S., Chauhan, D., Poria, S., Ekbal, A., & Bhattacharyya, P. (2018). Contextual inter-modal attention for multi-modal sentiment analysis. proceedings of the 2018 conference on empirical methods in natural language processing,

Howard, A. G., Zhu, M., Chen, B., Kalenichenko, D., Wang, W., Weyand, T., Andreetto, M., & Adam, H. (2017). Mobilenets: Efficient convolutional neural networks for mobile vision applications. *arXiv preprint arXiv:1704.04861*.

Krizhevsky, A., Sutskever, I., & Hinton, G. E. (2012). Imagenet classification with deep convolutional neural networks. *Advances in neural information processing systems*, *25*.

Li, Y., & Xie, Y. (2020). Is a picture worth a thousand words? An empirical study of image content and social media engagement. *Journal of Marketing Research*, *57*(1), 1-19.

Liu, X., Lee, D., & Srinivasan, K. (2019). Large-scale cross-category analysis of consumer review content on sales conversion leveraging deep learning. *Journal of Marketing Research*, *56*(6), 918-943.

Pennebaker, J. W., Boyd, R. L., Jordan, K., & Blackburn, K. (2015). *The development and psychometric properties of LIWC2015*.

Pilakal, M., & Ellis, D. (2020). *YAMNet*. https://github.com/tensorflow/models/tree/master/research/audioset/yamnet

Selvaraju, R. R., Cogswell, M., Das, A., Vedantam, R., Parikh, D., & Batra, D. (2017). Grad-cam: Visual explanations from deep networks via gradient-based localization. Proceedings of the IEEE international conference on computer vision,

Simonyan, K., & Zisserman, A. (2014). Very deep convolutional networks for large-scale image recognition. *arXiv preprint arXiv:1409.1556*.

Szegedy, C., Vanhoucke, V., Ioffe, S., Shlens, J., & Wojna, Z. (2016). Rethinking the inception architecture for computer vision. Proceedings of the IEEE conference on computer vision and pattern recognition,

Vashishth, S., Upadhyay, S., Tomar, G. S., & Faruqui, M. (2019). Attention interpretability across nlp tasks. *arXiv preprint arXiv:1909.11218*.

Vaswani, A., Shazeer, N., Parmar, N., Uszkoreit, J., Jones, L., Gomez, A. N., Kaiser, Ł., & Polosukhin, I. (2017). Attention is all you need. Advances in neural information processing systems,

Widex. (2016, August 9). *The human hearing range - what can you hear?* https://www.widex.com/en-us/blog/human-hearing-range-what-can-you-hear

Wu, Y., Schuster, M., Chen, Z., Le, Q. V., Norouzi, M., Macherey, W., Krikun, M., Cao, Y., Gao, Q., & Macherey, K. (2016). Google's neural machine translation system: Bridging the gap between human and machine translation. *arXiv preprint arXiv:1609.08144*.

Wu, Z., Jiang, Y.-G., Wang, J., Pu, J., & Xue, X. (2014). Exploring inter-feature and inter-class relationships with deep neural networks for video classification. Proceedings of the 22nd ACM international conference on Multimedia,

Xiao, H. (2018). *BERT-as-service*. https://bert-as-service.readthedocs.io/en/latest/section/faq.html#why-not-the-last-hidden-layer-why-second-to-last

Yue-Hei Ng, J., Hausknecht, M., Vijayanarasimhan, S., Vinyals, O., Monga, R., & Toderici, G. (2015). Beyond short snippets: Deep networks for video classification. *Proceedings of the IEEE conference on computer vision and pattern recognition*, 4694-4702.